\documentclass[runningheads]{llncs}

\usepackage{eccv}

\usepackage{eccvabbrv}

\usepackage{graphicx}
\usepackage{booktabs}

\usepackage{multirow}
\usepackage{makecell}
\usepackage{xcolor}
\usepackage[stable]{footmisc}
\usepackage{color}
\usepackage{colortbl}
\usepackage{caption}
\usepackage{adjustbox}
\usepackage{wrapfig}

\usepackage[accsupp]{axessibility}  % Improves PDF readability for those with disabilities.

\usepackage{hyperref}

\usepackage{orcidlink}

\newcommand{\igray}[1]{\textit{\textcolor{gray}{#1}}}
\definecolor{grayline}{gray}{0.95}

\begin{document}

\title{Phase-Aligned RoPE for Mixed-Resolution Diffusion Transformer}

\titlerunning{Phase-Aligned RoPE for Mixed-Res DiT}

\author{Haoyu Wu\inst{1}\orcidlink{0009-0008-5513-7553} \and
Jingyi Xu\inst{1}\orcidlink{0009-0003-5597-5534} \and
Qiaomu Miao\inst{1}\orcidlink{0000-0002-4091-2181} \and
Dimitris Samaras\inst{1}\orcidlink{0000-0002-1373-0294} \and
Hieu Le\inst{2}\orcidlink{0000-0001-7855-2778}
}

\authorrunning{H.~Wu et al.}

\institute{Stony Brook University, NY, USA \and
UNC-Charlotte, NC, USA}

\maketitle

\begin{abstract}
Rotary positional embeddings (RoPE) are widely used in diffusion transformers (DiTs) to encode spatial relationships, yet their behavior with mixed-resolution tokens remains underexplored.
A natural approach is to rescale token positions from different resolutions into a unified coordinate system before attention, but we show this fails.
Our analysis shows that with RoPE, the attention similarity score is a highly structured and periodic function of token distance, so rescaling distances across resolutions moves token pairs to different regions of this periodic function, leading to incorrect attention scores.
Motivated by this, we introduce Phase-Aligned Mixed-Resolution Attention (PMA), a training-free mechanism that stabilizes mixed-resolution attention. PMA modifies the RoPE position mapping to enforce a consistent positional scale for every query--key pair, ensuring that relative distances are evaluated under a single reference scale. To further improve local coherence near resolution transitions, we incorporate a lightweight boundary refinement module that softly exchanges features across adjacent scales. Experiments on image and video diffusion models validate our analysis and demonstrate consistent improvements in visual fidelity and computational efficiency. Project page: \url{https://hao-yu-wu.github.io/mixed_res/}.

 % \and Third keyword}
\end{abstract}

\section{Introduction}
\label{sec:intro}

\begin{figure}[tb]
  \centering
  \includegraphics[width=1\linewidth]{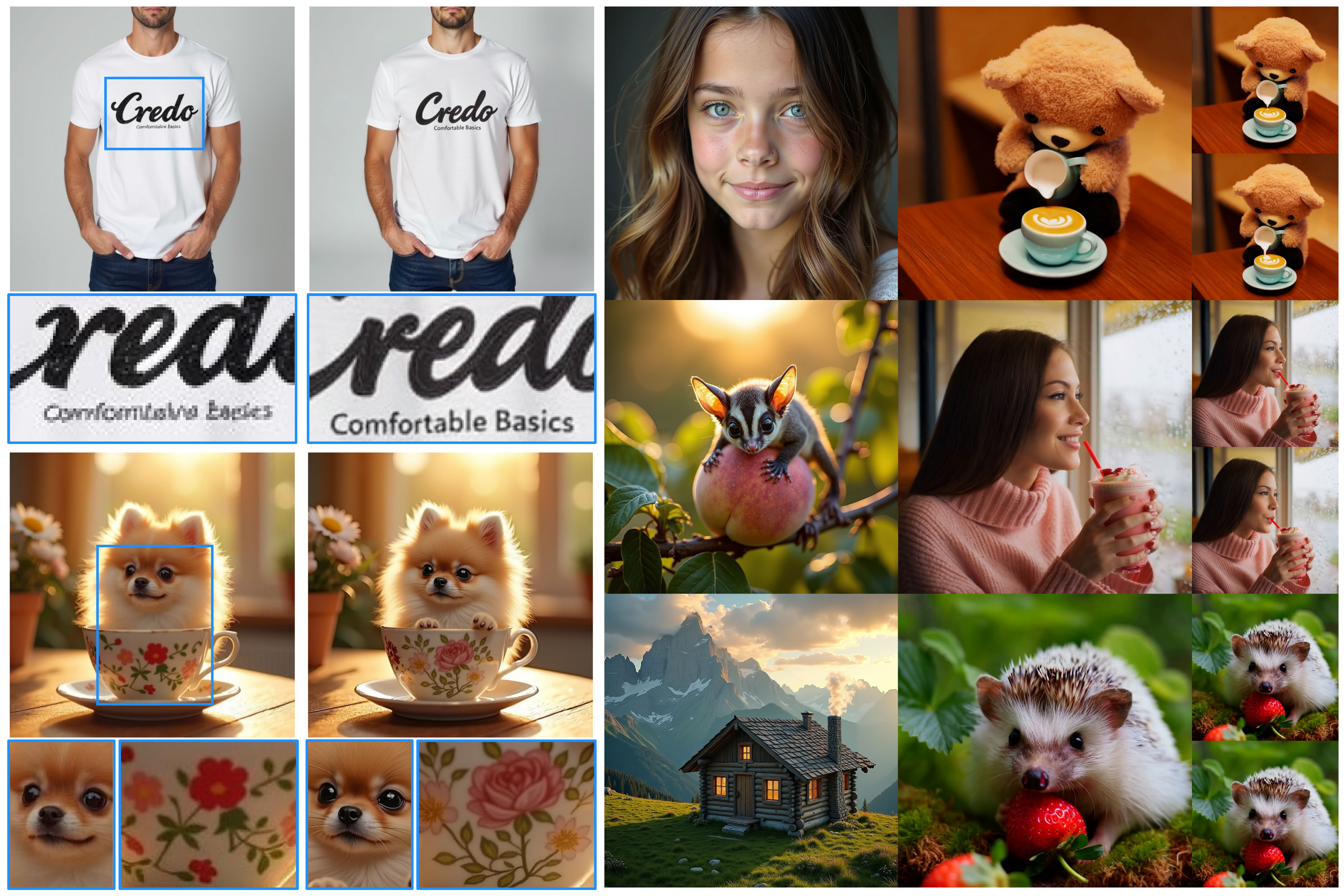}\par
  % \vspace{-0.3mm}
  \makebox[0.225\linewidth]{\centering (a) LR}%
  \makebox[0.225\linewidth]{\centering (b) Ours}%
  \makebox[0.55\linewidth]{\centering (c) Our Generated Samples}\par
  % \vspace{-2mm}
\caption{
Our method enables \textbf{stable mixed-resolution denoising} (b), performing high-resolution (HR) denoising on salient regions (blue boxes) while simultaneously denoising the remaining areas at low resolution (LR). (a) shows a low-resolution baseline. (c) presents image and video samples generated by our method.
}
  \label{fig:teaser}
  % \vspace{-6mm}
\end{figure}

Improving the efficiency of diffusion models has become a central challenge in modern generative modeling. As the resolution of images and videos increases, the quadratic cost of attention quickly dominates computation, making high-resolution generation increasingly expensive \cite{zhang2024token,bolya2023token,Wu_2025_ICCV}. A promising strategy to mitigate this cost is mixed-resolution processing \cite{Ronen2023VisionTW,jeong2025upsample,choudhury2025accelerating,su2025sat}: coarse resolution for background or non-essential regions, and higher resolution for important regions to capture fine-grained details. By allocating computation adaptively, mixed-resolution denoising can, in principle, deliver sharper details without the full cost of uniformly high-resolution attention.

In practice, however, we observe diffusion transformers (DiTs) \cite{labs2025flux1kontextflowmatching,jeong2025upsample,wan2025wan} often struggle under mixed-resolution processing. When tokens from different resolutions are processed together in a single attention operation, the results often exhibit blur and unstable artifacts, even when positions are carefully rescaled into a unified coordinate system, as shown in \cref{sec:diagnosis}. 
We find that the root cause lies in the interaction between attention and Rotary Positional Embeddings (RoPE) \cite{su2024roformer}, which is the standard positional encoding in modern large language models \cite{black2022gpt,touvron2023llama,bai2023qwen,dubey2024llama,guo2025deepseek,lu2024fit} and diffusion transformers \cite{labs2025flux1kontextflowmatching,zhuo2024lumina,hacohen2024ltx,yang2024cogvideox,wan2025wan}.

In essence, RoPE works by rotating token embeddings according to their positions. We find that such transformation introduces a strong position-dependent scale bias into every attention score. As we will show empirically (see \cref{fig:delta-curves}) and theoretically (see \cref{eq:kernel-mixture}), the expected cosine similarity between a query--key pair follows a sinusoidal-like curve $\kappa(\Delta)$, as a function of their relative token distance $\Delta$. This means that token pairs at certain distances tend to have artificially higher or lower similarity scores, on top of what token content alone would dictate. When all tokens share the same resolution, this bias is consistent and the model learns to work with it naturally. The problem arises in mixed-resolution settings: any attempt to unify LR and HR tokens into a shared positional space will distort the native distance scale of at least one group, compressing or stretching their pairwise distances and shifting them to different regions (phases) of the $\kappa(\Delta)$ curve. Some pairs get artificially high attention scores, others get suppressed, rendering the outputs blurry or with random image artifacts.

Motivated by this diagnosis, we introduce \textbf{P}hase-Aligned \textbf{M}ixed-Resolution \textbf{A}ttention (\textbf{PMA}), a training-free mechanism that stabilizes mixed-resolution attention. PMA modifies the RoPE position mapping so that, for every token-level attention, the token positions always share the same reference scale that the model was trained on.
To further improve local continuity near resolution transitions, we incorporate a lightweight boundary refinement module that softly exchanges features across adjacent scales. 
Combined with a practical coarse-to-fine denoising schedule, our approach enables stable, high-fidelity, and efficient generation for images and videos. As illustrated in \cref{fig:teaser}, by supporting targeted regions to be processed at higher resolution, our method unlocks fine-grained detail and new possibilities for controllable, high-quality content generation.

Our contributions are:
\begin{itemize}
\item We uncover and formally analyze a structural limitation of RoPE that makes mixed-resolution attention inherently unstable.

\item We propose Phase-Aligned Mixed-Resolution Attention (PMA), a training-free mechanism that restores stability by enforcing a consistent and native positional scale, along with a lightweight boundary refinement module.

\item We demonstrate stable and efficient mixed-resolution image and video generation with substantially enhanced fine-grained details.
\end{itemize}
\section{Related Work}
\noindent\textbf{Diffusion Transformers (DiTs).}
Replacing U-Nets with transformers has proven highly scalable for diffusion models, with DiTs operating on latent patches and achieving state-of-the-art generation \cite{peebles2023scalable}. 
Multimodal variants further improve text-to-image
\cite{chen2023pixart,esser2024scaling,li2024hunyuan,li2024efficient,tang2025exploring,xie2025sana,labs2025flux1kontextflowmatching}
and text-to-video
\cite{ma2024latte,hacohen2024ltx,zheng2024open,fei2024video,wan2025wan,pondaven2025video}
alignment and visual fidelity~\cite{xu2024sample,xu2024learning,li2024controlling,li2024enhancing}.
% Multimodal variants further improve text–image/video alignment and visual fidelity \cite{chen2023pixart,esser2024scaling,hacohen2024ltx,ma2024latte,li2024hunyuan,zheng2024open,li2024efficient,fei2024video,wan2025wan,labs2025flux1kontextflowmatching,pondaven2025video,tang2025exploring,xie2025sana}. 
% We study DiT efficiency under mixed resolutions, focusing on stability when attention spans tokens from heterogeneous grids.

\noindent\textbf{Positional Encoding.}
A key choice in DiTs is the positional encoding.
% A key choice in DiTs is the positional encoding used in self-attention.
RoPE \cite{su2024roformer} rotates queries and keys using frequency-based phases and is now standard across vision \cite{heo2024rotary,lu2024fit,tian2024u,crowson2024scalable,feng2025romantex} and language transformers \cite{chowdhery2023palm,team2023internlm,dubey2024llama,team2024gemma,chen2024rotary}. 
Work on extending RoPE beyond its trained context---linear interpolation, NTK-aware scaling, YaRN---targets long-range extrapolation \cite{chen2023extending,peng2023yarn,peng2023ntk,ding2024longrope}.
Trainable or Lie-group generalizations \cite{yu2025comrope,ostmeier2024liere} aim to improve robustness across resolutions, while alternatives like ALiBi \cite{press2021train}, XPos \cite{sun2023length}, and other relative schemes pursue greater stability.
However, prior efforts concentrate on language or long-context vision tokens \cite{zhao2025riflex}, instead of our mixed-resolution setting.
% where tokens come from incompatible spatial grids.

\noindent\textbf{Diffusion Acceleration.}
Many acceleration methods reduce either step count or per-step compute and are orthogonal to our focus \cite{du2025fewer}.
Token-reduction techniques merge, prune, or route tokens to lower cost with minimal degradation \cite{bolya2023token,zhang2025training,zhang2024token,you2025layer,changsparsedit,chen2025sparse}.
Model-compression approaches---quantization \cite{shang2023post,li2024svdquant,chen2025q,deng2025vq4dit}, distillation \cite{li2023snapfusion,feng2024relational,zhang2024accelerating}, block pruning \cite{fang2025tinyfusion,ma2024learning,seo2025skrr}, and structured or unstructured pruning \cite{fang2023structural,castells2024ld,zhang2024effortless,wan2025pruning,ganjdanesh2024not}---trade capacity for efficiency.
Feature-caching methods \cite{wimbauer2024cache,lv2024fastercache,ma2024learning,chen2024delta} reuse hidden states across steps \cite{ma2024deepcache,zou2024accelerating,kahatapitiya2025adaptive,liu2025timestep,agarwal2024approximate}, sometimes with selective token routing \cite{liu2025region,you2025layer,lou2024token}.
Sampler advancements cut step counts without retraining \cite{song2020denoising,zhao2023unipc,lu2022dpm,lu2025dpm,zheng2023dpm}, and progressive distillation or consistency models push toward few- or one-step generation \cite{salimans2022progressive,song2023consistency,luo2023latent}.
Coarse-to-fine and multi-stage diffusion pipelines reduce compute while maintaining fidelity, often operating in VAE latents \cite{rombach2022high,chen2024edt}. 
Two-stage \cite{podell2023sdxl,pernias2023wurstchen} and cascaded systems are common \cite{ho2022cascaded,saharia2022photorealistic,teng2023relay,jin2024pyramidal}.
Other approaches explore high-low-high strategies \cite{tian2025training}, region-adaptive sampling \cite{liu2025region}, latent-space super-resolution \cite{jeong2025latent}, and patch-based schemes \cite{ding2024patched}.

To the best of our knowledge, our work is the first to focus on mixed-resolution generation. The closest prior work is RALU \cite{jeong2025upsample}, which uses mixed-resolution as the intermediate stage of a coarse-to-fine pipeline; however, it relies on linear interpolation of RoPE, overlooks RoPE phase mismatch across resolutions, and compensates with additional noise and extra steps.

% Our work follows the coarse-to-fine paradigm but directly tackles attention collapse during mixed-resolution inference, enabling stable low→mixed→full-resolution generation.

% RALU \cite{jeong2025upsample} adopts a three-stage schedule that upsamples artifact-prone regions with noise-timestep rescheduling, but it ignores RoPE phase mismatch across resolutions and compensates with extra noise and steps.
\section{Understanding RoPE Interpolation Failures in Mixed-Resolution Denoising}
\label{sec:diagnosis}

In this section, we investigate why mixed-resolution denoising destabilizes DiTs. 
We show that RoPE imposes a strong positional scale bias on attention scores as a 
function of token distance, following a structured sinusoidal pattern and 
systematically inflates or deflates similarity scores on top of what token content 
alone would dictate.
In mixed-resolution settings, forcing LR and HR tokens into a shared positional coordinate space 
corrupts this distance structure, producing systematically wrong attention scores 
that manifest as blur and visual artifacts.

\subsection{Preliminary - RoPE}
\label{sec:rope-prelim}
We first recall a simple case of RoPE in 1D.
Let an attention head have even dimensionality $d$, and let $\omega_i$ denote the $i$-th angular frequency, typically a geometric sequence: $\omega_i = 10000^{-2i/d}, \ i \in \{0,1,\ldots,d/2-1\}$.
For a scalar position $p\in\mathbb{R}$, RoPE rotates each $(2i,2i{+}1)$ pair by angle $\theta_i(p)=\omega_i\,p$:
\begin{align}
R(\theta) &=
\begin{bmatrix}
\cos\theta & -\sin\theta\\
\sin\theta & \cos\theta
\end{bmatrix}, \\
\mathcal{R}(p) &= 
\mathrm{diag}\big(R(\theta_0(p)), \ldots, R(\theta_{d/2-1}(p))\big).
\end{align}
Given tokens $q,k\in\mathbb{R}^d$, RoPE applies $q\!\rightarrow\!\mathcal{R}(p_q)q$, $k\!\rightarrow\!\mathcal{R}(p_k)k$. The attention score obeys the relative property
\begin{equation}
\big(\mathcal{R}(p_q)q\big)^\top\big(\mathcal{R}(p_k)k\big)
~=~q^\top \mathcal{R}(\Delta)\,k,
\label{eq:rope-rel}
\end{equation}
\textit{i.e.}, RoPE converts absolute positions to a frequency-coded phase that depends only on the relative offset $\Delta=p_k-p_q$.

\subsection{Position Interpolation Fails for Mixed-Resolution Tokens}
\label{sec:lin-interp-fails}

\begin{figure}[t]
\centering
\includegraphics[width=0.96\linewidth]{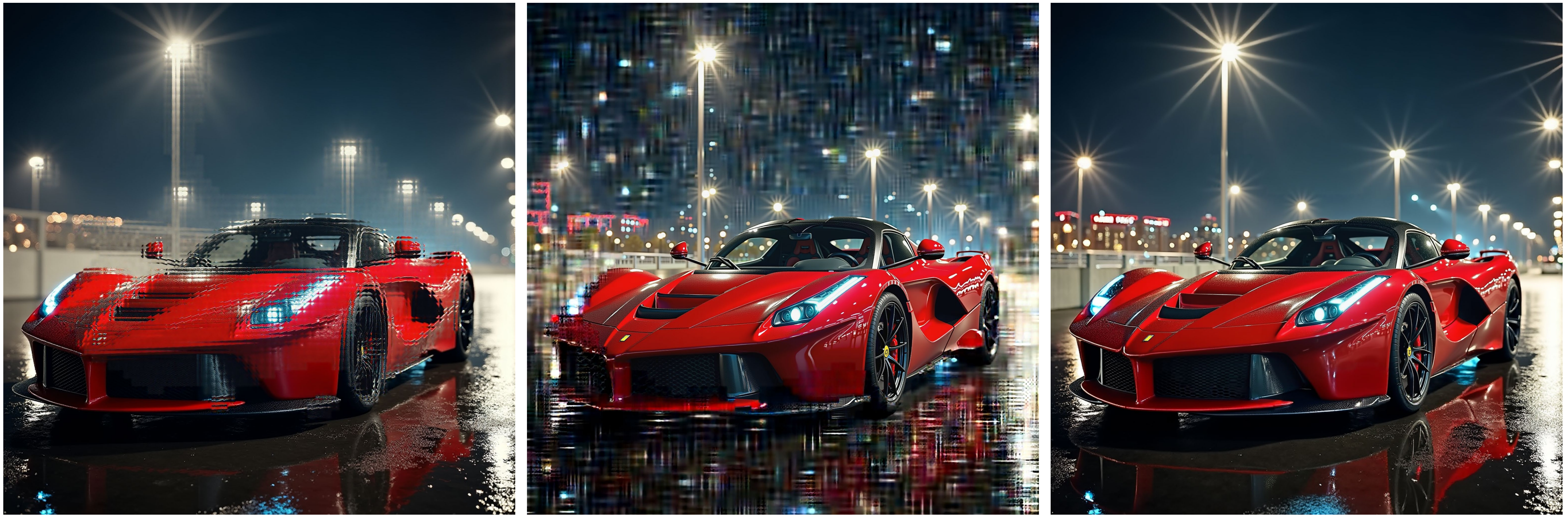}
\makebox[0.32\linewidth]{\centering (a) PI-LR}%
\makebox[0.32\linewidth]{\centering (b) PI-HR}%
\makebox[0.32\linewidth]{\centering (c) Ours}
% \vspace{-2mm}
\caption{Results for RoPE with linear position interpolation (PI) \cite{chen2023extending} to the low- or high-resolution grid, versus our method.}
\label{fig:interp_fail}
% \vspace{-5mm}
\end{figure}

We study the problem of mixed-resolution denoising: allocating high resolution to salient regions and lower resolution elsewhere. How attention behaves in this setting remains an open problem. 
The natural baseline is linear position interpolation (PI)~\cite{chen2023extending}, widely used in language models,  vision transformers, and DiTs, which maps all tokens onto a unified positional space.

Consider a 1D example. Starting from indices [0, 1, 2, {\color{blue}\textbf{3, 4}}, 
5, 6, 7, 8], suppose the middle segment $[3,4]$ is upsampled into a HR block. 
Two natural unification strategies are:

\noindent\emph{(i) Fractional unification (LR grid with fractional HR indices).}
\[
\underbrace{0\ \ 1\ \ 2}_{\text{LR}}\ \ \underbrace{3.0\ \ 3.5\ \ 4.0\ \ 4.5}_{\color{blue}{\text{HR}}}\ \ \underbrace{5\ \ 6\ \ 7\ \ 8}_{\text{LR}}.
\]

\noindent\emph{(ii) Integerized unification (warp LR zones to keep integer indices).}
Here, LR zones are stretched by 2 while the HR block remains dense:
\[
\underbrace{0\ \ 2\ \ 4}_{\text{LR}}\ \ \underbrace{6\ \ 7\ \ 8\ \ 9}_{\color{blue}{\text{HR}}}\ \ \underbrace{10\ \ 12\ \ 14\ \ 16}_{\text{LR}}.
\]

Both can be expressed as a piecewise-affine map on RoPE positions $p$:
\begin{equation}
\phi(p)=a_r + s_r\,(p-b_r),\qquad p\in r,
\label{eq:affine-map}
\end{equation}
where $r$ represents the region (LR or HR), $s_r$ is the per-region scale factor, $b_r$ is the region's original start index, and $a_r$ ensures continuity across boundaries. 
However, as shown in~\cref{fig:interp_fail}, both strategies fail systematically: one 
produces plausible LR regions but collapses HR, and the other does the opposite. 
We explain why next.

% \subsection{Attention Scores vs.\ Relative Distance (\texorpdfstring{$\Delta$}{Delta})}
\subsection{RoPE Imposes a Sinusoidal Scale Bias That Breaks Under Mixed Resolution}
\label{sec:head-measure}

\begin{figure}[t]
\begin{center}
\includegraphics[width=1.0\textwidth]{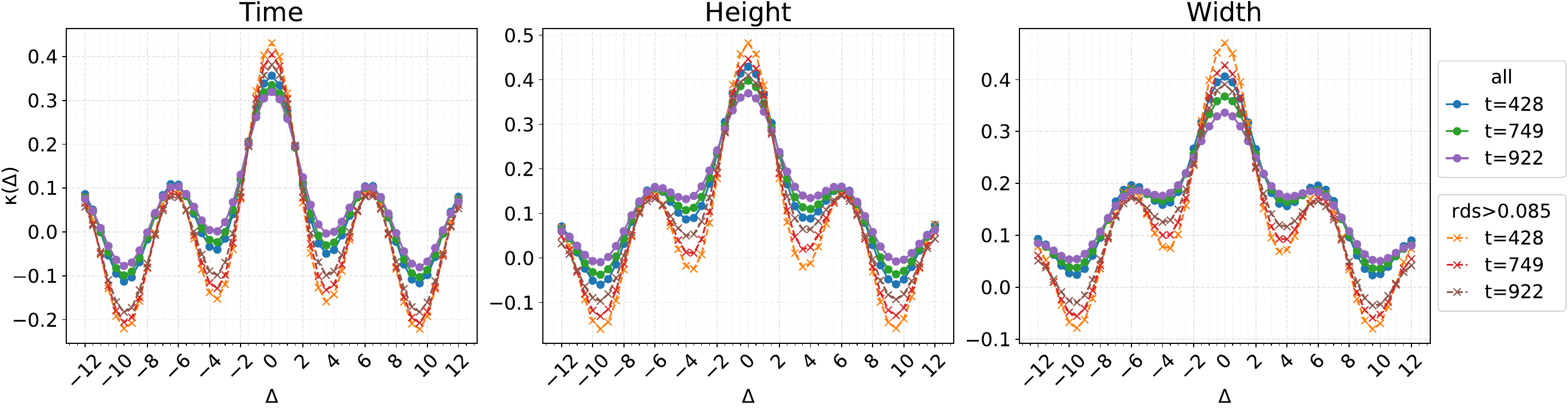}
\end{center}
% \vspace{-6mm}
\caption{\textbf{RoPE imposes a sinusoidal scale bias on the attention function.} We plot the mean normalized attention score $\kappa(\Delta)$ as a 
function of relative distance $\Delta$ on the Wan model~\cite{wan2025wan}, 
measured along three axes (time, height, width) across diffusion steps 
$t\in\{428,749,922\}$. $\kappa(\Delta)$ peaks sharply near $\Delta\approx 0$ and 
oscillates with a clear sinusoidal structure at larger offsets. This is independent 
of token content since we measure this with random token pairs. This bias is amplified in RoPE-dominant heads, \ie, heads with a RoPE-dominance score greater than 0.085 ($rds$), and remains stable across timesteps.}
\label{fig:delta-curves}
% \vspace{-6mm}
\end{figure}

To understand the failure above, we directly measure the relationship between attention scores and relative distance across attention heads in pre-trained DiTs.
Specifically, we compute the expected 
cosine similarity (scale-free proxy for the pre-softmax attention score), between a query and a RoPE-rotated key, as a function of their 
relative token distance $\Delta$:
\begin{equation}
\kappa(\Delta):=\mathbb{E}_{(q,k)}\big[\,
\langle \widehat{q},\, \mathcal{R}(\Delta)\,\widehat{k}\rangle\,\big],
\label{eq:kappa-ours}
\end{equation}
where hats denote $\ell_2$-normalization. Importantly, we average over \textbf{\textit{random}} token pairs. Thus,
$\kappa(\Delta)$ isolates the pure positional scale bias imposed by RoPE---independent 
of what the tokens actually represent. In \cref{fig:delta-curves}, we show the aggregated $\kappa(\Delta)$ across time, height, and width axes, and additionally analyze \emph{RoPE-dominant} heads selected by the RoPE-dominance score (\emph{rds}) 
following~\cite{chen2024rotary}.

\noindent\textbf{Empirical findings.}
As shown in~\cref{fig:delta-curves}, $\kappa(\Delta)$ exhibits three consistent 
properties across all layers and timesteps:
\begin{enumerate}
\item \textbf{Sharp peak near $\Delta \approx 0$.} The similarity is highest for 
nearby tokens and drops steeply within the first 2--3 offsets.
\item \textbf{Sinusoidal oscillations.} Beyond the initial peak, $\kappa(\Delta)$ 
does not decay smoothly—it oscillates, with alternating regions of high and low 
similarity at larger offsets.
\item \textbf{Stable across timesteps.} This structure persists from early to late 
diffusion steps and is amplified in RoPE-dominant heads, confirming it reflects a 
\emph{pretrained phase prior} rather than a denoising artifact.
\end{enumerate}

This sinusoidal scale bias is what makes mixed-resolution attention fundamentally 
problematic. In a single-resolution setting, all tokens share the same distance scale, \ie, the unit of positional spacing, so $\kappa(\Delta)$ acts consistently and the model learns to work within 
it. But when the positions of LR and HR tokens are forced into a shared coordinate space, any remapping inevitably assigns inconsistent pairwise distances to at least one group of tokens: either HR token positions get compressed, or LR token positions get stretched. Because 
$\kappa(\Delta)$ oscillates rather than decays monotonically, these distortions are unpredictable: compressing a distance from 4 to 2 may land on a peak, while compressing from 2 to 1 may land on a trough, or vice versa. Some token pairs become artificially over-attended, others spuriously suppressed, rendering the output blurry or with random visual artifacts. No single rescaling can fix this, because there is no position remapping that can preserve the oscillatory structure of $\kappa(\Delta)$ across two different positional scales simultaneously.

\noindent\textbf{Theoretical interpretation.}
Using the relative form of RoPE (\cref{eq:rope-rel}), the attention score can be 
written as a mixture of sinusoids:
\begin{equation}
\text{score}(q,k,\Delta)
=\sum_i C_i(q,k)\cos(\omega_i \Delta + \phi_i),
\label{eq:kernel-mixture}
\end{equation}
where $\omega_i$ are fixed RoPE frequencies and $(C_i,\phi_i)$ are content- and 
head-dependent coefficients (full proof in supplementary material). This confirms that each attention head implements a \emph{learned sinusoidal phase filter}: high-frequency components produce the steep near-zero peak and strong oscillations observed 
empirically. 
The filter is calibrated to a single positional scale during training. When distances are rescaled, token pairs get shifted to different phases (regions) of the same periodic pattern, so pairs that should align can become misaligned. Mixed-resolution processing therefore samples the filter at two incompatible scales simultaneously, which is the root cause of failures observed in \cref{sec:lin-interp-fails}.

Overall, the issue is not the choice of interpolation scheme but the lack of a uniform positional scale 
for $\Delta$.  
Stable mixed-resolution attention therefore requires all token positions to be expressed on a consistent scale that aligns with the attention heads in pre-trained DiTs.  
We introduce such a mechanism next.

\section{Method}
\label{sec:method}

\begin{figure}[t]
\centerline{\includegraphics[width=1\linewidth]{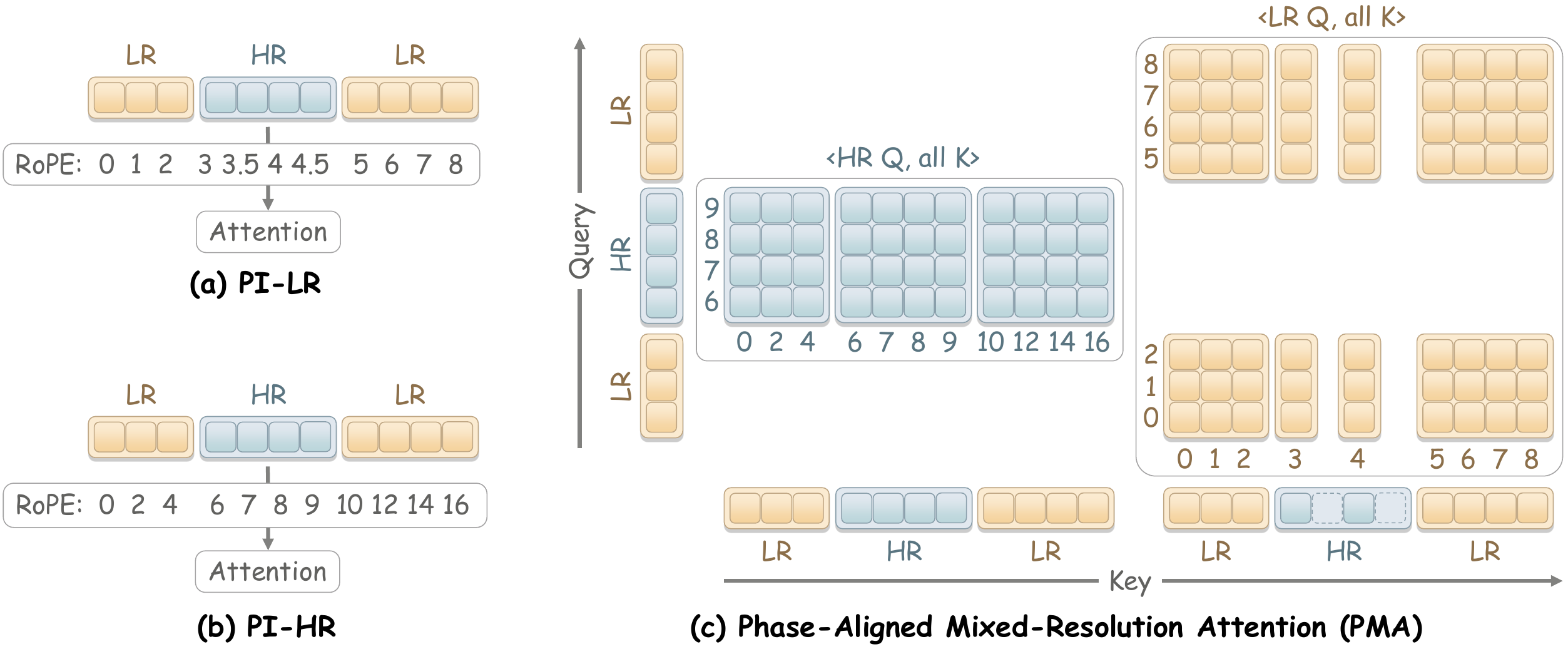}}
% \vspace{-2mm}
\caption{\textbf{Phase-Aligned Mixed-Resolution Attention (PMA).} PMA measures RoPE offsets in the query’s native units by rescaling key positions to the query grid, aligning RoPE phases across resolutions for stable mixed-resolution denoising. (a) and (b) illustrate baselines that interpolate positions to LR and HR grids.}
\label{fig:method}
% \vspace{-6.5mm}
\end{figure}

To stabilize attention with mixed-resolution tokens, we propose Phase-Aligned Mixed-Resolution Attention (PMA), a training-free mechanism that enforces a consistent positional scale across all tokens and is native to pre-trained DiT attention heads.
To further smooth content transitions between LR and HR regions, we also introduce a Boundary Expand-and-Replace module. Together, these components form a stable mixed-resolution denoising pipeline.

\subsection{Phase-Aligned Mixed-Resolution Attention (PMA)}
\label{sec:mix-res-attn}

The core insight from \cref{sec:head-measure} is that attention scores are only 
meaningful when relative token distances are expressed at the same positional scale 
as seen during pre-training. Any remapping that forces LR and HR tokens into a 
shared coordinate space violates this: it inevitably distorts the pairwise 
distances of at least one group of tokens, causing their attention scores to be 
evaluated at the wrong region of the sinusoidal-like curve $\kappa(\Delta)$, \ie, at a different phase of this periodic function.

Thus, the fix is straightforward: rather than remapping all tokens into a single global coordinate space, we perform remapping \emph{locally} for each query--key pair: always expressing the key's position in the query's native positional scale. This ensures that for every dot product attention, the relative offset $\Delta$ is measured in units consistent with pre-training.

% Let $S_q$ and $S_k$ denote the native resolution scale of the query and key tokens, \ie, HR has a larger $S$ than LR, and define the scale ratio $\alpha_{k\!\rightarrow\! q}\;=\;S_q/S_k$.
Let $S$ denote the native resolution scale of tokens, \ie, HR has a larger $S$ than LR, and define the query--key scale ratio: $\alpha_{k\!\rightarrow\! q}\;=\;S_q/S_k$.
For attention between each key and query, we re-index the position of the key to the query's reference grid:
% \[
% p_k^{(q)} \;=\; \alpha_{k\!\rightarrow\! q}\, p_k,\qquad
% \widetilde{q}\;=\;\mathcal{R}(p_q)\,q,\qquad
% \widetilde{k}\;=\;\mathcal{R}\!\big(p_k^{(q)}\big)\,k,
% \]
\begin{equation}
p_k^{(q)} \;=\; \alpha_{k\!\rightarrow\! q}\, p_k,
\label{eq:re_index_k2q}
\end{equation}
and compute the RoPE-based attention score
\begin{equation}
\langle \widetilde{q},\,\widetilde{k}\rangle
~=~ \big\langle \mathcal{R}(p_q)\,q,\; \mathcal{R}\!\big(p_k^{(q)}\big)\,k\big\rangle
~=~ \big\langle q,\, \mathcal{R}\!\big(\alpha_{k\!\rightarrow\! q}\,p_k - p_q\big)\,k\big\rangle.
\label{eq:re_indexed_rope_attn}
\end{equation}

This leads to two cases in practice (\cref{fig:method}):
\begin{itemize}
\item \textbf{HR query vs. all keys.} The HR query's native scale is the 
reference. We stretch LR key positions to match the HR grid. In the toy example, 
HR query positions remain at $[6,7,8,9]$, and LR key positions are stretched: 
$[0,1,2][5,6,7,8] \!\mapsto\! [0,2,4][10,12,14,16]$.
\item \textbf{LR query vs. all keys.} The LR query's native scale is the 
reference. We compress HR key positions to match the LR grid. Since LR queries 
only require coarse context, we also downsample HR keys to the LR grid via strided sampling, with their positions compressed: 
$[6,7,8,9]\!\mapsto\![3,4]$ (stride=2).
\end{itemize}

PMA is training-free and requires no architectural changes: it only modifies token positions before attention. It ensures that, for every attention between a query and a key, their token positions always share the same scale that is consistent with the pre-trained attention heads.
% By construction, every query--key pair now operates on a consistent positional scale.

\subsection{Extension: Boundary Expand-and-Replace (BER)}

% \begin{wrapfigure}{r}{0.6\linewidth}
% % \vspace{-7mm}
% \centering
% \includegraphics[width=0.75\linewidth]{fig/pad_v1.pdf}\par
% % \vspace{-2mm}
% \caption{\textbf{\mbox{Boundary Expand-and-Replace.}}~Around LR--HR boundaries, we dilate the masks and bidirectionally exchange upsampled and downsampled latent content within the narrow band.}
% % \vspace{-7mm}
% \label{fig:pad}
% \end{wrapfigure}

\begin{wrapfigure}{r}{0.45\linewidth} 
\centering
\includegraphics[width=1\linewidth]{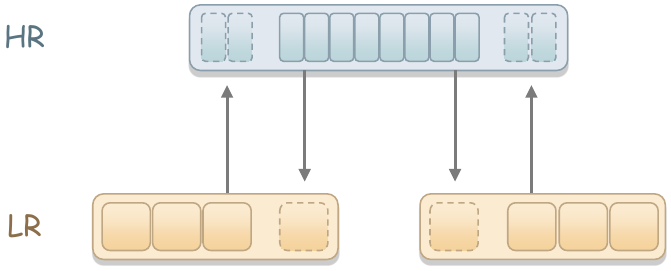}\par
\caption{\textbf{Boundary Expand-and-Replace.} Around LR--HR boundaries, we dilate the masks and bidirectionally exchange upsampled and downsampled latent content within the narrow band.}
\label{fig:pad}
\end{wrapfigure}

PMA resolves the attention scale mismatch between resolutions, but small visual discontinuities may still appear near LR--HR boundaries (\eg, slight density or texture shifts).  
We mitigate these with a lightweight, localized harmonization step that operates only in a narrow band around the boundaries (\cref{fig:pad}).

Given the LR and HR token masks, we first dilate each mask by $n_{\text{pad}}$ (\eg, 2) tokens to form an overlapping boundary band.
At each diffusion step $t$, with noisy latent $x_t$ and current clean estimate $x_0$, we perform a bidirectional content exchange within this band:
\begin{itemize}
\item \textbf{Low$\rightarrow$high.} Upsample the LR portion of $x_0$ using a learned latent upsampler, re-noise it to timestep $t-1$, and replace HR tokens in the band for the next denoising step.
\item \textbf{High$\rightarrow$low.} Downsample the HR portion of $x_0$, re-noise it similarly, and replace the LR tokens in the band.
\end{itemize}

We train the latent up/downsamplers using paired targets obtained by pixel-space resizing and re-encoding, optimized with $\ell_1$ in latent space and $\ell_1+\mathrm{LPIPS}$ in pixel space.
Because the exchanged tokens serve only as localized attention context rather than final output, a small resizer model (25M) is sufficient.

\subsection{Use Case: Saliency-Guided Mixed-Resolution Denoising}

% fig:wan
\begin{figure*}[t]
\centering
\makebox[0.25\linewidth]{\centering PI-LR~\cite{chen2023extending}}%
\makebox[0.25\linewidth]{\centering PI-HR~\cite{chen2023extending}}%
\makebox[0.25\linewidth]{\centering YaRN~\cite{peng2023yarn}}%
\makebox[0.25\linewidth]{\centering Ours}
\includegraphics[width=1\linewidth]{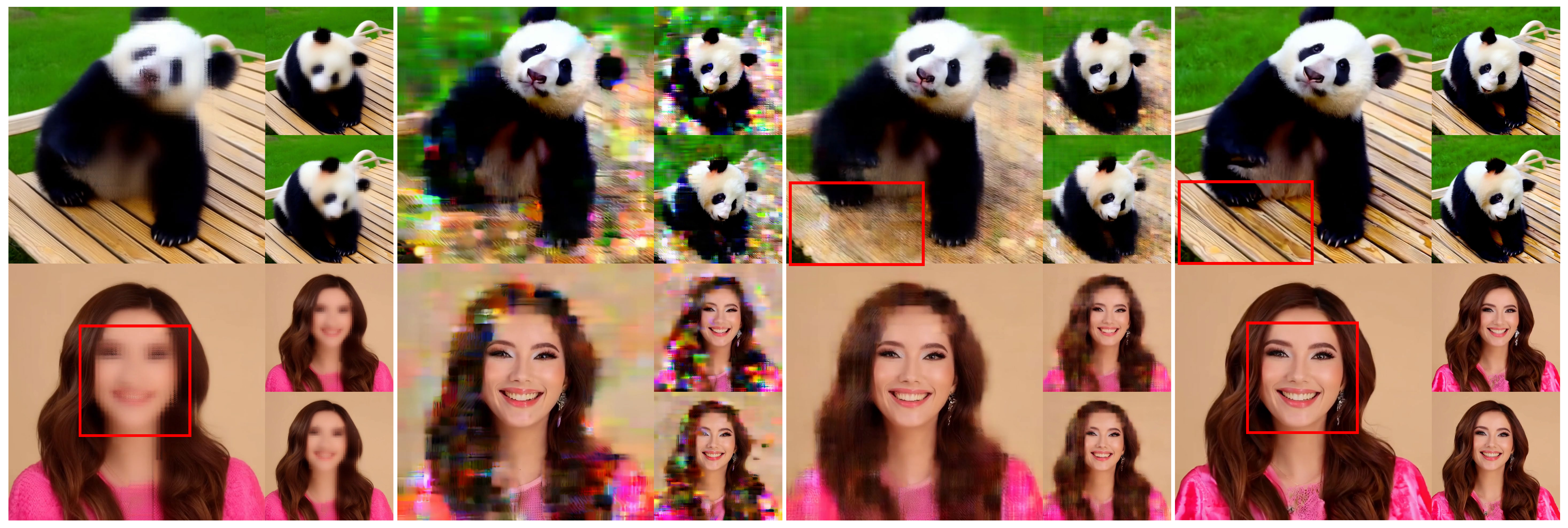}\par
% \vspace{-2mm}
\caption{
\textbf{Mixed-resolution video generation} on Wan 2.1 \cite{wan2025wan}. We compare our method with linear interpolation \cite{chen2023extending} to low- or high-resolution grids (PI-LR/PI-HR), YaRN \cite{peng2023yarn}. As highlighted in red boxes, our method produces the most stable results.
}
\label{fig:wan}
% \vspace{-3mm}
\end{figure*}

% fig:flux
\begin{figure*}[!t]
\centering
\includegraphics[width=1.0\linewidth]{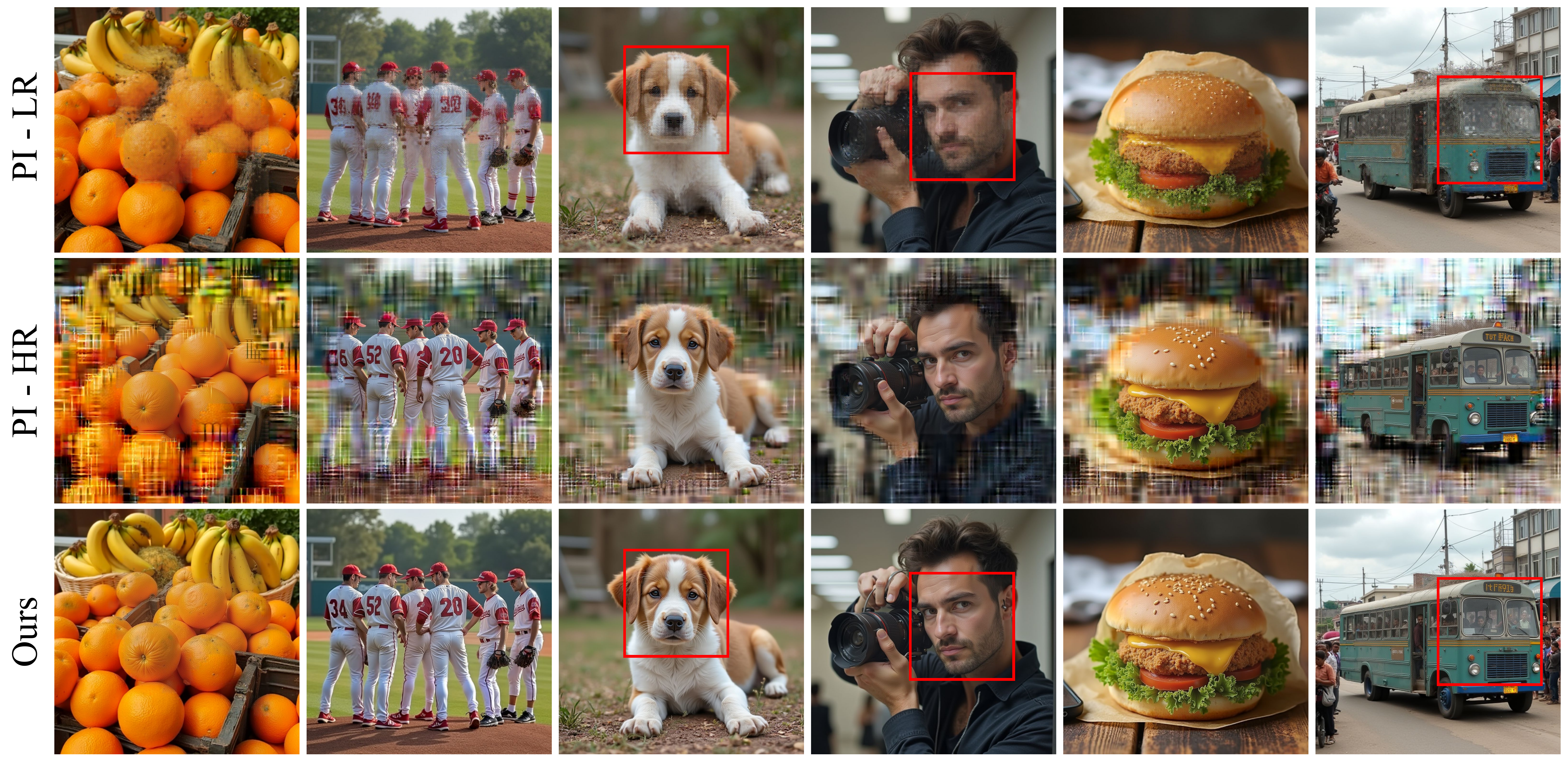}
% \vspace{-6mm}
\caption{
\textbf{Mixed-resolution image generation} on FLUX~\cite{labs2025flux1kontextflowmatching} comparing RoPE with linear interpolation to low- and high-resolution grids against our method.
}
\label{fig:flux}
% \vspace{-6mm}
\end{figure*}

PMA is an attention mechanism that, given LR tokens and HR tokens, stabilizes attention for mixed-resolution generation.
To demonstrate a concrete use case, we propose pairing PMA with an off-the-shelf lightweight saliency model to produce a spatial importance signal and enable mixed-resolution generation.
Our contribution is orthogonal to the specific choice of importance localizer: it only requires a reasonable spatial weighting over regions, such as the one produced by off-the-shelf saliency predictors or user-provided masks. Moreover, these saliency predictors are lightweight relative to the diffusion model itself, adding negligible overhead to the overall pipeline.
We adopt a coarse-to-fine pipeline:
\begin{enumerate}
\item \textbf{Coarse.} Low-resolution generation for a few denoising steps to quickly establish the global structure and motion of the scene.
\item \textbf{Mixed-resolution.} An off-the-shelf saliency model identifies regions of higher importance from the coarse result; these regions are switched to high resolution while the remainder stays at low resolution. Denoising then proceeds using PMA and BER.
\item \textbf{Fine (optional).} A small number of final steps at full resolution to enhance visual quality.
\end{enumerate}
As we will show in \cref{sec:exp}, this simple pipeline already achieves state-of-the-art results, and can also be seamlessly integrated with other diffusion acceleration techniques, highlighting its broad applicability.

% fig:wan_compare_acc
\begin{figure*}[t]
\centering
\makebox[0.25\linewidth]{\centering UniPC \cite{zhao2023unipc}}%
\makebox[0.25\linewidth]{\centering TeaCache \cite{liu2025timestep}}%
\makebox[0.25\linewidth]{\centering MagCache \cite{ma2025magcache}}%
\makebox[0.25\linewidth]{\centering Ours}
\includegraphics[width=1\linewidth]{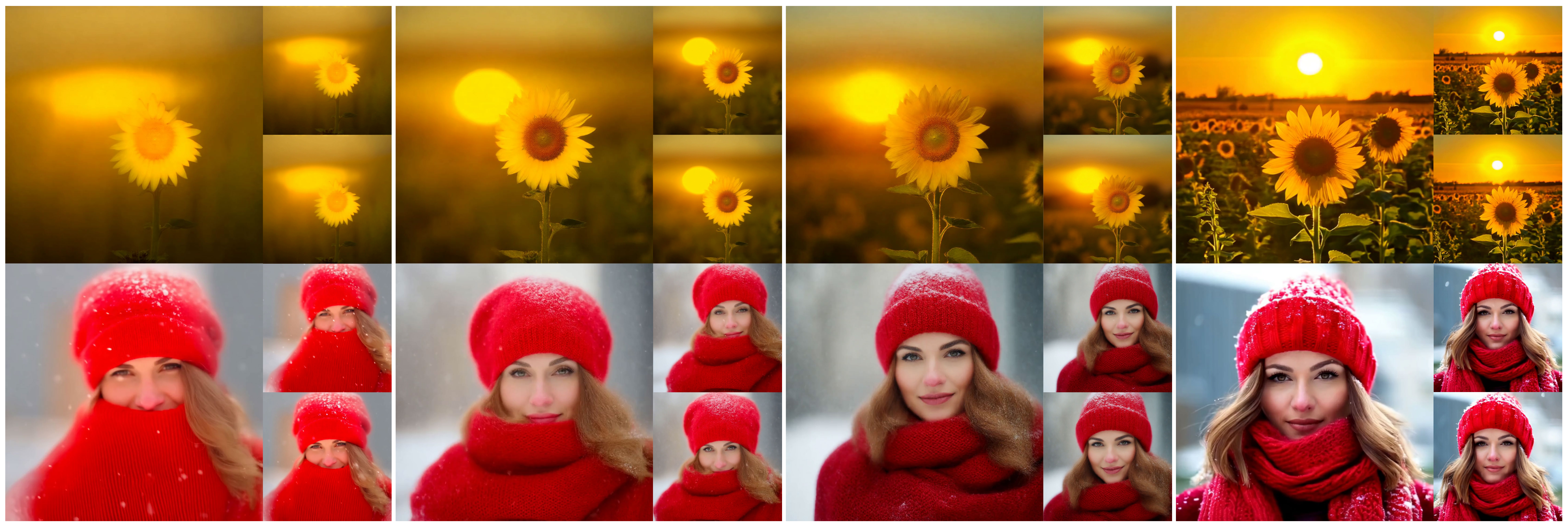}\par
% \vspace{-2mm}
\caption{
\textbf{Comparison with diffusion acceleration methods for video generation} on Wan 2.1 \cite{wan2025wan}. Our method achieves higher quality and greater fidelity.
}
\label{fig:wan_compare_acc}
% \vspace{-3mm}
\end{figure*}

% fig:flux_compare_acc
\begin{figure*}[!t]
\centering
\makebox[0.158\linewidth]{\centering FLUX-50}%
\makebox[0.28\linewidth]{\centering Bottleneck \cite{tian2025training}}%
\makebox[0.28\linewidth]{\centering RALU \cite{jeong2025upsample}}%
\makebox[0.28\linewidth]{\centering Ours}
\includegraphics[width=1\linewidth]{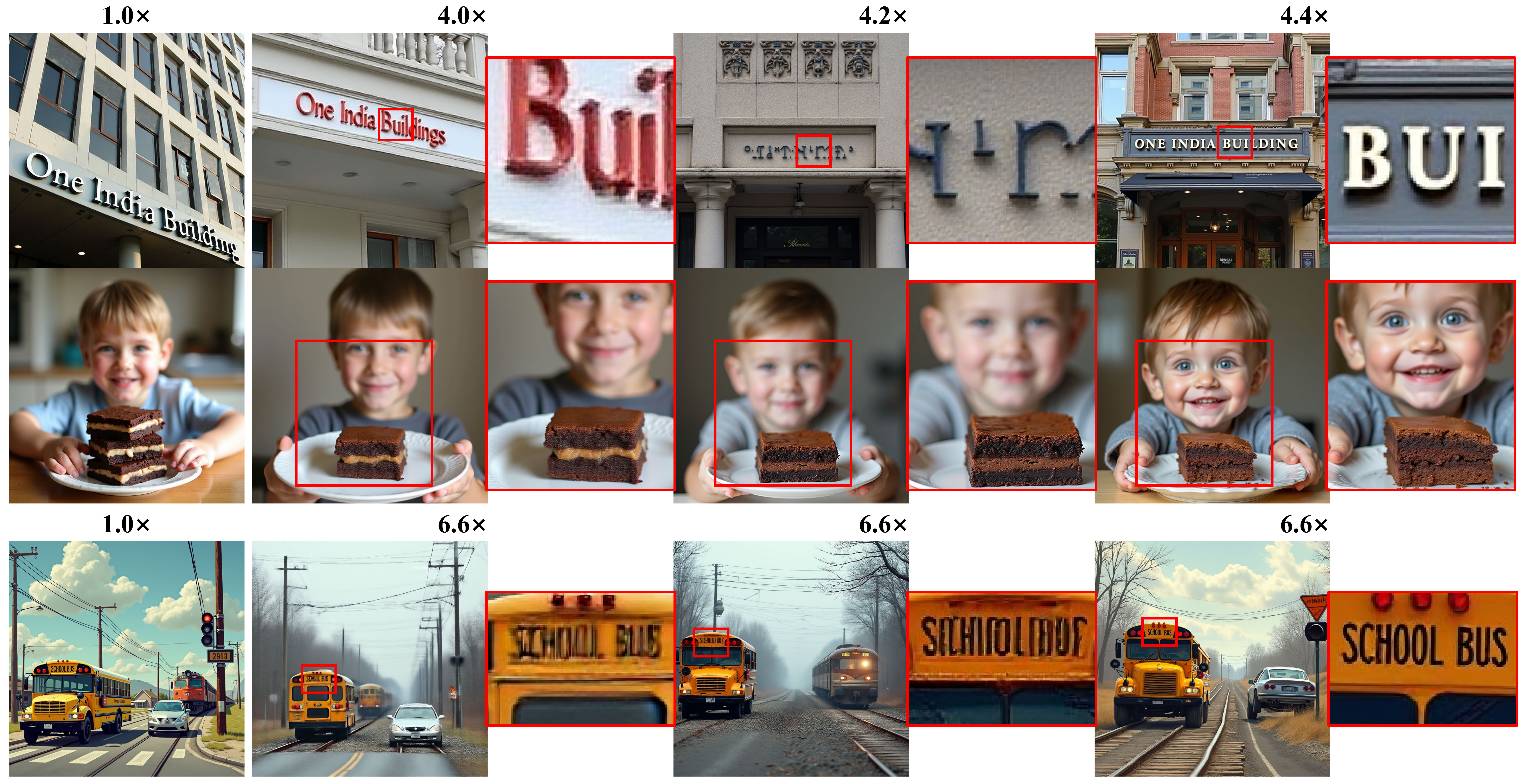}\par
% \vspace{-2mm}
\caption{
\textbf{Comparison with diffusion acceleration methods for image generation} on FLUX \cite{labs2025flux1kontextflowmatching}. We compare at \(4\times\) and \(6\times\) speedups, with 50-step FLUX generations as reference.
Red boxes highlight the improved level of detail in our generation. 
}
\label{fig:flux_compare_acc}
% \vspace{-6mm}
\end{figure*}

\section{Experiments}
\label{sec:exp}

\subsection{Experimental Settings}
\label{ssec:exp_setting}

\noindent\textbf{Model Configurations and Metrics.}
We use Wan2.1-1.3B \cite{wan2025wan} as the text-to-video model and FLUX.1-dev \cite{labs2025flux1kontextflowmatching} as the text-to-image model.
For video evaluation, we use VBench \cite{huang2024vbench} and DOVER \cite{wu2023dover}, and report generation latency, with the full set of VBench prompts. 
For images, we report ImageReward \cite{xu2023imagereward} for photorealism; MUSIQ \cite{ke2021musiq} and CLIP-IQA \cite{wang2022exploring} for image quality assessment; CLIP score \cite{radford2021learning} for prompt alignment, and generation latency.
We use MSCOCO 2014 validation dataset \cite{lin2014microsoft}, with 5K randomly sampled image and caption pairs.

% tab:wan
\begin{table}[t]
\centering
\caption{
\textbf{Quantitative comparison with RoPE interpolation methods} for mixed-resolution denoising on Wan2.1-1.3B \cite{wan2025wan} using DOVER \cite{wu2023dover} and VBench \cite{huang2024vbench}. We include an HR baseline that denoises at high resolution as a reference.
}
\label{tab:wan}
% \vspace{-3mm}
\begingroup
\scriptsize 
\setlength{\tabcolsep}{4.5pt}
\begin{adjustbox}{max width=\columnwidth}
\begin{tabular}{l ccc ccc c}
\toprule
 & \multicolumn{3}{c}{\textbf{DOVER} $\uparrow$} & \multicolumn{3}{c}{\textbf{VBench} $\uparrow$} & \textbf{Time} \\
\cmidrule(lr){2-4} \cmidrule(lr){5-7} %
\textbf{Method} & Aesthetic & Technical & Overall & Quality & Semantics & Total & (s) $\downarrow$ \\
\midrule
\igray{HR} & \igray{99.83} & \igray{10.43} & \igray{79.12} & \igray{80.12} & \igray{62.30} & \igray{76.56} & \igray{172.1} \\
\midrule
PI-LR~\cite{chen2023extending} & 98.10 & 8.01 & 63.39 & 75.93 & 54.92 & 71.73 &   \\
PI-HR~\cite{chen2023extending} & 86.52 & 4.94 & 35.04 & 70.38 & 49.41 & 66.18 &   \\
NTK~\cite{peng2023ntk}         & 92.76 & 5.89 & 44.52 & 71.80 & 52.93 & 68.02 & 43.2 \\
PI+NTK                         & 98.07 & 7.71 & 62.67 & 75.60 & 56.09 & 71.70 &   \\
YARN~\cite{peng2023yarn}       & \underline{98.56} & \underline{8.96} & \underline{66.38} & \underline{76.39} & \underline{56.72} & \underline{72.46} &  \\
\midrule
\textbf{Ours} & \textbf{99.63} & \textbf{10.01} & \textbf{75.34} & \textbf{80.76} & \textbf{62.17} & \textbf{77.04} & 43.2 \\
\bottomrule
\end{tabular}
\end{adjustbox}
\endgroup
% \vspace{-3.5mm}
\end{table}

% tab:flux
\begin{table}[t]
\centering
\caption{\textbf{Quantitative comparison with RoPE interpolation methods} for mixed-resolution denosing on FLUX.1-dev \cite{labs2025flux1kontextflowmatching}. }
\label{tab:flux}
% \vspace{-3mm}
\begingroup
\scriptsize 
\setlength{\tabcolsep}{4.5pt}
\begin{adjustbox}{max width=\columnwidth}
\begin{tabular}{l cccc c}
\toprule
\textbf{Method} & \textbf{ImgReward} $\uparrow$ & \textbf{CLIP-IQA} $\uparrow$ & \textbf{MUSIQ} $\uparrow$ & \textbf{CLIP} $\uparrow$ & \textbf{Time} $\downarrow$ \\
\midrule
\igray{HR} & \igray{1.062} & \igray{0.621} & \igray{70.47} & \igray{31.12} & \igray{3.4 s} \\
\midrule
PI-LR~\cite{chen2023extending}   & 0.659 & 0.411 & 53.96 & \underline{31.41} &  \\
PI-HR~\cite{chen2023extending}   & 0.935 & 0.523 & \underline{70.94} & 31.41 &  \\
NTK~\cite{peng2023ntk}           & \underline{0.953} & 0.542 & 70.62 & 31.37 & 2.4 s \\
PI+NTK                            & 0.810 & 0.479 & 60.23 & \textbf{31.45} &  \\
YARN~\cite{peng2023yarn}         & 0.926 & \underline{0.548} & 69.99 & 31.29 &  \\
\midrule
\textbf{Ours}                     & \textbf{0.978} & \textbf{0.623} & \textbf{71.81} & 31.31 & 2.4 s \\
\bottomrule
\end{tabular}
\end{adjustbox}
\endgroup
% \vspace{-6mm}
\end{table}

% tab:wan_compare_acc
\begin{table}[t]
\centering
\caption{
\textbf{Quantitative comparison with diffusion acceleration methods} on Wan2.1-1.3B \cite{wan2025wan}. We compare our method with caching (TeaCache \cite{liu2025timestep}, MagCache \cite{ma2025magcache}), advanced samplers (UniPC \cite{zhao2023unipc}, DPM++ \cite{lu2025dpm}), and token merging (ToMe \cite{bolya2023token}).
}
\label{tab:wan_compare_acc}
% \vspace{-3mm}
\begingroup
\scriptsize 
\setlength{\tabcolsep}{4.5pt}
\begin{adjustbox}{max width=\columnwidth}
\begin{tabular}{l ccc ccc cc}
\toprule
& \multicolumn{3}{c}{\textbf{DOVER} $\uparrow$} & \multicolumn{3}{c}{\textbf{VBench} $\uparrow$} & \multicolumn{2}{c}{\textbf{Acceleration}} \\
\cmidrule(lr){2-4} \cmidrule(lr){5-7} \cmidrule(lr){8-9} %
% & Aesthetic & Technical & Overall & Quality & Semantics & Total & Time(s) $\downarrow$ & Speed $\uparrow$ \\
\textbf{Method} & Aes. & Tech. & Overall & Qual. & Sem. & Total & Time(s)$\downarrow$ & Speed \\
% \textbf{Method} & Aesthetic & Technical & Overall & Quality & Semantics & Total & Time(s)$\downarrow$ & Speed \\
\midrule
\igray{HR} & \igray{99.83} & \igray{10.43} & \igray{79.12} & \igray{80.12} & \igray{62.30} & \igray{76.56} & \igray{172.1} & \igray{$1.0\times$} \\
\midrule
UniPC \cite{zhao2023unipc} 
& 99.42 & 8.54 & 70.43 & \underline{78.50} & 53.44 & 73.49 & 45.6 & $3.8\times$ \\
DPM++ \cite{lu2025dpm} 
& 99.10 & 7.89 & 66.78 & 77.91 & 51.12 & 72.55 & 44.7 & $3.9\times$ \\
ToMe \cite{bolya2023token} 
& 89.34 & 6.50 & 48.47 & 68.80 & 28.80 & 60.80 & 48.0 & $3.6\times$ \\
TeaCache \cite{liu2025timestep} 
& 99.30 & 8.54 & 69.53 & 76.78 & 53.19 & 72.06 & \underline{43.5} & $\underline{4.0}\times$ \\
MagCache \cite{ma2025magcache} 
& \underline{99.49} & \underline{9.84} & \underline{74.33} & 77.87 & \underline{57.91} & \underline{73.88} & 45.5 & $3.8\times$ \\
\midrule
\textbf{Ours} 
& \textbf{99.63} & \textbf{10.01} & \textbf{75.34} & \textbf{80.76} & \textbf{62.17} & \textbf{77.04} & \textbf{43.2} & $\mathbf{4.0}\times$ \\
\bottomrule
\end{tabular}
\end{adjustbox}
\endgroup
% \vspace{-3.5mm}
\end{table}

% tab: flux_compare_acc
\begin{table}[t]
\centering
\caption{\textbf{Quantitative comparison with diffusion acceleration methods} on FLUX \cite{labs2025flux1kontextflowmatching}.
We compare $4\times$ and $6\times$ speedups against temporal acceleration methods, including TeaCache \cite{liu2025timestep}, MagCache \cite{ma2025magcache}, and ToCa \cite{zou2024accelerating}, spatial acceleration methods such as RALU \cite{jeong2025upsample} and Bottleneck Sampling \cite{tian2025training}, as well as reducing steps of FLUX.
}
\label{tab:flux_compare_acc}
% \vspace{-3mm}
\begingroup
\scriptsize 
\setlength{\tabcolsep}{4.pt}
\begin{adjustbox}{max width=\columnwidth}
\begin{tabular}{l cccc cc}
\toprule 
\textbf{Method} &
\textbf{ImgReward} $\uparrow$ &
\textbf{CLIP-IQA} $\uparrow$ &
\textbf{MUSIQ} $\uparrow$ &
\textbf{CLIP} $\uparrow$ &
\textbf{Time(s)} $\downarrow$ &
\textbf{Speed} \\
\midrule
\igray{FLUX-50} & \igray{1.085} & \igray{0.647} & \igray{71.90} & \igray{30.82} & \igray{11.5} & \igray{$1.0\times$} \\
\midrule
FLUX-12 & 0.985 & 0.588 & 68.94 & 31.19 & 2.8 & $4.1\times$ \\
TeaCache \cite{liu2025timestep} & 0.803 & 0.519 & 63.36 & 30.91 & 2.8 & $4.1\times$ \\
MagCache \cite{ma2025magcache} & \underline{0.993} & 0.512 & 67.37 & 30.98 & 3.0 & $3.9\times$ \\
RALU~\cite{jeong2025upsample}  & 0.940 & \underline{0.592} & \underline{70.06} & 31.04 & \underline{2.7} & $\underline{4.2}\times$ \\
ToCa~\cite{zou2024accelerating}  & 0.956 & 0.498 & 66.49 & \textbf{31.25} & 2.9 & $4.0\times$ \\
Bottleneck~\cite{tian2025training}  & 0.903 & 0.485 & 65.29 & \underline{31.22} & 2.9 & $4.0\times$ \\
\rowcolor{grayline} \textbf{Ours} & \textbf{1.027} & \textbf{0.616} & \textbf{72.29} & 31.16 & \textbf{2.6} & $\textbf{4.4}\times$ \\
\midrule
FLUX-7 & \underline{0.905} & 0.484 & 62.85 & \underline{31.24} & 1.7 & $6.6\times$ \\
MagCache \cite{ma2025magcache} & 0.487 & 0.425 & 52.82 & 31.21 & 2.0 & $5.9\times$ \\
RALU~\cite{jeong2025upsample}  & 0.900 & \underline{0.533} & \underline{66.87} & 31.07 & 1.7 & $6.6\times$ \\
ToCa~\cite{zou2024accelerating} & 0.345 & 0.435 & 50.23 & 30.99 & 1.8 & $6.5\times$ \\
Bottleneck~\cite{tian2025training} & 0.753 & 0.424 & 58.34 & 31.18 & 1.7 & $6.6\times$ \\
\rowcolor{grayline} \textbf{Ours} & \textbf{0.929} & \textbf{0.565} & \textbf{69.01} & \textbf{31.25} & \textbf{1.7} & $\textbf{6.6}\times$ \\
\bottomrule
\end{tabular}
\end{adjustbox}
\endgroup
% \vspace{-3.5mm}
\end{table}

% tab:wan_combine_acc
\begin{table}[t]
\centering
\caption{\textbf{Integration with orthogonal diffusion acceleration methods.} We combine our method with feature caching methods and with the step-distillation model, DMD \cite{yin2024one}. Speedups are reported relative to Wan2.1-1.3B \cite{wan2025wan} for video generation.
}
\label{tab:wan_combine_acc}
% \vspace{-3mm}
\begingroup
\scriptsize 
\setlength{\tabcolsep}{4.5pt}
\begin{adjustbox}{max width=\columnwidth}
\begin{tabular}{l ccc ccc cc}
\toprule
 & \multicolumn{3}{c}{\textbf{DOVER} $\uparrow$} & \multicolumn{3}{c}{\textbf{VBench} $\uparrow$} & \multicolumn{2}{c}{\textbf{Acceleration}} \\
\cmidrule(lr){2-4} \cmidrule(lr){5-7} \cmidrule(lr){8-9} 
\textbf{Method} & Aes. & Tech. & Overall & Qual. & Sem. & Total & Time(s)$\downarrow$ & Speed \\
% \textbf{Method} & Aesthetic & Technical & Overall & Quality & Semantics & Total & Time(s)$\downarrow$ & Speed \\ 
\midrule
Ours & 99.63 & 10.01 & 75.34 & 80.76 & 62.17 & 77.04 & 43.2 & $4.0\times$ \\
\rowcolor{grayline} \quad + TeaCache \cite{liu2025timestep} & 99.59 & 10.05 & 75.34 & 80.33 & 61.28 & 76.52 & 23.9 & $7.2\times$ \\
\rowcolor{grayline} \quad + MagCache \cite{ma2025magcache} & 99.63 & 10.10 & 75.57 & 80.42 & 61.98 & 76.73 & 22.0 & $7.8\times$ \\
\midrule
DMD (8-step) \cite{yin2024one} & 99.96 & 12.70 & 87.13 & 82.51 & 71.31 & 80.27 & 22.6 & $7.6\times$ \\
\rowcolor{grayline} \quad + Ours & 99.95 & 12.62 & 86.32 & 82.63 & 69.72 & 80.05 & 14.1 & $12.2\times$ \\
\addlinespace[1pt]
DMD (4-step) & 99.97 & 13.83 & 88.95 & 82.04 & 72.55 & 80.15 & 10.9 & $15.8\times$ \\
\rowcolor{grayline} \quad + Ours & 99.96 & 14.03 & 87.97 & 82.32 & 71.31 & 80.11 & 5.6 & $30.7\times$ \\
\bottomrule
\end{tabular}
\end{adjustbox}
\endgroup
% \vspace{-6mm}
\end{table}

\noindent\textbf{Implementation Details.}
For video generation, we adopt a two-stage inference scheme with 50 denoising steps: 15 steps at $480$p and 35 mixed-resolution steps, with a high-resolution token ratio of 15\% during the mixed-resolution stage.
For two-stage image generation, we use 15 steps: 5 steps at $512$ and 10 mixed-resolution steps, with a 60\% high-resolution token ratio.
For the three-stage image generation involving coarse, mixed, and fine stages, we adopt a RALU-like schedule \cite{jeong2025upsample} with a small number of steps at the final high-resolution stage and apply noise rescheduling accordingly. To select salient regions, we first use tiny VAE decoders \cite{BoerBohan2025TAEHV} to reconstruct coarse images or videos (only 0.03s for video) following the low-resolution denoising stage, and then apply the off-the-shelf pre-trained DeepGaze model \cite{linardos2021deepgaze} to detect these regions. For training latent resizers, we use a batch size of 1, with Pexels \cite{pexels_license, languagebind_open_sora_plan_v1_1_0} and Aesthetic-Train-V2 datasets \cite{zhang2025diffusion4k,zhang2025ultrahighresolutionimagesynthesis}.

\subsection{Results}
\label{ssec:results}

\noindent\textbf{Comparison with RoPE interpolation methods.} As shown in \cref{tab:wan} and \cref{fig:wan}, our approach yields more stable video generation than RoPE interpolation baselines, including linear position interpolation \cite{chen2023extending} (to low- or high-resolution grids), NTK-aware interpolation \cite{peng2023ntk}, their hybrids, and YaRN \cite{peng2023yarn}. \Cref{tab:flux} and \cref{fig:flux} shows a similar trend for image generation.
% : our method outperforms all RoPE interpolation baselines.

% fig:quality_cost,fig:abl_pad
\begin{figure}[t]
  \centering
  \begin{minipage}[t]{0.51\linewidth}
    \centering
    \includegraphics[width=\linewidth]{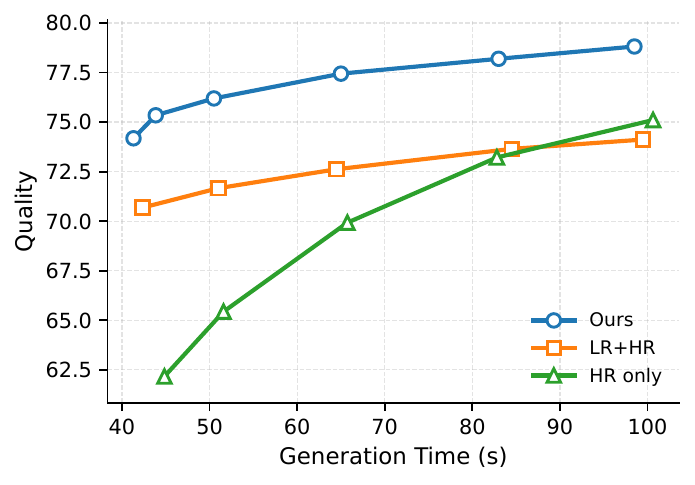}\par
    % \vspace{-2.5mm}
    \captionof{figure}{Quality--cost trade-off on Wan 2.1 \cite{wan2025wan}. We report generation quality using DOVER \cite{wu2023dover} versus generation time.}
    \label{fig:quality_cost}
  \end{minipage}\hfill
  \begin{minipage}[t]{0.45\linewidth}
    \centering
    \includegraphics[width=0.9\linewidth]{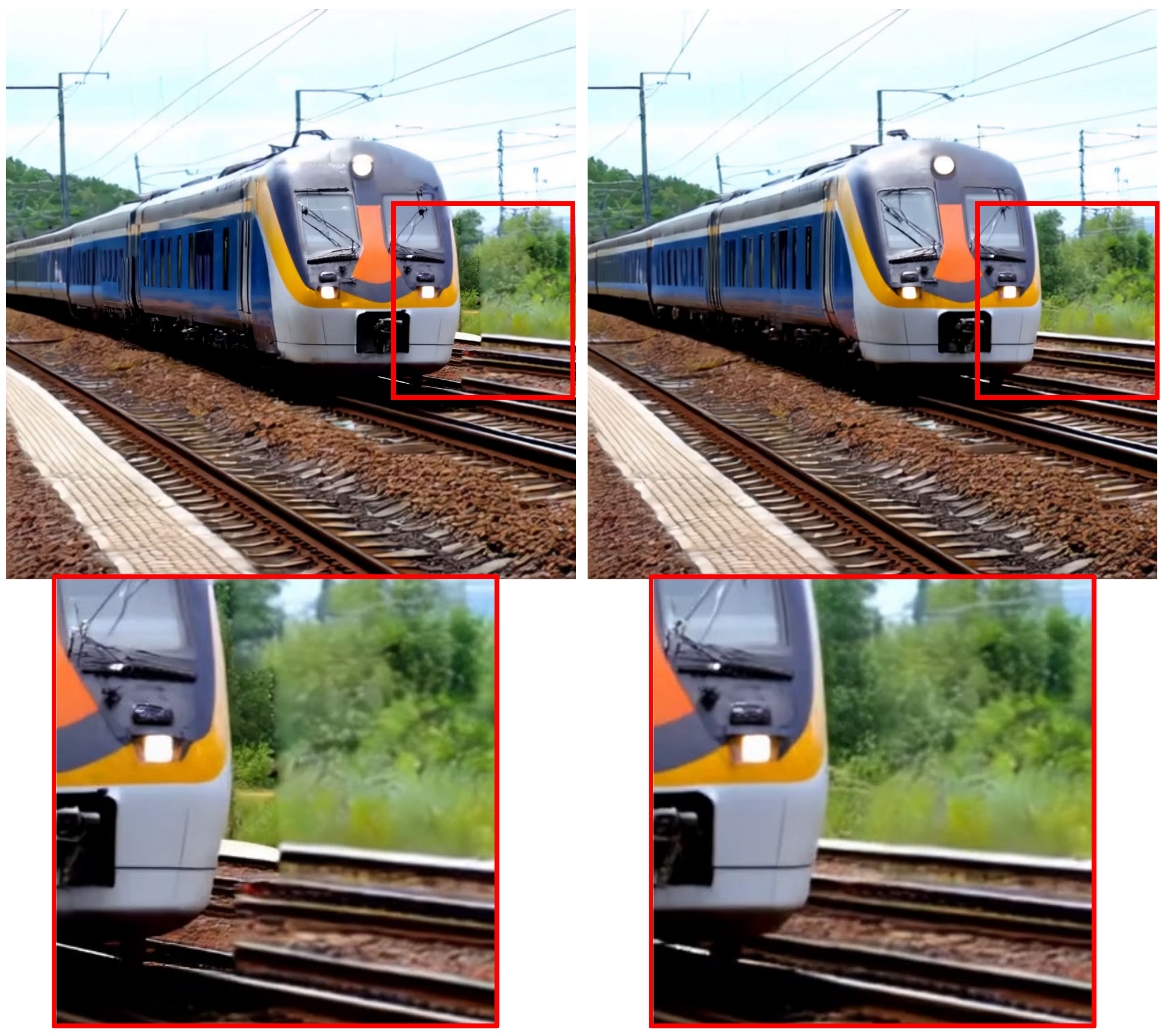}\par
    % \vspace{-2.5mm}
    \captionof{figure}{Generation without (left) vs.\ with (right) Boundary Expand-and-Replace module (BER).}
    \label{fig:abl_pad}
  \end{minipage}
  % \vspace{-2.mm}
\end{figure}

% tab:saliency_robust, tab:abl_pad
\begin{table}[t]
\centering
% ---------- Left table ----------
\begin{minipage}[t]{0.56\linewidth}
\centering
\captionof{table}{
Robustness of our method across saliency models, with each model’s runtime cost and model size reported.
}
\label{tab:saliency_robust}
% \vspace{-2.5mm}
\begingroup
\scriptsize
\setlength{\tabcolsep}{2pt}
\begin{adjustbox}{max width=\linewidth}
\begin{tabular}{l ccc ccc c c}
\toprule
& \multicolumn{3}{c}{\textbf{DOVER} $\uparrow$} & \multicolumn{3}{c}{\textbf{VBench} $\uparrow$} & \textbf{Size} & \textbf{Time} \\
\cmidrule(lr){2-4} \cmidrule(lr){5-7}
\textbf{Model} & Aes. & Tech. & All & Qual. & Sem. & Tot. & (M) & (s) \\
\midrule
DeepGazeI\cite{kummerer2014deep} & 99.64 & 10.26 & 76.13 & 80.79 & 62.01 & 77.04 & 2.5 & 0.01 \\
UNISAL\cite{droste2020unified}      & 99.66 & 10.24 & 76.26 & 80.64 & 61.83 & 76.88 & 3.7 & 0.01 \\
DeepGazeIIE\cite{linardos2021deepgaze} & 99.63 & 10.01 & 75.34 & 80.76 & 62.17 & 77.04 & 104 & 0.27 \\
Center Square & 99.59 & 9.69 & 74.22 & 80.50 & 61.74 & 76.75 & - & - \\
\bottomrule
\end{tabular}
\end{adjustbox}
\endgroup
\end{minipage}\hfill
% ---------- Right table ----------
\begin{minipage}[t]{0.405\linewidth}
\centering
\captionof{table}{
Ablation on overlapping boundary band size $n_{\text{pad}}$ for low (LR) and high resolution (HR).
}
\label{tab:abl_pad}
% \vspace{-2.5mm}
\begingroup
\scriptsize
\setlength{\tabcolsep}{2pt}
\begin{adjustbox}{max width=\linewidth}
\begin{tabular}{cc ccc ccc}
\toprule
\multicolumn{2}{c}{$n_{\text{pad}}$} & \multicolumn{3}{c}{\textbf{DOVER} $\uparrow$} & \multicolumn{3}{c}{\textbf{VBench} $\uparrow$} \\
\cmidrule(r){1-2} \cmidrule(lr){3-5} \cmidrule(l){6-8}
LR & HR & Aes. & Tech. & All & Qual. & Sem. & Tot. \\
\midrule
0 & 0 & 98.90 & 8.94 & 68.43 & 78.08 & 61.38 & 74.74 \\
2 & 2 & \textbf{99.63} & \textbf{10.01} & \textbf{75.34} & \textbf{80.76} & \textbf{62.17} & \textbf{77.04} \\
2 & 4 & 99.62 & 9.88 & 75.18 & 80.69 & 61.76 & 76.90 \\
\bottomrule
\end{tabular}
\end{adjustbox}
\endgroup
\end{minipage}
% \vspace{-1.5mm}
\end{table}

% tab:three_res_wan, tab:three_res_flux
\begin{table}[t]
\centering
% ---------- Left table ----------
\begin{minipage}[t]{0.505\linewidth}
\centering
\captionof{table}{3-mixed-resolution (480 + 960 + 1920p) video generation.}
\label{tab:three_res_wan}
% \vspace{-2.5mm}
\begingroup
\scriptsize
\setlength{\tabcolsep}{2pt}
\begin{adjustbox}{max width=\linewidth}
\begin{tabular}{l ccc ccc c}
\toprule
& \multicolumn{3}{c}{\textbf{DOVER} $\uparrow$} & \multicolumn{3}{c}{\textbf{VBench} $\uparrow$} & \textbf{Time} \\
\cmidrule(lr){2-4} \cmidrule(lr){5-7}
\textbf{Method} & Aes. & Tech. & All & Qual. & Sem. & Tot. & (s) $\downarrow$ \\
\midrule
\igray{Wan-2k}
& \igray{93.99} & \igray{5.85} & \igray{52.14}
& \igray{82.72} & \igray{42.75} & \igray{74.73}
& \igray{1995} \\
\midrule
PI-LR
& 62.29 & 4.73 & 19.87
& \underline{75.18} & \underline{41.39} & \underline{68.42}
& \multirow{4}{*}{288} \\
PI-HR
& 86.31 & \underline{6.48} & 45.23
& 74.01 & 39.96 & 67.20
&  \\
NTK
& 88.69 & 6.37 & \underline{46.76}
& 73.52 & 38.91 & 66.60
& \\
YaRN
& \underline{88.92} & 6.07 & 45.07
& 73.51 & 40.17 & 66.84
&  \\
\midrule
\textbf{Ours}
& \textbf{99.70} & \textbf{10.62} & \textbf{76.78}
& \textbf{81.77} & \textbf{61.55} & \textbf{77.73}
& \textbf{288} \\
\bottomrule
\end{tabular}
\end{adjustbox}
\endgroup

\end{minipage}\hfill
% ---------- Right table ----------
\begin{minipage}[t]{0.46\linewidth}
\centering
\captionof{table}{3-mixed-resolution (512 + 1024 + 2048) image generation. Higher is better for all metrics except for time.}
\label{tab:three_res_flux}
% \vspace{-2.5mm}
\begingroup
\scriptsize
\setlength{\tabcolsep}{2pt}
\begin{adjustbox}{max width=\linewidth}
\begin{tabular}{l cccc c}
\toprule
\textbf{Method} & \textbf{ImgR.} & \textbf{C.IQA} & \textbf{MUSIQ} & \textbf{CLIP} & \textbf{Time(s)} \\
\midrule
\igray{FLUX-2k} & \igray{0.919} & \igray{0.458} & \igray{55.05} & \igray{30.93} & \igray{21.2} \\
\midrule
PI-LR & \underline{0.536} & 0.292 & 40.19 & 28.63 & \multirow{4}{*}{9.8} \\
PI-HR & 0.391 & 0.276 & 45.82 & \underline{30.32} &  \\
NTK & 0.328 & \underline{0.295} & 44.12 & 30.23 &  \\
YaRN & 0.400 & 0.230 & \underline{52.97} & 29.99 & \\
\midrule
\textbf{Ours} & \textbf{0.983} & \textbf{0.468} & \textbf{57.20} & \textbf{31.28} & \textbf{9.8} \\
\bottomrule
\end{tabular}
\end{adjustbox}
\endgroup

\end{minipage}
% \vspace{-5mm}
\end{table}

\noindent\textbf{Comparison with diffusion acceleration methods.} In \cref{tab:wan_compare_acc}, we report video generation results showing that our method outperforms other acceleration approaches, including temporal feature caching methods (TeaCache \cite{liu2025timestep} and MagCache \cite{ma2025magcache}), advanced samplers (UniPC \cite{zhao2023unipc} and DPM++ \cite{lu2025dpm}), and token merging (ToMe \cite{bolya2023token}), with comparable or faster speed.
We also evaluate our three-stage generation setting (coarse, mixed, fine) for image generation in \cref{tab:flux_compare_acc}. We compare $4\times$ and $6\times$ speedups against temporal acceleration methods, including TeaCache \cite{liu2025timestep}, MagCache \cite{ma2025magcache}, and ToCa \cite{zou2024accelerating}, as well as spatial acceleration methods such as RALU \cite{jeong2025upsample} and Bottleneck Sampling \cite{tian2025training}.
As shown in \cref{fig:wan_compare_acc,fig:flux_compare_acc}, while other methods often produce reasonable outputs, our method better preserves key visual details at comparable or higher speed.

\noindent\textbf{Integration with orthogonal diffusion acceleration methods.} Our method is composable with other acceleration techniques. In \cref{tab:wan_combine_acc}, we show it combines effectively with caching and with step-distillation models like DMD \cite{yin2024one}, improving efficiency while maintaining comparable generation quality.

\noindent\textbf{Quality--cost trade-off.}
In \cref{fig:quality_cost}, we compare against LR+HR (LR steps followed by HR steps) and HR-only (all steps in HR). We match generation time for fairness: for our method, we adjust the time budget by changing the ratio of HR tokens, LR+HR uses the same total steps as ours while varying the HR-step ratio, and HR-only varies total steps. Across all time budgets, our method achieves a consistently better quality--cost trade-off.

\noindent\textbf{Ablation on BER}.
As shown in \cref{fig:abl_pad} and \cref{tab:abl_pad}, removing our BER module degrades generation quality. Outputs look plausible but show subtle content discrepancies near the boundary when the underlying token resolution changes. We found padding 2 tokens for both LR and HR to be a reasonable setting.

\noindent\textbf{Robustness to saliency models.}
\Cref{tab:saliency_robust} shows that our method performs consistently well across various off-the-shelf saliency detectors, and even when using a fixed center square as HR regions achieves competitive results. We also show that these detectors are lightweight and add negligible overhead.

\noindent\textbf{Multi-resolution study.}
Our method naturally extends to multi-resolutions. \Cref{tab:three_res_wan} reports 3-mixed-resolution results for video: we outperform the linear interpolation baseline and direct 2K generation while being 7$\times$ faster. \Cref{tab:three_res_flux} reports 3-mixed-resolution image results with a similar trend. 
All methods use resolution extrapolation (CineScale \cite{qiu2025cinescale}, DyPE \cite{issachar2025dype}).

\section{Conclusion}
\label{sec:conclusion}

We show that standard RoPE interpolation fundamentally breaks mixed resolution attention in DiTs: the model is forced to compare phases sampled at incompatible spatial rates, producing cross-scale aliasing and chaotic attention. To eliminate this failure mode, we introduce Phase-Aligned Mixed-Resolution Attention, a training-free mechanism that enforces a consistent and native positional scale, and adds a lightweight Boundary Expand-and-Replace step to smooth resolution transitions. 
Together, these changes enable reliable, high-fidelity image and video generation at reduced cost, supporting more sustainable and scalable deployment of diffusion models.

\section*{Acknowledgements}
We are grateful to Meher Gitika Karumuri, Brandon Smith, Amogh Gupta, and Vidya Narayanan for their insightful comments and valuable discussions. This work was supported in part by NSF grants IIS-2123920, IIS-2212046, and by the CCI startup fund at UNC Charlotte.

\clearpage
\bibliographystyle{splncs04}
\bibliography{main,new}

@String(CVPR  = {IEEE Conf. Comput. Vis. Pattern Recog.})

@String(ICCV  = {Int. Conf. Comput. Vis.})

@String(ECCV  = {Eur. Conf. Comput. Vis.})

@String(NeurIPS = {Adv. Neural Inform. Process. Syst.})

@String(ICLR  = {Int. Conf. Learn. Represent.})

@String(CVPRW = {IEEE Conf. Comput. Vis. Pattern Recog. Worksh.})

@String(AAAI  = {AAAI})

@String(ICPR  = {Int. Conf. Pattern Recog.})

@String(ICASSP=	{ICASSP})

@String(TMLR  = {Trans. Mach. Learn Res.})

@String(CVPR  = {CVPR})

@String(ICCV  = {ICCV})

@String(ECCV  = {ECCV})

@String(NeurIPS = {NeurIPS})

@String(ICLR  = {ICLR})

@String(CVPRW = {CVPRW})

@String(ICPR  = {ICPR})

@String(TMLR  = {TMLR})

@inproceedings{su2025sat,
  title={Sat-hmr: Real-time multi-person 3d mesh estimation via scale-adaptive tokens},
  author={Su, Chi and Ma, Xiaoxuan and Su, Jiajun and Wang, Yizhou},
  booktitle={Proceedings of the Computer Vision and Pattern Recognition Conference},
  pages={16796--16806},
  year={2025}
}

@article{choudhury2025accelerating,
  title={Accelerating Vision Transformers with Adaptive Patch Sizes},
  author={Choudhury, Rohan and Kim, JungEun and Park, Jinhyung and Yang, Eunho and Jeni, L{\'a}szl{\'o} A and Kitani, Kris M},
  journal={arXiv preprint arXiv:2510.18091},
  year={2025}
}

@InProceedings{Wu_2025_ICCV,
    author    = {Wu, Haoyu and Xu, Jingyi and Le, Hieu and Samaras, Dimitris},
    title     = {Importance-Based Token Merging for Efficient Image and Video Generation},
    booktitle = {Proceedings of the IEEE/CVF International Conference on Computer Vision (ICCV)},
    month     = {October},
    year      = {2025},
    pages     = {4983-4995}
}

@article{Ronen2023VisionTW,
  title={Vision Transformers with Mixed-Resolution Tokenization},
  author={Tomer Ronen and Omer Levy and Avram Golbert},
  journal={2023 IEEE/CVF Conference on Computer Vision and Pattern Recognition Workshops (CVPRW)},
  year={2023},
  pages={4613-4622},
  url={https://api.semanticscholar.org/CorpusID:257913635},
  note         = {Accessed: 2026-02-28},
}

@inproceedings{peebles2023scalable,
  title={Scalable diffusion models with transformers},
  author={Peebles, William and Xie, Saining},
  booktitle={Proceedings of the IEEE/CVF international conference on computer vision},
  pages={4195--4205},
  year={2023}
}

@inproceedings{esser2024scaling,
  title={Scaling rectified flow transformers for high-resolution image synthesis},
  author={Esser, Patrick and Kulal, Sumith and Blattmann, Andreas and Entezari, Rahim and M{\"u}ller, Jonas and Saini, Harry and Levi, Yam and Lorenz, Dominik and Sauer, Axel and Boesel, Frederic and others},
  booktitle={Forty-first international conference on machine learning},
  year={2024}
}

@inproceedings{pondaven2025video,
  title={Video motion transfer with diffusion transformers},
  author={Pondaven, Alexander and Siarohin, Aliaksandr and Tulyakov, Sergey and Torr, Philip and Pizzati, Fabio},
  booktitle={Proceedings of the Computer Vision and Pattern Recognition Conference},
  pages={22911--22921},
  year={2025}
}

@inproceedings{tang2025exploring,
  title={Exploring the Deep Fusion of Large Language Models and Diffusion Transformers for Text-to-Image Synthesis},
  author={Tang, Bingda and Zheng, Boyang and Paul, Sayak and Xie, Saining},
  booktitle={Proceedings of the Computer Vision and Pattern Recognition Conference},
  pages={28586--28595},
  year={2025}
}

@article{li2024hunyuan,
  title={Hunyuan-dit: A powerful multi-resolution diffusion transformer with fine-grained chinese understanding},
  author={Li, Zhimin and Zhang, Jianwei and Lin, Qin and Xiong, Jiangfeng and Long, Yanxin and Deng, Xinchi and Zhang, Yingfang and Liu, Xingchao and Huang, Minbin and Xiao, Zedong and others},
  journal={arXiv preprint arXiv:2405.08748},
  year={2024}
}

@misc{chen2023pixart,
      title={PixArt-$\alpha$: Fast Training of Diffusion Transformer for Photorealistic Text-to-Image Synthesis}, 
      author={Junsong Chen and Jincheng Yu and Chongjian Ge and Lewei Yao and Enze Xie and Yue Wu and Zhongdao Wang and James Kwok and Ping Luo and Huchuan Lu and Zhenguo Li},
      year={2023},
      eprint={2310.00426},
      archivePrefix={arXiv},
      primaryClass={cs.CV}
}

@article{zheng2024open,
  title={Open-sora: Democratizing efficient video production for all},
  author={Zheng, Zangwei and Peng, Xiangyu and Yang, Tianji and Shen, Chenhui and Li, Shenggui and Liu, Hongxin and Zhou, Yukun and Li, Tianyi and You, Yang},
  journal={arXiv preprint arXiv:2412.20404},
  year={2024}
}

@article{ma2024latte,
  title={Latte: Latent diffusion transformer for video generation},
  author={Ma, Xin and Wang, Yaohui and Jia, Gengyun and Chen, Xinyuan and Liu, Ziwei and Li, Yuan-Fang and Chen, Cunjian and Qiao, Yu},
  journal={arXiv preprint arXiv:2401.03048},
  year={2024}
}

@article{hacohen2024ltx,
  title={Ltx-video: Realtime video latent diffusion},
  author={HaCohen, Yoav and Chiprut, Nisan and Brazowski, Benny and Shalem, Daniel and Moshe, Dudu and Richardson, Eitan and Levin, Eran and Shiran, Guy and Zabari, Nir and Gordon, Ori and others},
  journal={arXiv preprint arXiv:2501.00103},
  year={2024}
}

@misc{labs2025flux1kontextflowmatching,
      title={FLUX.1 Kontext: Flow Matching for In-Context Image Generation and Editing in Latent Space},
      author={Black Forest Labs and Stephen Batifol and Andreas Blattmann and Frederic Boesel and Saksham Consul and Cyril Diagne and Tim Dockhorn and Jack English and Zion English and Patrick Esser and Sumith Kulal and Kyle Lacey and Yam Levi and Cheng Li and Dominik Lorenz and Jonas Müller and Dustin Podell and Robin Rombach and Harry Saini and Axel Sauer and Luke Smith},
      year={2025},
      eprint={2506.15742},
      archivePrefix={arXiv},
      primaryClass={cs.GR},
      url={https://arxiv.org/abs/2506.15742},
      note         = {Accessed: 2026-02-28},
}

@article{wan2025wan,
  title={Wan: Open and advanced large-scale video generative models},
  author={Wan, Team and Wang, Ang and Ai, Baole and Wen, Bin and Mao, Chaojie and Xie, Chen-Wei and Chen, Di and Yu, Feiwu and Zhao, Haiming and Yang, Jianxiao and others},
  journal={arXiv preprint arXiv:2503.20314},
  year={2025}
}

@article{li2024efficient,
  title={Efficient scaling of diffusion transformers for text-to-image generation},
  author={Li, Hao and Lal, Shamit and Li, Zhiheng and Xie, Yusheng and Wang, Ying and Zou, Yang and Majumder, Orchid and Manmatha, R and Tu, Zhuowen and Ermon, Stefano and others},
  journal={arXiv preprint arXiv:2412.12391},
  year={2024}
}

@article{fei2024video,
  title={Video diffusion transformers are in-context learners},
  author={Fei, Zhengcong and Qiu, Di and Li, Debang and Yu, Changqian and Fan, Mingyuan},
  journal={arXiv preprint arXiv:2412.10783},
  year={2024}
}

@article{su2024roformer,
  title={Roformer: Enhanced transformer with rotary position embedding},
  author={Su, Jianlin and Ahmed, Murtadha and Lu, Yu and Pan, Shengfeng and Bo, Wen and Liu, Yunfeng},
  journal={Neurocomputing},
  volume={568},
  pages={127063},
  year={2024},
  publisher={Elsevier}
}

@article{lu2024fit,
  title={Fit: Flexible vision transformer for diffusion model},
  author={Lu, Zeyu and Wang, Zidong and Huang, Di and Wu, Chengyue and Liu, Xihui and Ouyang, Wanli and Bai, Lei},
  journal={arXiv preprint arXiv:2402.12376},
  year={2024}
}

@article{zhuo2024lumina,
  title={Lumina-next: Making lumina-t2x stronger and faster with next-dit},
  author={Zhuo, Le and Du, Ruoyi and Xiao, Han and Li, Yangguang and Liu, Dongyang and Huang, Rongjie and Liu, Wenze and Zhu, Xiangyang and Wang, Fu-Yun and Ma, Zhanyu and others},
  journal={Advances in Neural Information Processing Systems},
  volume={37},
  pages={131278--131315},
  year={2024}
}

@article{chen2023extending,
  title={Extending context window of large language models via positional interpolation},
  author={Chen, Shouyuan and Wong, Sherman and Chen, Liangjian and Tian, Yuandong},
  journal={arXiv preprint arXiv:2306.15595},
  year={2023}
}

@article{peng2023yarn,
  title={Yarn: Efficient context window extension of large language models},
  author={Peng, Bowen and Quesnelle, Jeffrey and Fan, Honglu and Shippole, Enrico},
  journal={arXiv preprint arXiv:2309.00071},
  year={2023}
}

@misc{peng2023ntk,
  title={Ntk-aware scaled rope allows llama models to have extended (8k+) context size without any fine-tuning and minimal perplexity degradation},
  author={Peng, Bowen and Quesnelle, Jeffrey},
  year={2023},
  url={https://www.reddit.com/r/LocalLLaMA/comments/14lz7j5/ntkaware_scaled_rope_allows_llama_models_to_have/},
  note         = {Accessed: 2026-02-28},
}

@article{ding2024longrope,
  title={Longrope: Extending llm context window beyond 2 million tokens},
  author={Ding, Yiran and Zhang, Li Lyna and Zhang, Chengruidong and Xu, Yuanyuan and Shang, Ning and Xu, Jiahang and Yang, Fan and Yang, Mao},
  journal={arXiv preprint arXiv:2402.13753},
  year={2024}
}

@inproceedings{yu2025comrope,
  title={ComRoPE: Scalable and Robust Rotary Position Embedding Parameterized by Trainable Commuting Angle Matrices},
  author={Yu, Hao and Jiang, Tangyu and Jia, Shuning and Yan, Shannan and Liu, Shunning and Qian, Haolong and Li, Guanghao and Dong, Shuting and Yuan, Chun},
  booktitle={Proceedings of the Computer Vision and Pattern Recognition Conference},
  pages={4508--4517},
  year={2025}
}

@article{ostmeier2024liere,
  title={Liere: Generalizing rotary position encodings},
  author={Ostmeier, Sophie and Axelrod, Brian and Moseley, Michael E and Chaudhari, Akshay and Langlotz, Curtis},
  journal={arXiv preprint arXiv:2406.10322},
  volume={2},
  number={4},
  year={2024}
}

@article{press2021train,
  title={Train short, test long: Attention with linear biases enables input length extrapolation},
  author={Press, Ofir and Smith, Noah A and Lewis, Mike},
  journal={arXiv preprint arXiv:2108.12409},
  year={2021}
}

@inproceedings{sun2023length,
  title={A length-extrapolatable transformer},
  author={Sun, Yutao and Dong, Li and Patra, Barun and Ma, Shuming and Huang, Shaohan and Benhaim, Alon and Chaudhary, Vishrav and Song, Xia and Wei, Furu},
  booktitle={Proceedings of the 61st annual meeting of the association for computational linguistics (volume 1: long papers)},
  pages={14590--14604},
  year={2023}
}

@article{chen2024rotary,
  title={What rotary position embedding can tell us: Identifying query and key weights corresponding to basic syntactic or high-level semantic information},
  author={Chen, Yiting and Yan, Junchi},
  journal={Advances in Neural Information Processing Systems},
  volume={37},
  pages={54507--54528},
  year={2024}
}

@article{zhao2025riflex,
  title={Riflex: A free lunch for length extrapolation in video diffusion transformers},
  author={Zhao, Min and He, Guande and Chen, Yixiao and Zhu, Hongzhou and Li, Chongxuan and Zhu, Jun},
  journal={arXiv preprint arXiv:2502.15894},
  year={2025}
}

@article{ho2022cascaded,
  title={Cascaded diffusion models for high fidelity image generation},
  author={Ho, Jonathan and Saharia, Chitwan and Chan, William and Fleet, David J and Norouzi, Mohammad and Salimans, Tim},
  journal={Journal of Machine Learning Research},
  volume={23},
  number={47},
  pages={1--33},
  year={2022}
}

@article{pernias2023wurstchen,
  title={W{\"u}rstchen: An efficient architecture for large-scale text-to-image diffusion models},
  author={Pernias, Pablo and Rampas, Dominic and Richter, Mats L and Pal, Christopher J and Aubreville, Marc},
  journal={arXiv preprint arXiv:2306.00637},
  year={2023}
}

@article{saharia2022photorealistic,
  title={Photorealistic text-to-image diffusion models with deep language understanding},
  author={Saharia, Chitwan and Chan, William and Saxena, Saurabh and Li, Lala and Whang, Jay and Denton, Emily L and Ghasemipour, Kamyar and Gontijo Lopes, Raphael and Karagol Ayan, Burcu and Salimans, Tim and others},
  journal={Advances in neural information processing systems},
  volume={35},
  pages={36479--36494},
  year={2022}
}

@article{teng2023relay,
  title={Relay diffusion: Unifying diffusion process across resolutions for image synthesis},
  author={Teng, Jiayan and Zheng, Wendi and Ding, Ming and Hong, Wenyi and Wangni, Jianqiao and Yang, Zhuoyi and Tang, Jie},
  journal={arXiv preprint arXiv:2309.03350},
  year={2023}
}

@article{chen2024edt,
  title={Edt: An efficient diffusion transformer framework inspired by human-like sketching},
  author={Chen, Xinwang and Liu, Ning and Zhu, Yichen and Feng, Feifei and Tang, Jian},
  journal={Advances in Neural Information Processing Systems},
  volume={37},
  pages={134075--134106},
  year={2024}
}

@article{jin2024pyramidal,
  title={Pyramidal flow matching for efficient video generative modeling},
  author={Jin, Yang and Sun, Zhicheng and Li, Ningyuan and Xu, Kun and Jiang, Hao and Zhuang, Nan and Huang, Quzhe and Song, Yang and Mu, Yadong and Lin, Zhouchen},
  journal={arXiv preprint arXiv:2410.05954},
  year={2024}
}

@inproceedings{rombach2022high,
  title={High-resolution image synthesis with latent diffusion models},
  author={Rombach, Robin and Blattmann, Andreas and Lorenz, Dominik and Esser, Patrick and Ommer, Bj{\"o}rn},
  booktitle={Proceedings of the IEEE/CVF conference on computer vision and pattern recognition},
  pages={10684--10695},
  year={2022}
}

@article{podell2023sdxl,
  title={Sdxl: Improving latent diffusion models for high-resolution image synthesis},
  author={Podell, Dustin and English, Zion and Lacey, Kyle and Blattmann, Andreas and Dockhorn, Tim and M{\"u}ller, Jonas and Penna, Joe and Rombach, Robin},
  journal={arXiv preprint arXiv:2307.01952},
  year={2023}
}

@article{jeong2025upsample,
  title={Upsample What Matters: Region-Adaptive Latent Sampling for Accelerated Diffusion Transformers},
  author={Jeong, Wongi and Lee, Kyungryeol and Seo, Hoigi and Chun, Se Young},
  journal={arXiv preprint arXiv:2507.08422},
  year={2025}
}

@article{tian2025training,
  title={Training-free diffusion acceleration with bottleneck sampling},
  author={Tian, Ye and Xia, Xin and Ren, Yuxi and Lin, Shanchuan and Wang, Xing and Xiao, Xuefeng and Tong, Yunhai and Yang, Ling and Cui, Bin},
  journal={arXiv preprint arXiv:2503.18940},
  year={2025}
}

@inproceedings{jeong2025latent,
  title={Latent space super-resolution for higher-resolution image generation with diffusion models},
  author={Jeong, Jinho and Han, Sangmin and Kim, Jinwoo and Kim, Seon Joo},
  booktitle={Proceedings of the Computer Vision and Pattern Recognition Conference},
  pages={2355--2365},
  year={2025}
}

@inproceedings{ding2024patched,
  title={Patched Denoising Diffusion Models For High-Resolution Image Synthesis},
  author={Zheng Ding and Mengqi Zhang and Jiajun Wu and Zhuowen Tu},
  booktitle={The Twelfth International Conference on Learning Representations},
  year={2024}
}

@article{liu2025region,
  title={Region-adaptive sampling for diffusion transformers},
  author={Liu, Ziming and Yang, Yifan and Zhang, Chengruidong and Zhang, Yiqi and Qiu, Lili and You, Yang and Yang, Yuqing},
  journal={arXiv preprint arXiv:2502.10389},
  year={2025}
}

@inproceedings{du2025fewer,
  title={Fewer Denoising Steps or Cheaper Per-Step Inference: Towards Compute-Optimal Diffusion Model Deployment},
  author={Du, Zhenbang and Fu, Yonggan and Wang, Lifu and Qian, Jiayi and Luo, Xiao and Lin, Yingyan Celine},
  booktitle={Proceedings of the IEEE/CVF International Conference on Computer Vision},
  pages={3001--3010},
  year={2025}
}

@inproceedings{heo2024rotary,
  title={Rotary position embedding for vision transformer},
  author={Heo, Byeongho and Park, Song and Han, Dongyoon and Yun, Sangdoo},
  booktitle={European Conference on Computer Vision},
  pages={289--305},
  year={2024},
  organization={Springer}
}

@inproceedings{bolya2023token,
  title={Token merging for fast stable diffusion},
  author={Bolya, Daniel and Hoffman, Judy},
  booktitle={Proceedings of the IEEE/CVF conference on computer vision and pattern recognition},
  pages={4599--4603},
  year={2023}
}

@inproceedings{shang2023post,
  title={Post-training quantization on diffusion models},
  author={Shang, Yuzhang and Yuan, Zhihang and Xie, Bin and Wu, Bingzhe and Yan, Yan},
  booktitle={Proceedings of the IEEE/CVF conference on computer vision and pattern recognition},
  pages={1972--1981},
  year={2023}
}

@article{li2024svdquant,
  title={Svdquant: Absorbing outliers by low-rank components for 4-bit diffusion models},
  author={Li, Muyang and Lin, Yujun and Zhang, Zhekai and Cai, Tianle and Li, Xiuyu and Guo, Junxian and Xie, Enze and Meng, Chenlin and Zhu, Jun-Yan and Han, Song},
  journal={arXiv preprint arXiv:2411.05007},
  year={2024}
}

@inproceedings{chen2025q,
  title={Q-dit: Accurate post-training quantization for diffusion transformers},
  author={Chen, Lei and Meng, Yuan and Tang, Chen and Ma, Xinzhu and Jiang, Jingyan and Wang, Xin and Wang, Zhi and Zhu, Wenwu},
  booktitle={Proceedings of the Computer Vision and Pattern Recognition Conference},
  pages={28306--28315},
  year={2025}
}

@inproceedings{deng2025vq4dit,
  title={Vq4dit: Efficient post-training vector quantization for diffusion transformers},
  author={Deng, Juncan and Li, Shuaiting and Wang, Zeyu and Gu, Hong and Xu, Kedong and Huang, Kejie},
  booktitle={Proceedings of the AAAI Conference on Artificial Intelligence},
  volume={39},
  pages={16226--16234},
  year={2025}
}

@article{li2023snapfusion,
  title={Snapfusion: Text-to-image diffusion model on mobile devices within two seconds},
  author={Li, Yanyu and Wang, Huan and Jin, Qing and Hu, Ju and Chemerys, Pavlo and Fu, Yun and Wang, Yanzhi and Tulyakov, Sergey and Ren, Jian},
  journal={Advances in Neural Information Processing Systems},
  volume={36},
  pages={20662--20678},
  year={2023}
}

@inproceedings{feng2024relational,
  title={Relational diffusion distillation for efficient image generation},
  author={Feng, Weilun and Yang, Chuanguang and An, Zhulin and Huang, Libo and Diao, Boyu and Wang, Fei and Xu, Yongjun},
  booktitle={Proceedings of the 32nd ACM international conference on multimedia},
  pages={205--213},
  year={2024}
}

@article{zhang2024accelerating,
  title={Accelerating diffusion models with one-to-many knowledge distillation},
  author={Zhang, Linfeng and Ma, Kaisheng},
  journal={arXiv preprint arXiv:2410.04191},
  year={2024}
}

@article{xie2025sana,
  title={Sana 1.5: Efficient scaling of training-time and inference-time compute in linear diffusion transformer},
  author={Xie, Enze and Chen, Junsong and Zhao, Yuyang and Yu, Jincheng and Zhu, Ligeng and Wu, Chengyue and Lin, Yujun and Zhang, Zhekai and Li, Muyang and Chen, Junyu and others},
  journal={arXiv preprint arXiv:2501.18427},
  year={2025}
}

@inproceedings{fang2025tinyfusion,
  title={Tinyfusion: Diffusion transformers learned shallow},
  author={Fang, Gongfan and Li, Kunjun and Ma, Xinyin and Wang, Xinchao},
  booktitle={Proceedings of the Computer Vision and Pattern Recognition Conference},
  pages={18144--18154},
  year={2025}
}

@article{ma2024learning,
  title={Learning-to-cache: Accelerating diffusion transformer via layer caching},
  author={Ma, Xinyin and Fang, Gongfan and Bi Mi, Michael and Wang, Xinchao},
  journal={Advances in Neural Information Processing Systems},
  volume={37},
  pages={133282--133304},
  year={2024}
}

@article{seo2025skrr,
  title={Skrr: Skip and Re-use Text Encoder Layers for Memory Efficient Text-to-Image Generation},
  author={Seo, Hoigi and Jeong, Wongi and Seo, Jae-sun and Chun, Se Young},
  journal={arXiv preprint arXiv:2502.08690},
  year={2025}
}

@inproceedings{zhang2025training,
  title={Training-free and hardware-friendly acceleration for diffusion models via similarity-based token pruning},
  author={Zhang, Evelyn and Tang, Jiayi and Ning, Xuefei and Zhang, Linfeng},
  booktitle={Proceedings of the AAAI Conference on Artificial Intelligence},
  volume={39},
  pages={9878--9886},
  year={2025}
}

@article{zhang2024token,
  title={Token pruning for caching better: 9 times acceleration on stable diffusion for free},
  author={Zhang, Evelyn and Xiao, Bang and Tang, Jiayi and Ma, Qianli and Zou, Chang and Ning, Xuefei and Hu, Xuming and Zhang, Linfeng},
  journal={arXiv preprint arXiv:2501.00375},
  year={2024}
}

@inproceedings{you2025layer,
  title={Layer-and Timestep-Adaptive Differentiable Token Compression Ratios for Efficient Diffusion Transformers},
  author={You, Haoran and Barnes, Connelly and Zhou, Yuqian and Kang, Yan and Du, Zhenbang and Zhou, Wei and Zhang, Lingzhi and Nitzan, Yotam and Liu, Xiaoyang and Lin, Zhe and others},
  booktitle={Proceedings of the Computer Vision and Pattern Recognition Conference},
  pages={18072--18082},
  year={2025}
}

@inproceedings{changsparsedit,
  title={SparseDiT: Token Sparsification for Efficient Diffusion Transformer},
  author={Chang, Shuning and Pichao, WANG and Tang, Jiasheng and Wang, Fan and Yang, Yi},
  booktitle={The Thirty-ninth Annual Conference on Neural Information Processing Systems},
  year={2025}
}

@article{chen2025sparse,
  title={Sparse-vDiT: Unleashing the Power of Sparse Attention to Accelerate Video Diffusion Transformers},
  author={Chen, Pengtao and Zeng, Xianfang and Zhao, Maosen and Ye, Peng and Shen, Mingzhu and Cheng, Wei and Yu, Gang and Chen, Tao},
  journal={arXiv preprint arXiv:2506.03065},
  year={2025}
}

@inproceedings{fang2023structural,
  title={Structural pruning for diffusion models},
  author={Gongfan Fang and Xinyin Ma and Xinchao Wang},
  booktitle={Advances in Neural Information Processing Systems},
  year={2023},
}

@inproceedings{castells2024ld,
  title={Ld-pruner: Efficient pruning of latent diffusion models using task-agnostic insights},
  author={Castells, Thibault and Song, Hyoung-Kyu and Kim, Bo-Kyeong and Choi, Shinkook},
  booktitle={Proceedings of the IEEE/CVF Conference on Computer Vision and Pattern Recognition},
  pages={821--830},
  year={2024}
}

@article{zhang2024effortless,
  title={Effortless efficiency: Low-cost pruning of diffusion models},
  author={Zhang, Yang and Jin, Er and Dong, Yanfei and Khakzar, Ashkan and Torr, Philip and Stegmaier, Johannes and Kawaguchi, Kenji},
  journal={arXiv preprint arXiv:2412.02852},
  year={2024}
}

@inproceedings{wan2025pruning,
  title={Pruning for Sparse Diffusion Models Based on Gradient Flow},
  author={Wan, Ben and Zheng, Tianyi and Chen, Zhaoyu and Wang, Yuxiao and Wang, Jia},
  booktitle={ICASSP 2025-2025 IEEE International Conference on Acoustics, Speech and Signal Processing (ICASSP)},
  pages={1--5},
  year={2025},
  organization={IEEE}
}

@article{ganjdanesh2024not,
  title={Not all prompts are made equal: Prompt-based pruning of text-to-image diffusion models},
  author={Ganjdanesh, Alireza and Shirkavand, Reza and Gao, Shangqian and Huang, Heng},
  journal={arXiv preprint arXiv:2406.12042},
  year={2024}
}

@inproceedings{wimbauer2024cache,
  title={Cache me if you can: Accelerating diffusion models through block caching},
  author={Wimbauer, Felix and Wu, Bichen and Schoenfeld, Edgar and Dai, Xiaoliang and Hou, Ji and He, Zijian and Sanakoyeu, Artsiom and Zhang, Peizhao and Tsai, Sam and Kohler, Jonas and others},
  booktitle={Proceedings of the IEEE/CVF Conference on Computer Vision and Pattern Recognition},
  pages={6211--6220},
  year={2024}
}

@article{lv2024fastercache,
  title={Fastercache: Training-free video diffusion model acceleration with high quality},
  author={Lv, Zhengyao and Si, Chenyang and Song, Junhao and Yang, Zhenyu and Qiao, Yu and Liu, Ziwei and Wong, Kwan-Yee K},
  journal={arXiv preprint arXiv:2410.19355},
  year={2024}
}

@article{chen2024delta,
  title={A Training-Free Acceleration Method Tailored for Diffusion Transformers},
  author={Chen, Pengtao and Shen, Mingzhu and Ye, Peng and Cao, Jianjian and Tu, Chongjun and Bouganis, Christos-Savvas and Zhao, Yiren and Chen, Tao},
  journal={arXiv preprint arXiv:2406.01125},
  year={2024}
}

@inproceedings{ma2024deepcache,
  title={Deepcache: Accelerating diffusion models for free},
  author={Ma, Xinyin and Fang, Gongfan and Wang, Xinchao},
  booktitle={Proceedings of the IEEE/CVF conference on computer vision and pattern recognition},
  pages={15762--15772},
  year={2024}
}

@article{zou2024accelerating,
  title={Accelerating diffusion transformers with token-wise feature caching},
  author={Zou, Chang and Liu, Xuyang and Liu, Ting and Huang, Siteng and Zhang, Linfeng},
  journal={arXiv preprint arXiv:2410.05317},
  year={2024}
}

@inproceedings{kahatapitiya2025adaptive,
  title={Adaptive caching for faster video generation with diffusion transformers},
  author={Kahatapitiya, Kumara and Liu, Haozhe and He, Sen and Liu, Ding and Jia, Menglin and Zhang, Chenyang and Ryoo, Michael S and Xie, Tian},
  booktitle={Proceedings of the IEEE/CVF International Conference on Computer Vision},
  pages={15240--15252},
  year={2025}
}

@inproceedings{liu2025timestep,
  title={Timestep Embedding Tells: It's Time to Cache for Video Diffusion Model},
  author={Liu, Feng and Zhang, Shiwei and Wang, Xiaofeng and Wei, Yujie and Qiu, Haonan and Zhao, Yuzhong and Zhang, Yingya and Ye, Qixiang and Wan, Fang},
  booktitle={Proceedings of the Computer Vision and Pattern Recognition Conference},
  pages={7353--7363},
  year={2025}
}

@article{lou2024token,
  title={Token caching for diffusion transformer acceleration},
  author={Lou, Jinming and Luo, Wenyang and Liu, Yufan and Li, Bing and Ding, Xinmiao and Hu, Weiming and Cao, Jiajiong and Li, Yuming and Ma, Chenguang},
  journal={arXiv preprint arXiv:2409.18523},
  year={2024}
}

@inproceedings{agarwal2024approximate,
  title={Approximate caching for efficiently serving $\{$Text-to-Image$\}$ diffusion models},
  author={Agarwal, Shubham and Mitra, Subrata and Chakraborty, Sarthak and Karanam, Srikrishna and Mukherjee, Koyel and Saini, Shiv Kumar},
  booktitle={21st USENIX Symposium on Networked Systems Design and Implementation (NSDI 24)},
  pages={1173--1189},
  year={2024}
}

@article{salimans2022progressive,
  title={Progressive distillation for fast sampling of diffusion models},
  author={Salimans, Tim and Ho, Jonathan},
  journal={arXiv preprint arXiv:2202.00512},
  year={2022}
}

@article{song2023consistency,
  title={Consistency Models},
  author={Song, Yang and Dhariwal, Prafulla and Chen, Mark and Sutskever, Ilya},
  journal={arXiv preprint arXiv:2303.01469},
  year={2023},
}

@article{luo2023latent,
  title={Latent consistency models: Synthesizing high-resolution images with few-step inference},
  author={Luo, Simian and Tan, Yiqin and Huang, Longbo and Li, Jian and Zhao, Hang},
  journal={arXiv preprint arXiv:2310.04378},
  year={2023}
}

@article{zhao2023unipc,
  title={UniPC: A Unified Predictor-Corrector Framework for Fast Sampling of Diffusion Models},
  author={Zhao, Wenliang and Bai, Lujia and Rao, Yongming and Zhou, Jie and Lu, Jiwen},
  journal={NeurIPS},
  year={2023}
}

@article{song2020denoising,
  title={Denoising diffusion implicit models},
  author={Song, Jiaming and Meng, Chenlin and Ermon, Stefano},
  journal={arXiv preprint arXiv:2010.02502},
  year={2020}
}

@article{lu2022dpm,
  title={Dpm-solver: A fast ode solver for diffusion probabilistic model sampling in around 10 steps},
  author={Lu, Cheng and Zhou, Yuhao and Bao, Fan and Chen, Jianfei and Li, Chongxuan and Zhu, Jun},
  journal={Advances in neural information processing systems},
  volume={35},
  pages={5775--5787},
  year={2022}
}

@article{lu2025dpm,
  title={Dpm-solver++: Fast solver for guided sampling of diffusion probabilistic models},
  author={Lu, Cheng and Zhou, Yuhao and Bao, Fan and Chen, Jianfei and Li, Chongxuan and Zhu, Jun},
  journal={Machine Intelligence Research},
  pages={1--22},
  year={2025},
  publisher={Springer}
}

@article{zheng2023dpm,
  title={Dpm-solver-v3: Improved diffusion ode solver with empirical model statistics},
  author={Zheng, Kaiwen and Lu, Cheng and Chen, Jianfei and Zhu, Jun},
  journal={Advances in Neural Information Processing Systems},
  volume={36},
  pages={55502--55542},
  year={2023}
}

@inproceedings{huang2024vbench,
  title={Vbench: Comprehensive benchmark suite for video generative models},
  author={Huang, Ziqi and He, Yinan and Yu, Jiashuo and Zhang, Fan and Si, Chenyang and Jiang, Yuming and Zhang, Yuanhan and Wu, Tianxing and Jin, Qingyang and Chanpaisit, Nattapol and others},
  booktitle={Proceedings of the IEEE/CVF Conference on Computer Vision and Pattern Recognition},
  pages={21807--21818},
  year={2024}
}

@inproceedings{wu2023dover,
      title={Exploring Video Quality Assessment on User Generated Contents from Aesthetic and Technical Perspectives}, 
      author={Wu, Haoning and Zhang, Erli and Liao, Liang and Chen, Chaofeng and Hou, Jingwen Hou and Wang, Annan and Sun, Wenxiu Sun and Yan, Qiong and Lin, Weisi},
      year={2023},
      booktitle={International Conference on Computer Vision (ICCV)},
}

@inproceedings{ke2021musiq,
  title={Musiq: Multi-scale image quality transformer},
  author={Ke, Junjie and Wang, Qifei and Wang, Yilin and Milanfar, Peyman and Yang, Feng},
  booktitle={Proceedings of the IEEE/CVF international conference on computer vision},
  pages={5148--5157},
  year={2021}
}

@inproceedings{wang2022exploring,
    author = {Wang, Jianyi and Chan, Kelvin CK and Loy, Chen Change},
    title = {Exploring CLIP for Assessing the Look and Feel of Images},
    booktitle = {AAAI},
    year = {2023}
}

@inproceedings{radford2021learning,
  title={Learning transferable visual models from natural language supervision},
  author={Radford, Alec and Kim, Jong Wook and Hallacy, Chris and Ramesh, Aditya and Goh, Gabriel and Agarwal, Sandhini and Sastry, Girish and Askell, Amanda and Mishkin, Pamela and Clark, Jack and others},
  booktitle={International conference on machine learning},
  pages={8748--8763},
  year={2021},
  organization={PmLR}
}

@inproceedings{linardos2021deepgaze,
  title={DeepGaze IIE: Calibrated prediction in and out-of-domain for state-of-the-art saliency modeling},
  author={Linardos, Akis and K{\"u}mmerer, Matthias and Press, Ori and Bethge, Matthias},
  booktitle={Proceedings of the IEEE/CVF International Conference on Computer Vision},
  pages={12919--12928},
  year={2021}
}

@misc {BoerBohan2025TAEHV,
  author = {Boer Bohan, Ollin},
  title = {TAEHV: Tiny AutoEncoder for Hunyuan Video},
  year = {2025},
  howpublished = {\url{https://github.com/madebyollin/taehv}},
  note         = {Accessed: 2026-02-28},
}

@article{itti2002model,
  title={A model of saliency-based visual attention for rapid scene analysis},
  author={Itti, Laurent and Koch, Christof and Niebur, Ernst},
  journal={IEEE Transactions on pattern analysis and machine intelligence},
  volume={20},
  number={11},
  pages={1254--1259},
  year={2002},
  publisher={Ieee}
}

@article{kienzle2006nonparametric,
  title={A nonparametric approach to bottom-up visual saliency},
  author={Kienzle, Wolf and Wichmann, Felix A and Franz, Matthias and Sch{\"o}lkopf, Bernhard},
  journal={Advances in neural information processing systems},
  volume={19},
  year={2006}
}

@article{zhang2008sun,
  title={SUN: A Bayesian framework for saliency using natural statistics},
  author={Zhang, Lingyun and Tong, Matthew H and Marks, Tim K and Shan, Honghao and Cottrell, Garrison W},
  journal={Journal of vision},
  volume={8},
  number={7},
  pages={32--32},
  year={2008},
  publisher={The Association for Research in Vision and Ophthalmology}
}

@inproceedings{NIPS2006_gbvs,
 author = {Harel, Jonathan and Koch, Christof and Perona, Pietro},
 booktitle = {Advances in Neural Information Processing Systems},
 editor = {B. Sch\"{o}lkopf and J. Platt and T. Hoffman},
 pages = {},
 publisher = {MIT Press},
 title = {Graph-Based Visual Saliency},
 url = {https://proceedings.neurips.cc/paper_files/paper/2006/file/4db0f8b0fc895da263fd77fc8aecabe4-Paper.pdf},
 volume = {19},
 year = {2006},
 note         = {Accessed: 2026-02-28},
}

@inproceedings{jiang2015salicon,
  title={Salicon: Saliency in context},
  author={Jiang, Ming and Huang, Shengsheng and Duan, Juanyong and Zhao, Qi},
  booktitle={Proceedings of the IEEE conference on computer vision and pattern recognition},
  pages={1072--1080},
  year={2015}
}

@article{deepgazei,
  title={Deep gaze i: Boosting saliency prediction with feature maps trained on imagenet},
  author={K{\"u}mmerer, Matthias and Theis, Lucas and Bethge, Matthias},
  journal={arXiv preprint arXiv:1411.1045},
  year={2014}
}

@article{alexnet,
  title={Imagenet classification with deep convolutional neural networks},
  author={Krizhevsky, Alex and Sutskever, Ilya and Hinton, Geoffrey E},
  journal={Advances in neural information processing systems},
  volume={25},
  year={2012}
}

@article{vggnet,
  title={Very deep convolutional networks for large-scale image recognition},
  author={Simonyan, Karen and Zisserman, Andrew},
  journal={arXiv preprint arXiv:1409.1556},
  year={2014}
}

@article{wang2017deep,
  title={Deep visual attention prediction},
  author={Wang, Wenguan and Shen, Jianbing},
  journal={IEEE Transactions on Image Processing},
  volume={27},
  number={5},
  pages={2368--2378},
  year={2017},
  publisher={IEEE}
}

@inproceedings{cornia2016deep,
  title={A deep multi-level network for saliency prediction},
  author={Cornia, Marcella and Baraldi, Lorenzo and Serra, Giuseppe and Cucchiara, Rita},
  booktitle={2016 23rd International Conference on Pattern Recognition (ICPR)},
  pages={3488--3493},
  year={2016},
  organization={IEEE}
}

@article{kruthiventi2017deepfix,
  title={Deepfix: A fully convolutional neural network for predicting human eye fixations},
  author={Kruthiventi, Srinivas SS and Ayush, Kumar and Babu, R Venkatesh},
  journal={IEEE Transactions on Image Processing},
  volume={26},
  number={9},
  pages={4446--4456},
  year={2017},
  publisher={IEEE}
}

@article{cornia2018predicting,
  title={Predicting human eye fixations via an lstm-based saliency attentive model},
  author={Cornia, Marcella and Baraldi, Lorenzo and Serra, Giuseppe and Cucchiara, Rita},
  journal={IEEE Transactions on Image Processing},
  volume={27},
  number={10},
  pages={5142--5154},
  year={2018},
  publisher={IEEE}
}

@article{liu2018deep,
  title={A deep spatial contextual long-term recurrent convolutional network for saliency detection},
  author={Liu, Nian and Han, Junwei},
  journal={IEEE Transactions on Image Processing},
  volume={27},
  number={7},
  pages={3264--3274},
  year={2018},
  publisher={IEEE}
}

@article{wang2019revisiting,
  title={Revisiting video saliency prediction in the deep learning era},
  author={Wang, Wenguan and Shen, Jianbing and Xie, Jianwen and Cheng, Ming-Ming and Ling, Haibin and Borji, Ali},
  journal={IEEE transactions on pattern analysis and machine intelligence},
  volume={43},
  number={1},
  pages={220--237},
  year={2019},
  publisher={IEEE}
}

@article{lou2022transalnet,
  title={TranSalNet: Towards perceptually relevant visual saliency prediction},
  author={Lou, Jianxun and Lin, Hanhe and Marshall, David and Saupe, Dietmar and Liu, Hantao},
  journal={Neurocomputing},
  volume={494},
  pages={455--467},
  year={2022},
  publisher={Elsevier}
}

@article{vaswani2017attention,
  title={Attention is all you need},
  author={Vaswani, Ashish and Shazeer, Noam and Parmar, Niki and Uszkoreit, Jakob and Jones, Llion and Gomez, Aidan N and Kaiser, {\L}ukasz and Polosukhin, Illia},
  journal={Advances in neural information processing systems},
  volume={30},
  year={2017}
}

@article{vit,
  title={An image is worth 16x16 words: Transformers for image recognition at scale},
  author={Dosovitskiy, Alexey},
  journal={arXiv preprint arXiv:2010.11929},
  year={2020}
}

@article{zhou2023transformer,
  title={Transformer-based multi-scale feature integration network for video saliency prediction},
  author={Zhou, Xiaofei and Wu, Songhe and Shi, Ran and Zheng, Bolun and Wang, Shuai and Yin, Haibing and Zhang, Jiyong and Yan, Chenggang},
  journal={IEEE Transactions on Circuits and Systems for Video Technology},
  volume={33},
  number={12},
  pages={7696--7707},
  year={2023},
  publisher={IEEE}
}

@article{ma2022video,
  title={Video saliency forecasting transformer},
  author={Ma, Cheng and Sun, Haowen and Rao, Yongming and Zhou, Jie and Lu, Jiwen},
  journal={IEEE transactions on circuits and systems for video technology},
  volume={32},
  number={10},
  pages={6850--6862},
  year={2022},
  publisher={IEEE}
}

@article{jia2020eml,
  title={Eml-net: An expandable multi-layer network for saliency prediction},
  author={Jia, Sen and Bruce, Neil DB},
  journal={Image and vision computing},
  volume={95},
  pages={103887},
  year={2020},
  publisher={Elsevier}
}

@article{ma2025magcache,
  title={MagCache: Fast Video Generation with Magnitude-Aware Cache},
  author={Ma, Zehong and Wei, Longhui and Wang, Feng and Zhang, Shiliang and Tian, Qi},
  journal={arXiv preprint arXiv:2506.09045},
  year={2025}
}

@inproceedings{lin2014microsoft,
  title={Microsoft coco: Common objects in context},
  author={Lin, Tsung-Yi and Maire, Michael and Belongie, Serge and Hays, James and Perona, Pietro and Ramanan, Deva and Doll{\'a}r, Piotr and Zitnick, C Lawrence},
  booktitle={Computer Vision--ECCV 2014: 13th European Conference, Zurich, Switzerland, September 6-12, 2014, Proceedings, Part V 13},
  pages={740--755},
  year={2014},
  organization={Springer}
}

@article{qiu2025cinescale,
  title={CineScale: Free Lunch in High-Resolution Cinematic Visual Generation},
  author={Qiu, Haonan and Yu, Ning and Huang, Ziqi and Debevec, Paul and Liu, Ziwei},
  journal={arXiv preprint arXiv:2508.15774},
  year={2025}
}

@inproceedings{qiu2025freescale,
  title={Freescale: Unleashing the resolution of diffusion models via tuning-free scale fusion},
  author={Qiu, Haonan and Zhang, Shiwei and Wei, Yujie and Chu, Ruihang and Yuan, Hangjie and Wang, Xiang and Zhang, Yingya and Liu, Ziwei},
  booktitle={Proceedings of the IEEE/CVF International Conference on Computer Vision},
  pages={16893--16903},
  year={2025}
}

@article{touvron2023llama,
  title={Llama: Open and efficient foundation language models},
  author={Touvron, Hugo and Lavril, Thibaut and Izacard, Gautier and Martinet, Xavier and Lachaux, Marie-Anne and Lacroix, Timoth{\'e}e and Rozi{\`e}re, Baptiste and Goyal, Naman and Hambro, Eric and Azhar, Faisal and others},
  journal={arXiv preprint arXiv:2302.13971},
  year={2023}
}

@article{guo2025deepseek,
  title={Deepseek-r1: Incentivizing reasoning capability in llms via reinforcement learning},
  author={Guo, Daya and Yang, Dejian and Zhang, Haowei and Song, Junxiao and Zhang, Ruoyu and Xu, Runxin and Zhu, Qihao and Ma, Shirong and Wang, Peiyi and Bi, Xiao and others},
  journal={arXiv preprint arXiv:2501.12948},
  year={2025}
}

@inproceedings{feng2025romantex,
  title={Romantex: Decoupling 3d-aware rotary positional embedded multi-attention network for texture synthesis},
  author={Feng, Yifei and Yang, Mingxin and Yang, Shuhui and Zhang, Sheng and Yu, Jiaao and Zhao, Zibo and Liu, Yuhong and Jiang, Jie and Guo, Chunchao},
  booktitle={Proceedings of the IEEE/CVF International Conference on Computer Vision},
  pages={17203--17213},
  year={2025}
}

@inproceedings{crowson2024scalable,
  title={Scalable high-resolution pixel-space image synthesis with hourglass diffusion transformers},
  author={Crowson, Katherine and Baumann, Stefan Andreas and Birch, Alex and Abraham, Tanishq Mathew and Kaplan, Daniel Z and Shippole, Enrico},
  booktitle={Forty-first International Conference on Machine Learning},
  year={2024}
}

@article{tian2024u,
  title={U-dits: Downsample tokens in u-shaped diffusion transformers},
  author={Tian, Yuchuan and Tu, Zhijun and Chen, Hanting and Hu, Jie and Xu, Chao and Wang, Yunhe},
  journal={Advances in Neural Information Processing Systems},
  volume={37},
  pages={51994--52013},
  year={2024}
}

@inproceedings{black2022gpt,
  title={Gpt-neox-20b: An open-source autoregressive language model},
  author={Black, Sidney and Biderman, Stella and Hallahan, Eric and Anthony, Quentin and Gao, Leo and Golding, Laurence and He, Horace and Leahy, Connor and McDonell, Kyle and Phang, Jason and others},
  booktitle={Proceedings of BigScience Episode\# 5--Workshop on Challenges \& Perspectives in Creating Large Language Models},
  pages={95--136},
  year={2022}
}

@article{bai2023qwen,
  title={Qwen technical report},
  author={Bai, Jinze and Bai, Shuai and Chu, Yunfei and Cui, Zeyu and Dang, Kai and Deng, Xiaodong and Fan, Yang and Ge, Wenbin and Han, Yu and Huang, Fei and others},
  journal={arXiv preprint arXiv:2309.16609},
  year={2023}
}

@article{chowdhery2023palm,
  title={Palm: Scaling language modeling with pathways},
  author={Chowdhery, Aakanksha and Narang, Sharan and Devlin, Jacob and Bosma, Maarten and Mishra, Gaurav and Roberts, Adam and Barham, Paul and Chung, Hyung Won and Sutton, Charles and Gehrmann, Sebastian and others},
  journal={Journal of Machine Learning Research},
  volume={24},
  number={240},
  pages={1--113},
  year={2023}
}

@article{yang2024cogvideox,
  title={Cogvideox: Text-to-video diffusion models with an expert transformer},
  author={Yang, Zhuoyi and Teng, Jiayan and Zheng, Wendi and Ding, Ming and Huang, Shiyu and Xu, Jiazheng and Yang, Yuanming and Hong, Wenyi and Zhang, Xiaohan and Feng, Guanyu and others},
  journal={arXiv preprint arXiv:2408.06072},
  year={2024}
}

@article{dubey2024llama,
  title={The llama 3 herd of models},
  author={Dubey, Abhimanyu and Jauhri, Abhinav and Pandey, Abhinav and Kadian, Abhishek and Al-Dahle, Ahmad and Letman, Aiesha and Mathur, Akhil and Schelten, Alan and Yang, Amy and Fan, Angela and others},
  journal={arXiv e-prints},
  pages={arXiv--2407},
  year={2024}
}

@article{team2024gemma,
  title={Gemma: Open models based on gemini research and technology},
  author={Team, Gemma and Mesnard, Thomas and Hardin, Cassidy and Dadashi, Robert and Bhupatiraju, Surya and Pathak, Shreya and Sifre, Laurent and Rivi{\`e}re, Morgane and Kale, Mihir Sanjay and Love, Juliette and others},
  journal={arXiv preprint arXiv:2403.08295},
  year={2024}
}

@misc{team2023internlm,
  title={Internlm: A multilingual language model with progressively enhanced capabilities},
  author={Team, InternLM},
  year={2023}
}

@inproceedings{yin2024one,
  title={One-step diffusion with distribution matching distillation},
  author={Yin, Tianwei and Gharbi, Micha{\"e}l and Zhang, Richard and Shechtman, Eli and Durand, Fredo and Freeman, William T and Park, Taesung},
  booktitle={Proceedings of the IEEE/CVF conference on computer vision and pattern recognition},
  pages={6613--6623},
  year={2024}
}

@misc{languagebind_open_sora_plan_v1_1_0,
  author       = {{LanguageBind}},
  title        = {Open-Sora-Plan-v1.1.0 Dataset},
  howpublished = {\url{https://huggingface.co/datasets/LanguageBind/Open-Sora-Plan-v1.1.0}},
  note         = {Accessed: 2026-02-28. This dataset package includes content sourced from Pexels.},
  year         = {2024}
}

@misc{pexels_license,
  author       = {{Pexels}},
  title        = {Pexels License},
  howpublished = {\url{https://www.pexels.com/license/}},
  note         = {Accessed: 2026-02-28},
  year         = {2024}
}

@inproceedings{zhang2025diffusion4k,
    title={Diffusion-4K: Ultra-High-Resolution Image Synthesis with Latent Diffusion Models},
    author={Zhang, Jinjin and Huang, Qiuyu and Liu, Junjie and Guo, Xiefan and Huang, Di},
    year={2025},
    booktitle={IEEE/CVF Conference on Computer Vision and Pattern Recognition (CVPR)},
}

@misc{zhang2025ultrahighresolutionimagesynthesis,
    title={Ultra-High-Resolution Image Synthesis: Data, Method and Evaluation},
    author={Zhang, Jinjin and Huang, Qiuyu and Liu, Junjie and Guo, Xiefan and Huang, Di},
    year={2025},
    note={arXiv:2506.01331},
}

@article{issachar2025dype,
  title={DyPE: Dynamic Position Extrapolation for Ultra High Resolution Diffusion},
  author={Issachar, Noam and Yariv, Guy and Benaim, Sagie and Adi, Yossi and Lischinski, Dani and Fattal, Raanan},
  journal={arXiv preprint arXiv:2510.20766},
  year={2025}
}

@article{xu2023imagereward,
  title={Imagereward: Learning and evaluating human preferences for text-to-image generation},
  author={Xu, Jiazheng and Liu, Xiao and Wu, Yuchen and Tong, Yuxuan and Li, Qinkai and Ding, Ming and Tang, Jie and Dong, Yuxiao},
  journal={Advances in Neural Information Processing Systems},
  volume={36},
  pages={15903--15935},
  year={2023}
}

@article{kummerer2014deep,
  title={Deep gaze i: Boosting saliency prediction with feature maps trained on imagenet},
  author={K{\"u}mmerer, Matthias and Theis, Lucas and Bethge, Matthias},
  journal={arXiv preprint arXiv:1411.1045},
  year={2014}
}

@inproceedings{droste2020unified,
  title={Unified image and video saliency modeling},
  author={Droste, Richard and Jiao, Jianbo and Noble, J Alison},
  booktitle={European Conference on Computer Vision},
  pages={419--435},
  year={2020},
  organization={Springer}
}

@misc{blackforestlabs_flux1_schnell_2024,
  author       = {{Black Forest Labs}},
  title        = {{FLUX.1 [schnell]}},
  year         = {2024},
  howpublished = {\url{https://huggingface.co/black-forest-labs/FLUX.1-schnell}},
  note         = {Hugging Face model card, accessed 2026-03-09}
}

@inproceedings{yin2025causvid,
    title={From Slow Bidirectional to Fast Autoregressive Video Diffusion Models},
    author={Yin, Tianwei and Zhang, Qiang and Zhang, Richard and Freeman, William T and Durand, Fredo and Shechtman, Eli and Huang, Xun},
    booktitle={CVPR},
    year={2025}
}

@String(CVPR= {IEEE Conf. Comput. Vis. Pattern Recog.})

@String(ICCV= {Int. Conf. Comput. Vis.})

@String(ECCV= {Eur. Conf. Comput. Vis.})

@String(ICPR = {Int. Conf. Pattern Recog.})

@String(ICLR = {Int. Conf. Learn. Represent.})

@String(AAAI = {AAAI})

@String(CVPRW= {IEEE Conf. Comput. Vis. Pattern Recog. Worksh.})

@String(CVPRW= {CVPRW})

@inproceedings{xu2024sample,
  author = {Jingyi Xu and Hieu Le and Dimitris Samaras},
  title = {Assessing Sample Quality via the Latent Space of Generative Models},
  booktitle = {ECCV},
  year = {2024}
}

@article{li2024controlling,
  author = {Shuangqi Li and C. Liu and T. Zhang and Hieu Le and S. Süsstrunk and M. Salzmann},
  title = {Controlling the Fidelity and Diversity of Deep Generative Models via Pseudo Density},
  journal = {TMLR},
  year = {2024}
}

@article{li2024enhancing,
  title={Enhancing Compositional Text-to-Image Generation with Reliable Random Seeds},
  author={Li, Shuangqi and Hieu Le and Jingyi Xu and Mathieu Salzmann},
  journal={ICLR},
  year={2025}
}

@article{xu2024learning,
  author    = {Jingyi Xu and Hieu Le and Zhixin Shu and Yang Wang and Yi-Hsuan Tsai and Dimitris Samaras},
  title     = {Learning Frame-Wise Emotion Intensity for Audio-Driven Talking-Head Generation},
  journal   = {arXiv},
  year      = {2024},
  eprint    = {2409.19501}
}

\clearpage
{
\centering
\Large
\textbf{Phase-Aligned RoPE for Mixed-Resolution Diffusion Transformer (Supplementary Material)}\\
\vspace{1.0em}
}
\appendix
In \cref{sec:additional_results}, we provide additional results, comparisons, ablations, overhead breakdowns, and analyses. 
In \cref{sec:derivation}, we present the derivation of Equation~6 from the main text. 
\cref{sec:add_experimental_settings} details further experimental settings, while \cref{sec:add_backgrounds} explains the background concepts used throughout the paper. 
\cref{sec:discussion} presents a discussion of limitations.
In \cref{sec:prompts}, we list the prompts used to generate qualitative results.
We also include a supplementary video for comparisons on text-to-video generation.

% \paragraph{List of Figures.}
List of Figures:
\begin{itemize}
    \item \Cref{fig:flux_compare_acc_supp_4x}: Additional visual comparisons with diffusion acceleration methods for image generation on FLUX.1-dev at \(4\times\) speedup.
    \item \Cref{fig:rb_sal_vis}: Illustration of importance-based region selection for upsampling into high-resolution tokens during mixed-resolution inference.
    \item \Cref{fig:other_use}: Mixed-resolution generation with selected regions rendered at ultra-high resolution while the remaining areas stay at lower resolution.
    \item \Cref{fig:rope_only}: RoPE-only $\kappa(\Delta)$ curves on Wan2.1-1.3B compared with the corresponding learned curves from the trained model.
    \item \Cref{fig:flux_delta_curve_and_rope_only}: Learned and RoPE-only $\kappa(\Delta)$ curves on FLUX.1-dev along the text, height, and width axes.
    \item \Cref{fig:fail_cases}: Failure cases of our method.
    \item \Cref{fig:saliency_vis}: Illustration of importance-based region selection for video.
    \item \Cref{fig:flux_compare_acc_supp_6x}: Additional visual comparisons with diffusion acceleration methods for image generation on FLUX.1-dev at \(6\times\) speedup.
    \item \Cref{fig:flux_supp}: Additional visual comparison with RoPE interpolation methods applied to mixed-resolution denoising on FLUX.1-dev.
    \item \Cref{fig:flux_integrate_acc}: Integration of our method with orthogonal diffusion acceleration techniques for image generation on FLUX.1-dev.
    \item \Cref{fig:flux_2k_ours}: 2048$\times$2048 image samples generated by our method, using DyPE for resolution adaptation.
    \item \Cref{fig:rb_more_vis}: Visual comparison with acceleration methods for 2K image generation on FLUX.1-dev at \(4\times\) speedup.
    \item \Cref{fig:rb_kappa}: Additional $\kappa(\Delta)$ curves. 
\end{itemize}
\section{Additional Results and Analyses}
\label{sec:additional_results}

\subsection{Additional Results}
\label{ssec:add_results}

\begin{figure*}[!t]
\centering
\includegraphics[width=1.0\textwidth]{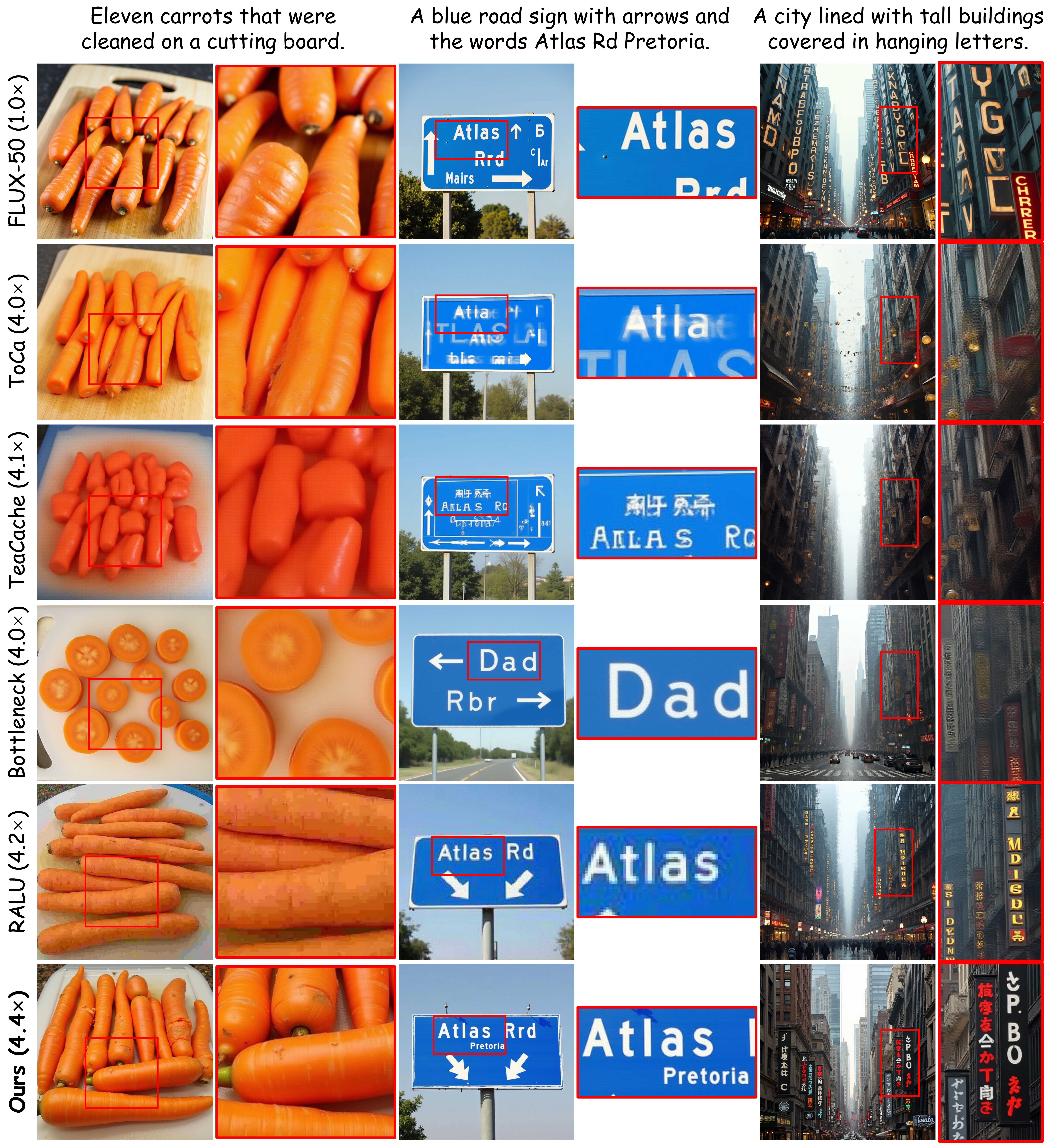}
% \vspace{-6mm}
\caption{
\textbf{Additional visual comparison with diffusion acceleration methods for image generation} on FLUX.1-dev \cite{labs2025flux1kontextflowmatching}. We compare at \(4\times\) speedups, with 50-step FLUX generations as reference.
Red boxes highlight the improved level of detail in our generation. 
}
\label{fig:flux_compare_acc_supp_4x}
% \vspace{-6mm}
\end{figure*}

\noindent\textbf{Additional qualitative results.}
In \cref{fig:flux_compare_acc_supp_4x,fig:flux_compare_acc_supp_6x}, we provide additional visual comparison with diffusion acceleration methods for image generation.
In \cref{fig:flux_supp}, we provide additional qualitative comparison with RoPE interpolation methods for mixed-resolution denoising on FLUX \cite{labs2025flux1kontextflowmatching}.
In the supplementary video, we show text-to-video comparisons with RoPE interpolation methods and acceleration methods on Wan2.1-1.3B \cite{wan2025wan}, along with integration with acceleration techniques and our results on the Wan2.1-14B model.

% fig:rb_sal_vis
\begin{figure}[t]
\centering
\includegraphics[width=1\linewidth]{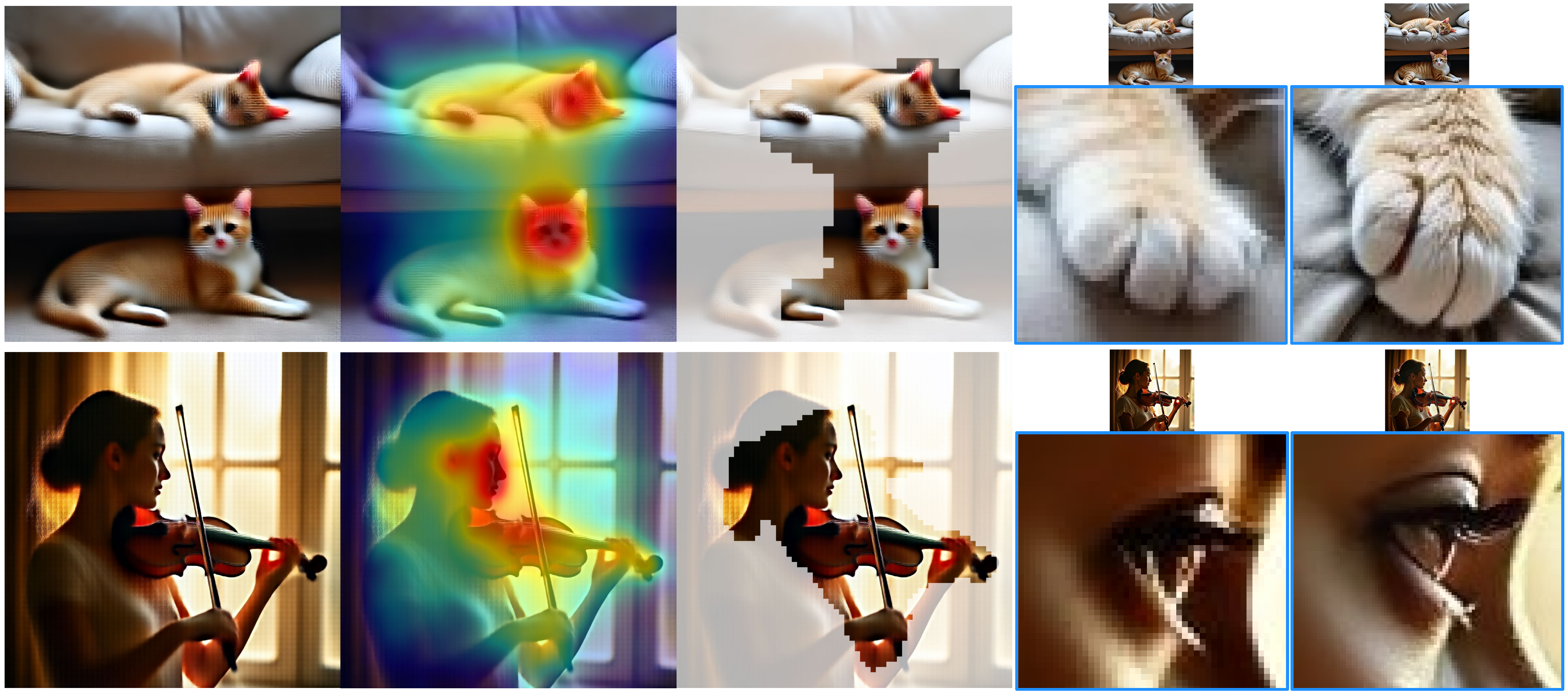}
{
\makebox[0.214\linewidth]{(a) Early LR}%
\makebox[0.214\linewidth]{(b) Saliency}%
\makebox[0.214\linewidth]{(c) HR Mask}%
\makebox[0.358\linewidth]{(d) Full-LR vs. Ours}%
\par} 
\caption{\textbf{Importance-based region selection.} After the early LR denoising stage, we obtain coarse output (a) and infer a saliency map (b). Given the HR token budget, we select the highest-scoring regions as HR mask (c) for mixed-resolution denoising. (d) shows selected regions in the final outputs have better quality than full-LR generation.}
\label{fig:rb_sal_vis} 
\end{figure}

% tab:flux_integrate_acc
\begin{table}[!t]
\centering 
\caption{\textbf{Integration with orthogonal diffusion acceleration methods} on FLUX.1-dev \cite{labs2025flux1kontextflowmatching}. We combine our method with feature caching methods and with the step-distillation model, FLUX.1-schnell \cite{blackforestlabs_flux1_schnell_2024}. Speedups are reported relative to FLUX.1-dev for image generation.}
\label{tab:flux_integrate_acc}
% \vspace{-3mm}
\begingroup
% \small 
% \setlength{\tabcolsep}{3pt}
\begin{adjustbox}{max width=\columnwidth}
\begin{tabular}{l cccc cc}
\toprule
\textbf{Method} & \textbf{ImgR.} $\uparrow$ & \textbf{CLIP-IQA} $\uparrow$ & \textbf{MUSIQ} $\uparrow$ & \textbf{CLIP} $\uparrow$ & \textbf{Time(s)} $\downarrow$ & \textbf{Speed} \\
\midrule
Ours & 0.978 & 0.623 & 71.81 & 31.31 & 2.44 & $4.8\times$ \\
\rowcolor{grayline} \quad + TeaCache \cite{liu2025timestep} & 0.911 & 0.591 & 70.72 & 31.30 & 1.75 & $6.5\times$ \\
\rowcolor{grayline} \quad + MagCache \cite{ma2025magcache} & 0.913 & 0.577 & 70.42 & 31.34 & 1.60 & $7.2\times$ \\
\midrule
FLUX.1-schnell (4-step) \cite{blackforestlabs_flux1_schnell_2024} & 0.990 & 0.637 & 70.48 & 31.12 & 0.93 & $12.4\times$ \\
\rowcolor{grayline} \quad + Ours & 1.110 & 0.611 & 72.85 & 32.03 & 0.78 & $14.8\times$ \\
\bottomrule
\end{tabular}
\end{adjustbox}
\endgroup
% \vspace{-3mm}
\end{table}

% fig:other_use
\begin{figure*}[t]
\centering
\includegraphics[width=1.0\textwidth]{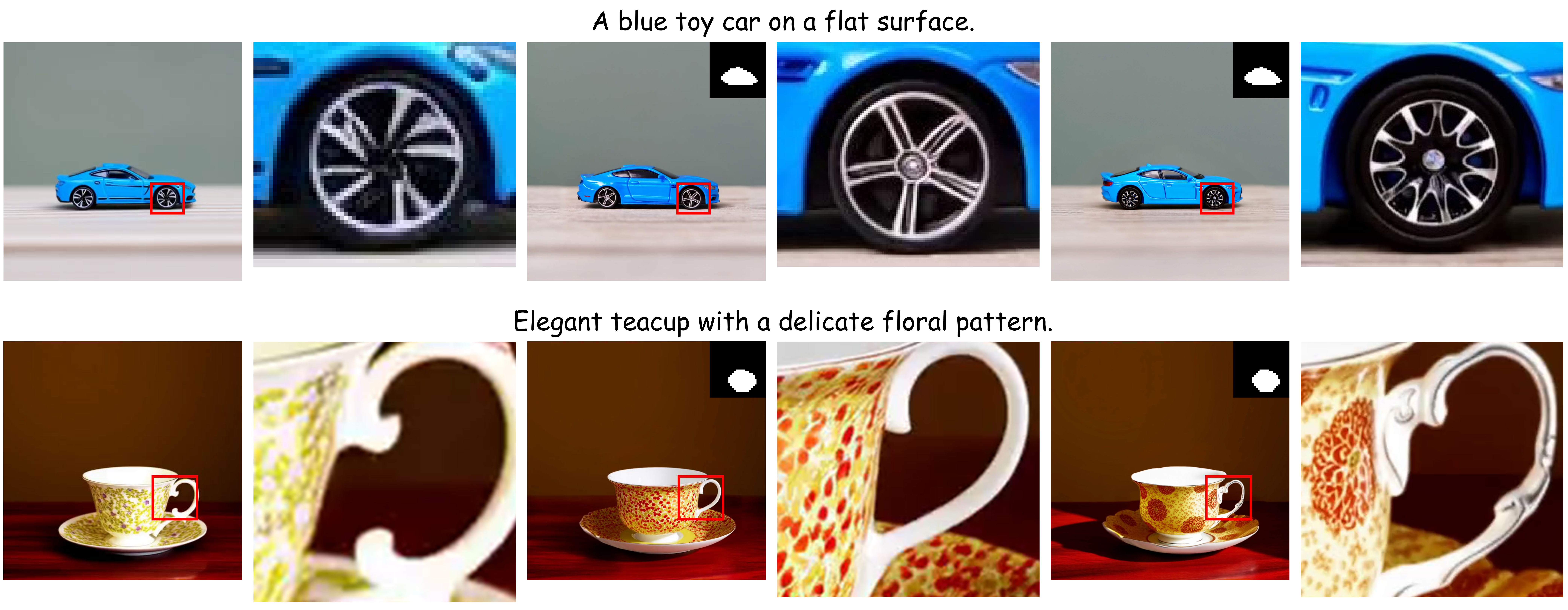}\par
% \vspace{-1mm}
\makebox[0.333\textwidth]{\centering (a) 480p}%
\makebox[0.333\textwidth]{\centering (b) 480p + 960p}%
\makebox[0.333\textwidth]{\centering (c) 480p + 1920p}
% \vspace{-5.5mm}
\caption{
We show that our method enables selected regions to be rendered at ultra-high resolution while the remaining areas stay at lower resolution. Specifically, we present results using the Wan2.1-14B model~\cite{wan2025wan} for: (a) 480$\times$480 only, (b) 480$\times$480 + 960$\times$960, and (c) 480$\times$480 + 1920$\times$1920. The high-resolution regions are indicated by the upper-right masks, and we provide zoomed-in views of these areas. As the target resolution increases, the visual quality and level of detail in the high-resolution regions improve, while the overall generation remains consistent.
}
\label{fig:other_use}
% \vspace{-3mm}
\end{figure*}

% tab:higher_res_wan
\begin{table}[t]
\centering
\caption{
\textbf{Higher-resolution video generation} (1088$\times$1920) on Wan2.1-1.3B \cite{wan2025wan}, using CineScale \cite{qiu2025cinescale} for resolution adaptation.
}
\label{tab:higher_res_wan}
% \vspace{-3mm}
\begingroup
\setlength{\tabcolsep}{4.5pt}
\begin{adjustbox}{max width=\columnwidth}
\begin{tabular}{l ccc ccc c}
\toprule
 & \multicolumn{3}{c}{\textbf{DOVER} $\uparrow$} & \multicolumn{3}{c}{\textbf{VBench} $\uparrow$} & \textbf{Time} \\
\cmidrule(lr){2-4} \cmidrule(lr){5-7} %
\textbf{Method} & Aes. & Tech. & Overall & Qual. & Sem. & Total & (s) $\downarrow$ \\
\midrule
Wan w/ CineScale & 99.86 & 10.81 & 80.03 & 80.95 & 61.00 & 76.96 & 716 \\
\rowcolor{grayline} Wan w/ CineScale + Ours & 99.85 & 9.72 & 80.19 & 80.59 & 65.23 & 77.51 & 161 \\
\bottomrule
\end{tabular}
\end{adjustbox}
\endgroup
% \vspace{-3mm}
\end{table}

% tab:higher_res_wan
\begin{table}[!t]
\centering
\caption{
\textbf{Higher-resolution image generation} (2048$\times$2048) on FLUX.1-dev \cite{labs2025flux1kontextflowmatching}, using DyPE \cite{issachar2025dype} for resolution adaptation.
}
\label{tab:higher_res_flux}
% \vspace{-3mm}
\begingroup
\setlength{\tabcolsep}{3pt}
\begin{adjustbox}{max width=\columnwidth}
\begin{tabular}{l ccccc c}
\toprule
\textbf{Method} & \textbf{ImgR.} $\uparrow$ & \textbf{CLIP-IQA} $\uparrow$ & \textbf{MUSIQ} $\uparrow$ & \textbf{CLIP} $\uparrow$ & \textbf{Time(s)} $\downarrow$ \\
\midrule
FLUX (w/ DyPE) & 0.919 & 0.458 & 55.05 & 30.93 & 21.2 \\
\rowcolor{grayline} FLUX + Ours (w/o DyPE) & 1.014 & 0.522 & 56.74 & 31.01 & 9.2 \\
\rowcolor{grayline} FLUX + Ours (w/ DyPE) & 1.023 & 0.533 & 58.14 & 30.96 & 9.2 \\
\bottomrule
\end{tabular}
\end{adjustbox}
\endgroup
% \vspace{-3mm}
\end{table}

% tab:rb_four_res_wan
\begin{table}[t] 
\centering
\caption{
\textbf{4-mixed-resolution (480, 960, 1920, 3840p) video generation} with Wan2.1-1.3B \cite{wan2025wan}.
}
\label{tab:rb_four_res_wan}
\begingroup
\scriptsize 
\setlength{\tabcolsep}{4.5pt}
\begin{adjustbox}{max width=\columnwidth}
\begin{tabular}{l ccc ccc c}
\toprule
 & \multicolumn{3}{c}{\textbf{DOVER} $\uparrow$} & \multicolumn{3}{c}{\textbf{VBench} $\uparrow$} & \textbf{Time} \\
\cmidrule(lr){2-4} \cmidrule(lr){5-7} %
\textbf{Method} & Aesthetic & Technical & Overall & Quality & Semantics & Total & (s) $\downarrow$ \\
\midrule
\igray{Wan-3.8k}
& \igray{86.64} & \igray{3.01} & \igray{25.37}
& \igray{68.46} & \igray{14.37} & \igray{57.65}
& \igray{6213} \\
\midrule
PI-LR~\cite{chen2023extending}
& 59.67 & 3.85 & 16.73
& 69.47 & 35.48 & 62.67
& \multirow{4}{*}{354} \\
PI-HR~\cite{chen2023extending}
& \underline{64.90} & \underline{3.98} & \underline{21.15}
& \underline{71.34} & \underline{38.66} & \underline{64.81}
& \\
NTK~\cite{peng2023ntk}
& 42.09 & 2.31 & 9.74
& 67.85 & 28.01 & 59.88
&  \\
YaRN~\cite{peng2023yarn}
& 36.18 & 2.37 & 8.78
& 67.72 & 32.85 & 60.74
&  \\
\midrule
\textbf{Ours}
& \textbf{99.25} & \textbf{9.39} & \textbf{71.04}
& \textbf{81.16} & \textbf{53.66} & \textbf{75.66}
& \textbf{354} \\
\bottomrule
\end{tabular}
\end{adjustbox}
\endgroup
\end{table}

% tab:rb_four_res_flux
\begin{table}[t] 
\centering
\caption{
\textbf{4-mixed-resolution (320, 640, 1280, 2560) image generation} with FLUX.1-dev \cite{labs2025flux1kontextflowmatching}.
}
\label{tab:rb_four_res_flux}
\begingroup
\scriptsize 
\setlength{\tabcolsep}{4.5pt}
\begin{adjustbox}{max width=\columnwidth}
\begin{tabular}{l cccc c}
\toprule
\textbf{Method} & \textbf{ImgReward} $\uparrow$ & \textbf{CLIP-IQA} $\uparrow$ & \textbf{MUSIQ} $\uparrow$ & \textbf{CLIP} $\uparrow$ & \textbf{Time} $\downarrow$ \\
\midrule
\igray{FLUX-2.5k} & \igray{0.477} & \igray{0.428} & \igray{35.64} & \igray{30.93} & \igray{38.9 s} \\
\midrule
PI-LR~\cite{chen2023extending} & 0.118 & \underline{0.399} & 36.19 & 25.24 & \multirow{4}{*}{11.2 s} \\
PI-HR~\cite{chen2023extending} & \underline{0.173} & 0.366 & 33.25 & \underline{28.74} &  \\
NTK~\cite{peng2023ntk} & 0.157 & 0.295 & 38.21 & 26.88 &  \\
YaRN~\cite{peng2023yarn} & 0.132 & 0.227 & \underline{39.16} & 24.84 & \\
\midrule
\textbf{Ours} & \textbf{0.836} & \textbf{0.556} & \textbf{40.74} & \textbf{31.30} & \textbf{11.2 s} \\
\bottomrule
\end{tabular}
\end{adjustbox}
\endgroup 
\end{table}

% tab:abl_subsample_hr_kv
\begin{table}[!t]
\centering
\caption{
\textbf{Ablation on HR key/value handling for LR queries}: different approaches yield similar performance, suggesting that LR queries mainly rely on coarse contextual information.
}
\label{tab:abl_subsample_hr_kv}
% \vspace{-3mm}
\begingroup
\setlength{\tabcolsep}{4.5pt}
\begin{adjustbox}{max width=\columnwidth}
\begin{tabular}{l ccc ccc c}
\toprule
 & \multicolumn{3}{c}{\textbf{DOVER} $\uparrow$} & \multicolumn{3}{c}{\textbf{VBench} $\uparrow$} & \textbf{Time} \\
\cmidrule(lr){2-4} \cmidrule(lr){5-7} %
% \textbf{Method} & Aesthetic & Technical & Overall & Quality & Semantics & Total & (s) $\downarrow$ \\
\textbf{Method} & Aes. & Tech. & Overall & Qual. & Sem. & Total & (s) $\downarrow$ \\
\midrule
Strided Downsample (Ours) & \underline{99.63} & 10.01 & 75.34 & \underline{80.76} & \textbf{62.17} & \textbf{77.04} & \textbf{43.2} \\
Avg-Pool Downsample & \textbf{99.64} & 10.01 & \underline{75.44} & 80.63 & \underline{62.12} & \underline{76.93} & \underline{43.2} \\
No Downsample & 99.63 & 10.01 & \textbf{75.44} & \textbf{80.77} & 61.51 & 76.92 & 44.1 \\
\bottomrule
\end{tabular}
\end{adjustbox}
\endgroup
% \vspace{-3.5mm}
\end{table}

% tab:wan_overhead
\begin{table}[!t]
\centering
\caption{We report overhead of each component in our method with Wan2.1-1.3B \cite{wan2025wan}.}
\label{tab:wan_overhead}
% \vspace{-3mm}
\begingroup
\setlength{\tabcolsep}{4.5pt}
\begin{adjustbox}{max width=\columnwidth}
\begin{tabular}{lcc}
    \toprule
    \textbf{Component} & \textbf{Latency (s)} & \textbf{Percentage} \\
    \midrule
    Ours (total)                & 43.2          & 100 \% \\
    \midrule
    PMA                        & 0.137       & 0.3 \% \\
    Tiny VAE~\cite{BoerBohan2025TAEHV}                    & 0.030       & 0.07 \% \\
    Saliency model~\cite{linardos2021deepgaze}              & 0.273       & 0.6 \% \\
    Latent up/down sampler      & 1.3         & 2.9 \% \\
    \bottomrule
\end{tabular}
\end{adjustbox}
\endgroup
% \vspace{-3.5mm}
\end{table}

\noindent\textbf{Importance-based region selection.}
In \cref{fig:rb_sal_vis}, we provide an illustration of importance-based region selection: the LR generations at early timesteps, along with the important regions identified by saliency prediction that will be upsampled into HR tokens.
In \cref{fig:saliency_vis}, we show visualizations for video generation.

\noindent\textbf{Integration with acceleration methods for image generation.}
As shown in \cref{tab:flux_integrate_acc} and \cref{fig:flux_integrate_acc}, our approach integrates seamlessly with orthogonal diffusion acceleration techniques, enabling further reductions in inference time while maintaining comparable quality for image generation.

\noindent\textbf{Higher-resolution generation.}
In \cref{tab:higher_res_wan,tab:higher_res_flux,fig:flux_2k_ours}, we show that our method can be extended to higher-resolution generation, \ie, 2K, when combined with resolution adaptation techniques. Even without such adaptation, it remains competitive. We believe this is because, during mixed-resolution denoising, the number of tokens remains relatively small and close to the range seen during training at standard resolutions. 
In \cref{fig:rb_more_vis}, we provide visual comparisons against baseline methods at 2K resolution.
For resolution adaptation, we use CineScale \cite{qiu2025cinescale} for video generation, \ie, an extension of FreeScale \cite{qiu2025freescale}, and Dynamic Position Extrapolation (DyPE) \cite{issachar2025dype} for image generation.

\noindent\textbf{Mixed-resolution with ultra-high-resolution regions.}
In \cref{fig:other_use}, we show that our method has the potential to make selected regions reach ultra-high resolution while other areas remain low resolution. We use Wan2.1-14B \cite{wan2025wan}, with the low-resolution region set to 480p and the ultra-high-resolution region set to 1920p.

\noindent\textbf{Multi-resolution scaling.}
In \cref{tab:rb_four_res_wan,tab:rb_four_res_flux}, we show our method naturally extends to 4-mixed-resolution with stable image and video generation results.

\noindent\textbf{Ablation on subsampling HR keys.}
In \cref{tab:abl_subsample_hr_kv}, we conduct an ablation study on how HR keys are handled for LR queries (Sec. 4.1 in the main text for PMA, under \emph{LR queries vs. all keys}). Besides our default setting that downsamples HR keys (and values) to the LR grid via strided downsampling, we also test two alternatives: keeping all HR keys without downsampling, and replacing strided downsampling with mean average pooling. All three variants yield very similar results, indicating that LR queries mainly require coarse contextual information. Keeping all HR keys leads to slightly slower generation, while providing no clear performance gain.

\noindent\textbf{Overhead of each component.}
In \cref{tab:wan_overhead}, we report the overhead incurred by each component in our method, including Phase-Aligned Mixed-Resolution Attention (PMA), the tiny VAE \cite{BoerBohan2025TAEHV}, the saliency model (DeepGazeIIE) \cite{linardos2021deepgaze}, and the latent up/downsampler.

\noindent\textbf{Detailed VBench results.}
In \cref{tab:wan_rope_detail_vbench,tab:wan_compare_acc_detail_vbench}, we show detailed quantitative evaluation results across all 16 dimensions of VBench \cite{huang2024vbench}, which correspond to Tables~1 and~3 in the main text, respectively.

\begin{figure*}[t]
\centering
\includegraphics[width=1.0\textwidth]{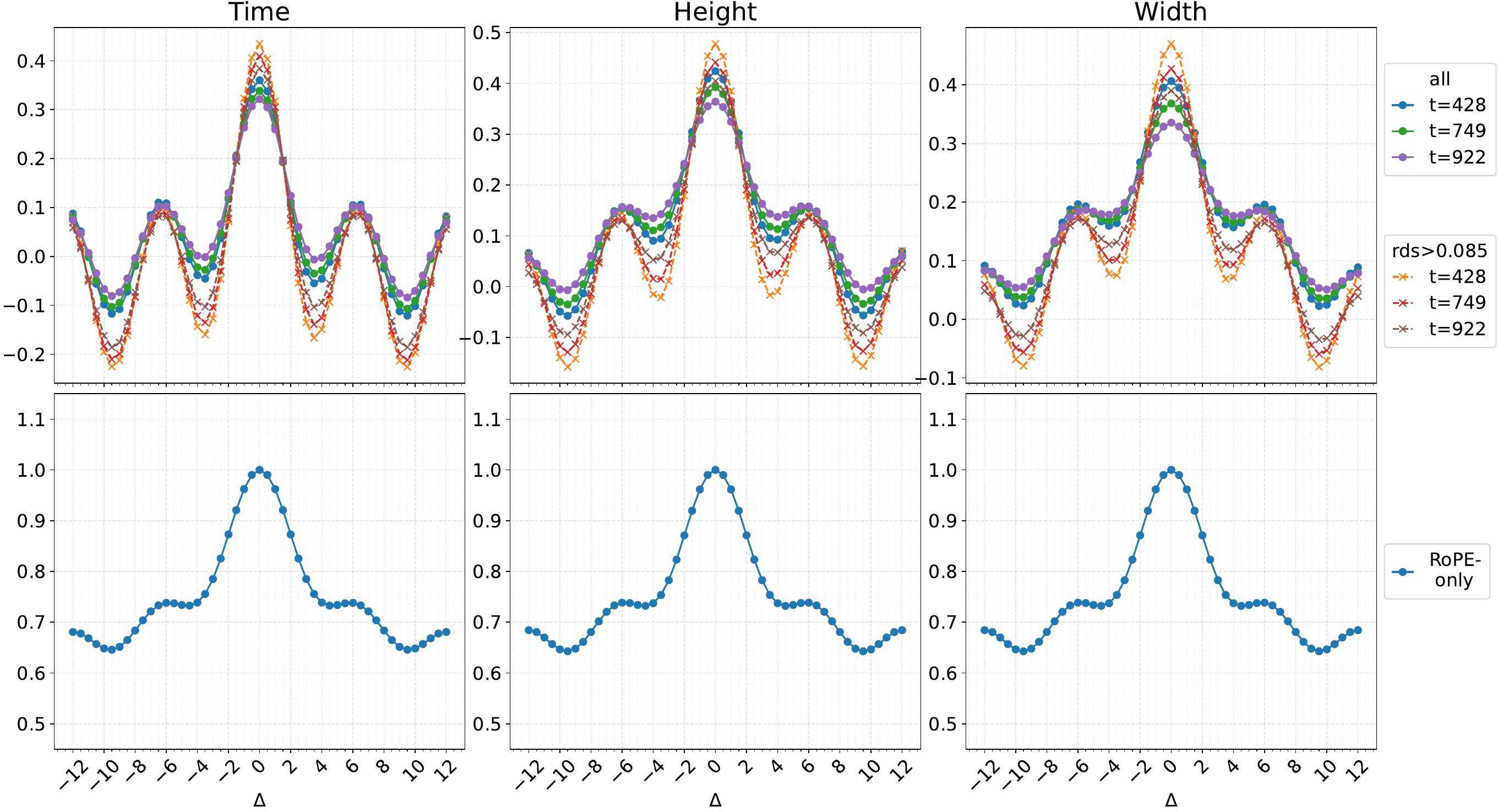}
% \vspace{-5mm}
\caption{
\textbf{RoPE-only $\kappa(\Delta)$ curves on Wan 2.1-1.3B} \cite{wan2025wan}. For comparison with the trained model (Fig.~3 in the main text is replotted as the first row in this figure), we plot in the second row the 
$\kappa(\Delta)$ curves produced by RoPE alone, \ie, without any influence from token content or attention heads. 
Each curve is a superposition of sinusoids with frequencies defined by RoPE. We observe that the trained model significantly alters the frequency composition compared to the RoPE-only scenario, exhibiting different value ranges, magnitudes, and strengths of peaks and troughs. Moreover, along different axes, \ie, time versus height/width, the nature of these changes differs substantially.
}
\label{fig:rope_only}
% \vspace{-3mm}
\end{figure*}

\begin{figure*}[t]
\centering
\includegraphics[width=1.0\textwidth]{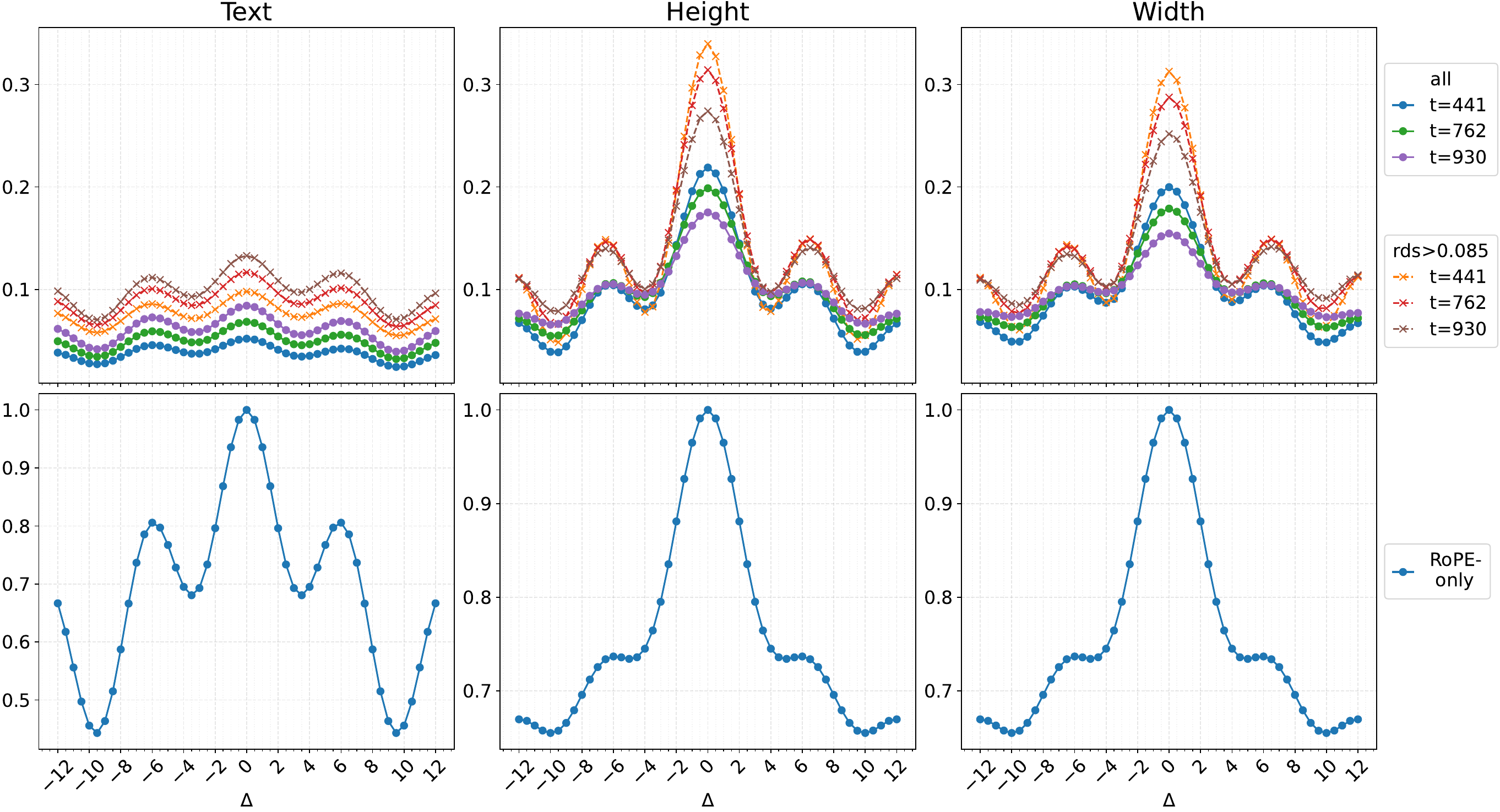}
% \vspace{-5mm}
\caption{
\textbf{$\kappa(\Delta)$ curves on FLUX.1-dev} \cite{labs2025flux1kontextflowmatching}.
The first row shows the measured $\kappa(\Delta)$ curves on FLUX.1-dev along the text, image height, and image width axes. The second row shows RoPE-only curves, \ie, without influences from token content or attention heads. Along the height and width axes, the learned curves remain strongly periodic with sharp peaks and troughs. Along the text axis, however, the learned curve is much smoother, even if the RoPE-only curve is still sharply periodic. This contrast indicates that attention heads learn modality-dependent corrections to the RoPE prior, reshaping the effective attention bias differently for text and image tokens.
}
\label{fig:flux_delta_curve_and_rope_only}
% \vspace{-3mm}
\end{figure*}

\subsection{Additional Analyses}
\label{ssec:add_analyses}

\noindent\textbf{RoPE-only $\kappa(\Delta)$ on Wan} \cite{wan2025wan}. To isolate the positional bias imposed by RoPE itself, we compute the mean normalized attention score $\kappa(\Delta)$ as a function of relative distance $\Delta$ using only the RoPE rotations, removing any contribution from token content and learned attention weights. As shown in \cref{fig:rope_only}, the resulting curve is a superposition of sinusoids with frequencies defined by RoPE. 
This curve serves as a structural baseline: in the trained model 
(Fig. 3 in the main text), attention heads reshape this profile substantially: amplifying or suppressing specific frequencies and altering the relative peak magnitudes. In other words, each head develops preferences for particular phase increments that reflect regularities in the training data, \eg, common spatial offsets or geometric patterns.

\noindent\textbf{Analysis of $\kappa(\Delta)$ on FLUX} \cite{labs2025flux1kontextflowmatching}. We further study $\kappa(\Delta)$ on FLUX.1-dev. As shown in \cref{fig:flux_delta_curve_and_rope_only}, on height and width axes, $\kappa(\Delta)$ curves have clear periodicity and sharp peaks and troughs. In contrast, on the text axis, the learned attention heads produce a substantially smoother profile, even if the RoPE-only baseline still displays periodic sharp peaks and troughs. This indicates that attention heads learn different biases across modalities. These results highlight the importance of empirical study of $\kappa(\Delta)$, since the effective attention bias can differ substantially from the RoPE-only baseline and can vary significantly across modalities even within the same model.

\section{Derivation of Equation 6}
\label{sec:derivation}

\noindent\textbf{Intuition.}
RoPE works by rotating every 2D pair of query and key features by an angle proportional to their positions. 
When we compute the dot product between these rotated vectors, the rotation angles appear inside cosine and sine terms. 
As a result, the RoPE-modulated attention score for a single pair behaves like a single sinusoid in the relative offset $\Delta = p_k - p_q$. 
Summing over all rotary pairs yields a mixture of sinusoids at the predefined RoPE frequencies.
This means that the attention score as a function of $\Delta$ is effectively a multi-frequency positional kernel whose amplitudes and phase shifts depend on the token content $(q,k)$, while the frequencies are fixed by RoPE. This structure is what we make explicit below.

\noindent\textbf{Derivation.} 
Consider a single attention head with dimensionality $d$ equipped with RoPE.
RoPE associates each coordinate pair $(2i,2i{+}1)$, $i\in\{0,\ldots,d/2-1\}$, with an angular frequency $\omega_i$. For a scalar position $p\in\mathbb{R}$, the $i$-th 2D subvector is rotated by angle $\theta_i(p)=\omega_i p$ using the $2\times 2$ rotation
\[
R(\theta)=
\begin{bmatrix}
\cos\theta & -\sin\theta\\
\sin\theta & \cos\theta
\end{bmatrix}.
\]
Splitting $q,k\in\mathbb{R}^{d}$ into 2D pairs
\[
q_i=\begin{bmatrix}q_{2i}\\ q_{2i+1}\end{bmatrix},\quad
k_i=\begin{bmatrix}k_{2i}\\ k_{2i+1}\end{bmatrix},
\]
the RoPE relative-position property (Eq.~(3) in the main text) implies that the pre-softmax score between a query at position $p_q$ and a key at position $p_k$ depends on the relative offset $\Delta = p_k - p_q$ and can be written as
\[
\text{score}(q,k,\Delta)=\sum_i q_i^\top R(\omega_i\Delta)\,k_i.
\]
Expanding each term yields
\[
q_i^\top R(\omega_i\Delta)\,k_i
= A_i(q,k)\,\cos(\omega_i\Delta) + B_i(q,k)\,\sin(\omega_i\Delta),
\]
where
\[
\begin{aligned}
A_i(q,k) &= q_{2i}k_{2i}+q_{2i+1}k_{2i+1}, \\
B_i(q,k) &= q_{2i+1}k_{2i}-q_{2i}k_{2i+1}.
\end{aligned}
\]
Convert to amplitude–phase form by defining
\[
\begin{aligned}
C_i(q,k) &= \sqrt{A_i(q,k)^2+B_i(q,k)^2}\;\ge 0, \\
\phi_i &= \mathrm{atan2}\big(-B_i(q,k),\,A_i(q,k)\big),
\end{aligned}
\]
so that:
\[
A_i\cos(\omega_i\Delta)+B_i\sin(\omega_i\Delta)
= C_i(q,k)\cos\!\big(\omega_i\Delta+\phi_i\big).
\]
Summing over $i$ gives
\[
\text{score}(q,k,\Delta)
=\sum_i C_i(q,k)\,\cos\!\big(\omega_i\Delta+\phi_i\big),
\]
which motivates the expected-score kernel approximation used in Equation 6 of the main text.

\section{Additional Experimental Settings}
\label{sec:add_experimental_settings}

\noindent\textbf{Metrics.}
For the evaluation of video generation, we follow VBench \cite{huang2024vbench} and generate using the full set of prompts provided in the benchmark. We also use the reference-free metric DOVER \cite{wu2023dover} to evaluate these videos and report the corresponding DOVER metrics. For VBench, we report quality, semantics, and total scores; for DOVER, we report aesthetic, technical, and overall scores.
For the evaluation of image generation, we report ImageReward \cite{xu2023imagereward} for photorealism; MUSIQ \cite{ke2021musiq} and CLIP-IQA \cite{wang2022exploring} for image quality assessment; and CLIP score \cite{radford2021learning} for prompt alignment, using the MSCOCO 2014 validation dataset~\cite{lin2014microsoft}, with 5K randomly sampled image and caption pairs.

\noindent\textbf{Latent up/downsampler.}
Our latent up/downsampler for Wan \cite{wan2025wan} is a 3D convolutional network with 3 residual blocks and one upsampling or downsampling layer, with hidden dimension 384 and about 25M parameters for each resizer model.
Our latent up/downsampler for FLUX \cite{labs2025flux1kontextflowmatching} is a 2D convolutional network with 20 residual blocks and one upsampling or downsampling layer, with hidden dimension 128 and about 6M parameters for each resizer model.

As already mentioned in the main text, we train the latent up/downsamplers using paired targets obtained by pixel-space resizing and re-encoding, optimized with $\ell_1$ in latent space and $\ell_1+\mathrm{LPIPS}$ in pixel space. Specifically, the weight for the latent-space $\ell_1$ loss is 0.01, while the weights for the pixel-space $\ell_1$ loss and LPIPS loss are 1 and 0.1, respectively. We use a batch size of 1, with the Pexels dataset \cite{pexels_license, languagebind_open_sora_plan_v1_1_0} for images and Aesthetic-Train-V2 dataset \cite{zhang2025diffusion4k,zhang2025ultrahighresolutionimagesynthesis} for videos.

\noindent\textbf{Measuring $\kappa(\Delta)$.}  
We attach forward hooks to the self-attention blocks and collect pre-attention tokens for each self-attention head, at three denoising steps (we select 20\%, 50\%, and 80\% quantiles). 
We randomly sample 10K query-key pairs, regardless of token positions. 
Then, we apply standard RoPE rotation with offset $\Delta \in [-12,12]$ (with a step size of $0.5$), and compute their cosine similarity score for each RoPE axis. 
Finally, we average the scores over tokens and heads to obtain $\kappa(\Delta)$. 
For both Wan \cite{wan2025wan} and FLUX \cite{labs2025flux1kontextflowmatching}, we use 300 randomly generated text prompts to collect tokens and compute the scores. 

We find that 300 prompts already produce a stable estimate: in \cref{fig:rb_kappa} (a), we show that 1K prompts yield similar curves. In addition, due to noise shift in the scheduler, the timesteps are non-uniform. In \cref{fig:rb_kappa} (b), we show that uniform timesteps lead to similar results.

\noindent\textbf{Our configuration details.}
For video diffusion with Wan2.1-1.3B \cite{wan2025wan}, we use 15 denoising steps at 480p and 35 mixed-resolution steps that combine 480p and 960p, with a 15\% high-resolution token ratio during the mixed-resolution stage and a CFG scale of 5.0, to generate 49-frame videos. For saliency detection, we adopt the off-the-shelf model DeepGazeIIE \cite{linardos2021deepgaze}, which takes only 0.27 s.

For image generation with FLUX.1-dev \cite{labs2025flux1kontextflowmatching}, in the two-stage setting, we use 15 denoising steps: 5 steps at 512 and 10 mixed-resolution steps that combine 512 and 1024 resolutions, with a 60\% high-resolution token ratio and a CFG scale of 3.5. 
In the three-stage setting involving coarse, mixed, and fine stages, we set the number of inference steps for each stage to $N = [6, 7, 5]$ for 4$\times$ acceleration and $N = [2, 3, 5]$ for 6$\times$ acceleration, with a 30\% high-resolution token ratio. We adopt the off-the-shelf saliency model DeepGazeI \cite{deepgazei}, which takes only 0.01 s.

We also integrate our method with acceleration techniques. 
When combined with caching, for video generation, we set the skipping threshold $\delta$ to 0.08 for TeaCache \cite{liu2025timestep} and 0.1 for MagCache \cite{ma2025magcache}, while for image generation we use $\delta$ as 0.4 for TeaCache and 0.1 for MagCache. 
When combined with step distillation, we use the DMD model of Wan 2.1, \ie, distribution matching distillation \cite{yin2024one}, obtained from CausVid \cite{yin2025causvid} for video generation, setting LR steps to 1 for both 4-step and 8-step generation. For image generation, we use FLUX.1-schnell \cite{blackforestlabs_flux1_schnell_2024} with LR steps set to 1 and 4 total steps.

\noindent\textbf{Baseline configuration details.}
For the baseline RoPE interpolation methods in Table~1 and~2 of the main text: in the NTK-aware interpolation, we set the NTK scaling factor to 2; in the hybrid linear and NTK-aware interpolation, the linear position scaling factor is set to 1.5, while the NTK scaling factor is set to 1.333.

For the baseline diffusion acceleration methods, we use the following configurations to ensure a fair comparison under matched acceleration settings. 
For advanced diffusion inference samplers used in video generation, including UniPC \cite{zhao2023unipc} and DPM++ \cite{lu2025dpm}, we set the number of denoising steps to 13. 
For ToMe \cite{bolya2023token}, a token merging method that merges tokens before attention and restores them afterward, we use a merging ratio of 0.5 and apply merging only to the middle transformer blocks for better performance.
TeaCache \cite{liu2025timestep} and MagCache \cite{ma2025magcache} are temporal feature caching methods that selectively reuse intermediate outputs during denoising. In video generation, we set the skipping threshold to $\delta=0.27$ for TeaCache and $\delta=0.8$ for MagCache. In image generation, we use $\delta=1.8$ for TeaCache and $\delta=1.6$ for MagCache under 4$\times$ acceleration, and $\delta=4$ for MagCache under 6$\times$ acceleration.
RALU \cite{jeong2025upsample} uses coarse-, mixed-, and fine-resolution denoising, similar to our method. However, it does not address the issue of mixed-resolution attention and therefore requires more high-resolution steps to compensate. We set the stage-wise inference steps to $N=[5,6,7]$ for 4$\times$ acceleration and $N=[2,3,5]$ for 6$\times$ acceleration, with a 30\% high-resolution token ratio.
For Bottleneck Sampling \cite{tian2025training}, a training-free method based on high--low--high resolution denoising with noise reintroduction and scheduler re-shifting, we set $N=[4,10,5]$ for 4$\times$ acceleration and $N=[2,7,3]$ for 6$\times$ acceleration. 
For ToCa \cite{zou2024accelerating}, which selectively caches token features based on importance, we set total steps $N=18$ for 4$\times$ acceleration and $N=8$ for 6$\times$ acceleration.

\section{Additional Backgrounds and Definitions}
\label{sec:add_backgrounds}

\noindent\textbf{RoPE in 1D and higher dimensions.}
We first recall a simple case of RoPE in 1D.
Let an attention head have even dimensionality $d$, and let $\omega_i$ denote the $i$-th angular frequency, typically a geometric sequence: $\omega_i = 10000^{-2i/d}, \ i \in \{0,1,\ldots,d/2-1\}$.
For a scalar position $p\in\mathbb{R}$, RoPE rotates each $(2i,2i{+}1)$ pair by angle $\theta_i(p)=\omega_i\,p$:
\begin{align}
R(\theta) &=
\begin{bmatrix}
\cos\theta & -\sin\theta\\
\sin\theta & \cos\theta
\end{bmatrix}, \\
\mathcal{R}(p) &= 
\mathrm{diag}\big(R(\theta_0(p)), \ldots, R(\theta_{d/2-1}(p))\big).
\end{align}
Given tokens $q,k\in\mathbb{R}^d$, RoPE applies $q\!\rightarrow\!\mathcal{R}(p_q)q$, $k\!\rightarrow\!\mathcal{R}(p_k)k$. The attention score obeys the relative property
\begin{equation}
\big(\mathcal{R}(p_q)q\big)^\top\big(\mathcal{R}(p_k)k\big)
~=~q^\top \mathcal{R}(\Delta)\,k,
\label{eq:rope-rel-supp}
\end{equation}
\textit{i.e.}, RoPE converts absolute positions to a frequency-coded phase that depends only on the relative offset $\Delta=p_k-p_q$.

For images (2D) and videos (3D), positions are tuples $p=(p^h,p^w)$ or $(p^t,p^h,p^w)$. Standard practices, \eg, Wan \cite{wan2025wan} and FLUX \cite{labs2025flux1kontextflowmatching}, assign disjoint coordinate pairs (or channel groups) to each axis and uses separable rotations:
$\mathcal{R}(p)=\mathcal{R}_h(p^h)\oplus\mathcal{R}_w(p^w)$ (and $\oplus\,\mathcal{R}_t(p^t)$ for video).

\noindent\textbf{RoPE-dominance score (rds).}
Following \cite{chen2024rotary}, we define the RoPE-dominance score (rds) for an attention head as the average alignment between the pairwise RoPE subspaces of its query (or key) weights. For a head $h$ with weight matrix $W^{(h)} \in \mathbb{R}^{d_h \times d_{\text{model}}}$, RoPE groups dimensions into $d/2$ rotary pairs $(2i,2i+1)$. For the $i$-th pair we consider the corresponding row vectors $W^{(h)}_{2i}, W^{(h)}_{2i+1}$ and define
\[
\cos \alpha_i^{(h)}
= \frac{\langle W^{(h)}_{2i},\, W^{(h)}_{2i+1} \rangle}
       {\|W^{(h)}_{2i}\| \,\|W^{(h)}_{2i+1}\|}.
\]
The RoPE dominance score of head $h$ is then
\[
rds^{(h)} \;=\; \frac{2}{d} \sum_{i=0}^{d/2-1} \bigl|\cos \alpha_i^{(h)}\bigr|.
\]
Heads with large $rds^{(h)}$ are position-dominated, while heads with small $rds^{(h)}$ are token-content-dominated (their RoPE subspaces vary strongly with the input).

\noindent\textbf{NTK-aware scaling of RoPE (NTK).}
NTK~\cite{peng2023ntk} parameterization can be understood as a frequency-rescaling of RoPE that preserves the precision of high-frequency components while scaling the low-frequency components to accommodate longer sequence lengths. NTK modifies the RoPE frequencies according to:
\[
\omega_i^{\prime} = (\lambda \cdot 10000)^{-2i/d}, 
\qquad 
\lambda = s^{\, d/ (d - 2)}, 
\]
where $i = 0, \ldots, d/2-1$ and $s = L' / L$.

\noindent\textbf{YaRN.}
YaRN~\cite{peng2023yarn} proposes a detailed scheme for modifying the RoPE frequencies and rescaling the attention logits. In empirical evaluations, it achieves stronger training-free extrapolation than NTK and attains high performance on longer sequences with only modest fine-tuning on the target context length.
First, YaRN partitions all frequencies into three regions according to how many cycles they complete over the training length, quantified by
\[
    r_i = \frac{L \,\omega_i}{2\pi},
    \qquad i = 0,\ldots,\frac{d}{2}-1.
\]
Given two fixed thresholds $\alpha, \beta$ satisfying
\[
    r_{d/2-1} \le \alpha < \beta \le r_0,
\]
YaRN updates the RoPE frequencies via
\[
    \omega_i^{\prime}
    = \gamma(r_i)\,\omega_i + \bigl(1 - \gamma(r_i)\bigr)\,\frac{\omega_i}{s},
\]
where the interpolation coefficient $\gamma(r_i)$ is defined piecewise as
\[
    \gamma(r_i) =
    \begin{cases}
        1, & \text{if } r_i > \beta,\\[4pt]
        0, & \text{if } r_i < \alpha,\\[6pt]
        \dfrac{r_i - \alpha}{\beta - \alpha}, & \text{otherwise}.
    \end{cases}
\]

In addition, YaRN applies an attention scaling to stabilize training-free
extrapolation at long context lengths. Let $A$ denote the usual (pre-softmax)
attention logits matrix and let $\tau > 0$ be a temperature parameter. YaRN
replaces $A$ by
\[
    A^{\prime} \;=\; \frac{A}{\tau},
\]
that is, it evaluates the attention weights as
\[
    \operatorname{Attn}(Q,K,V)
    \;=\;
    \operatorname{softmax}\!\left(\frac{A}{\tau}\right)V,
\]
with $\tau$ chosen to balance the sharpness of attention between the original
and extrapolated context lengths.

\noindent\textbf{Saliency prediction.}
Saliency prediction aims to model human visual attention on images, evolving from early biologically inspired, bottom-up hand-crafted feature models \cite{itti2002model, kienzle2006nonparametric, zhang2008sun, NIPS2006_gbvs} to deep learning approaches facilitated by large datasets such as SALICON \cite{jiang2015salicon}. Leveraging pretrained CNNs, models such as DeepGaze I \cite{deepgazei, alexnet}, VGG-based architectures \cite{vggnet, wang2017deep, cornia2016deep, kruthiventi2017deepfix}, and LSTM-enhanced methods \cite{cornia2018predicting, liu2018deep, wang2019revisiting} achieved substantial improvements, with later works combining multiple backbones \cite{jia2020eml, linardos2021deepgaze}. Recent transformer-based models \cite{vaswani2017attention, vit} further advance the field by capturing long-range context for both image and video saliency prediction \cite{lou2022transalnet, zhou2023transformer, ma2022video}.

\section{Discussion and Limitations}
\label{sec:discussion}

\begin{figure*}[t]
\centering
\includegraphics[width=1.0\textwidth]{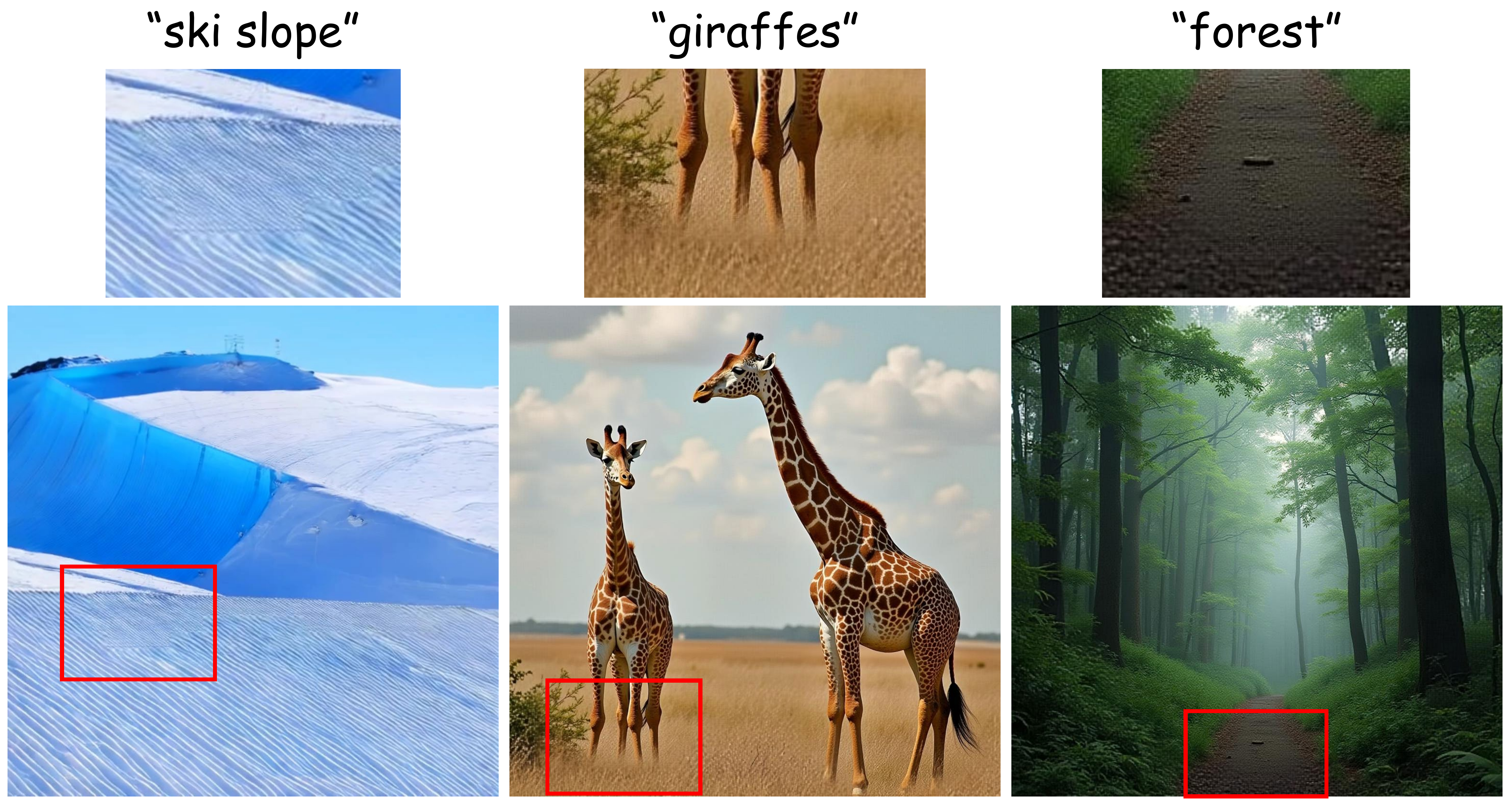} 
\caption{
\textbf{Failure cases.} Red boxes mark regions where complex LR-HR texture transitions still produce visible artifacts.
}
\label{fig:fail_cases} 
\end{figure*}

\begin{figure*}[t]
\centering
\includegraphics[width=1.0\textwidth]{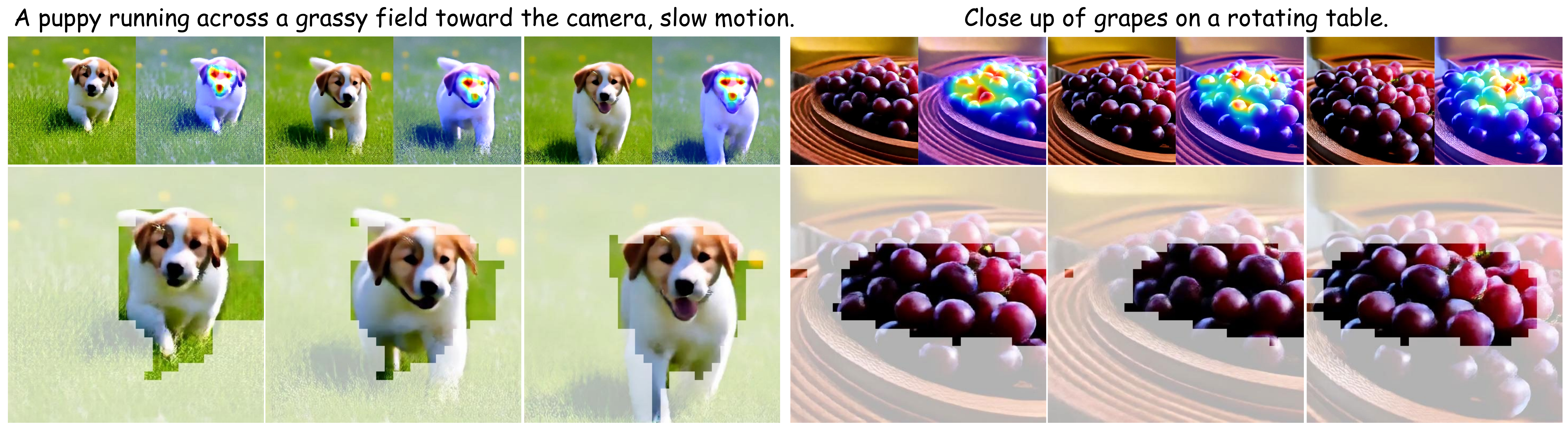}
% \vspace{-5mm}
\caption{Illustration of our importance-based region selection for text-to-video generation. Saliency prediction identifies the regions that will be upsampled into high-resolution tokens for mixed-resolution inference. The visualization is based on low-resolution outputs at inference step 15 (out of 50 total steps) using Wan 2.1 \cite{wan2025wan}.}
\label{fig:saliency_vis}
% \vspace{-2mm}
\end{figure*}

\begin{figure*}[t]
\centering
\includegraphics[width=1.0\textwidth]{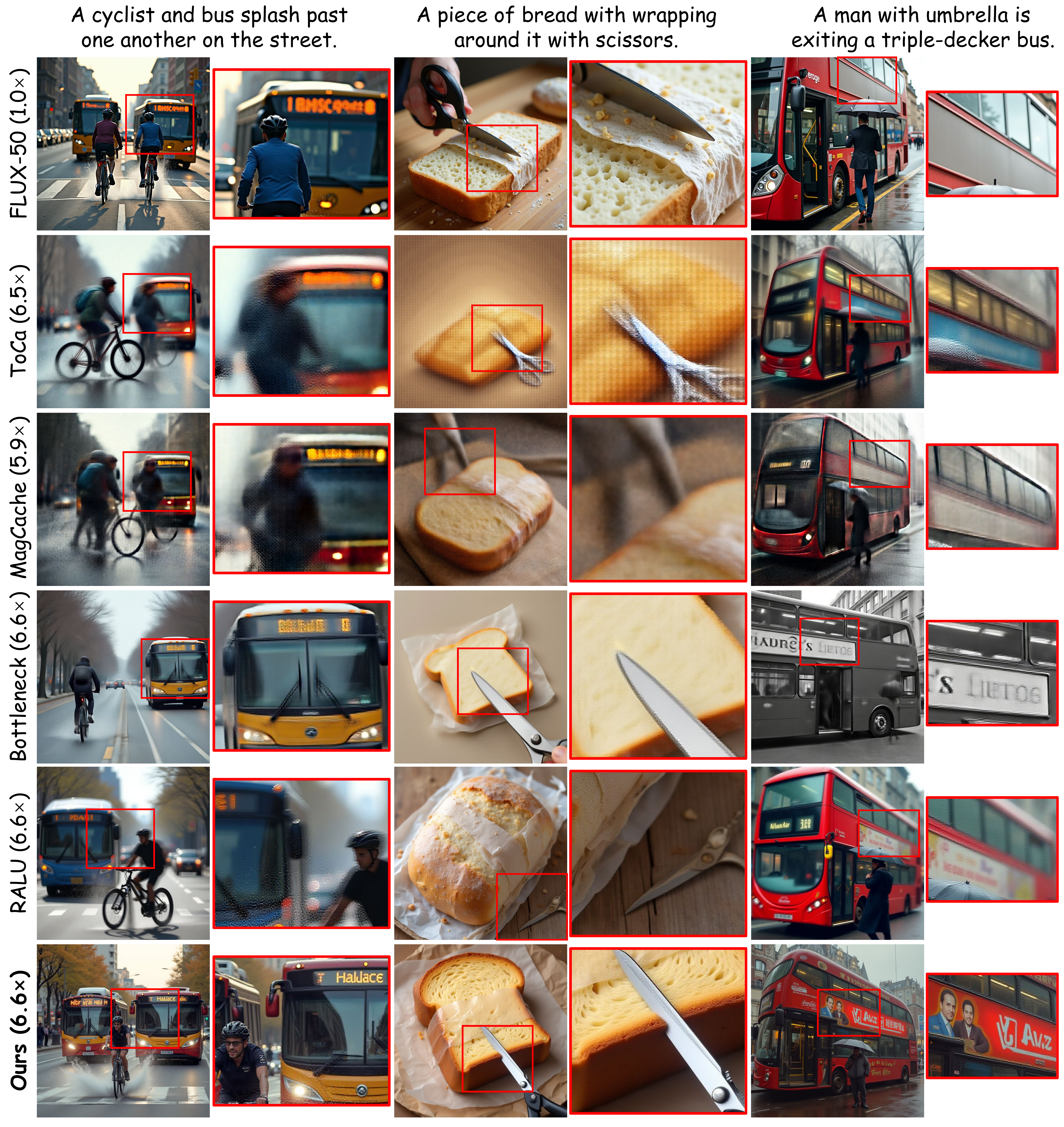}
\caption{
\textbf{Additional visual comparison with diffusion acceleration methods for image generation} on FLUX.1-dev \cite{labs2025flux1kontextflowmatching}. We compare at \(6\times\) speedups, with 50-step FLUX generations as reference.
Red boxes highlight the improved level of detail in our generation. 
}
\label{fig:flux_compare_acc_supp_6x}
\end{figure*}

\begin{figure*}[t]
\centering
\includegraphics[width=1.0\textwidth]{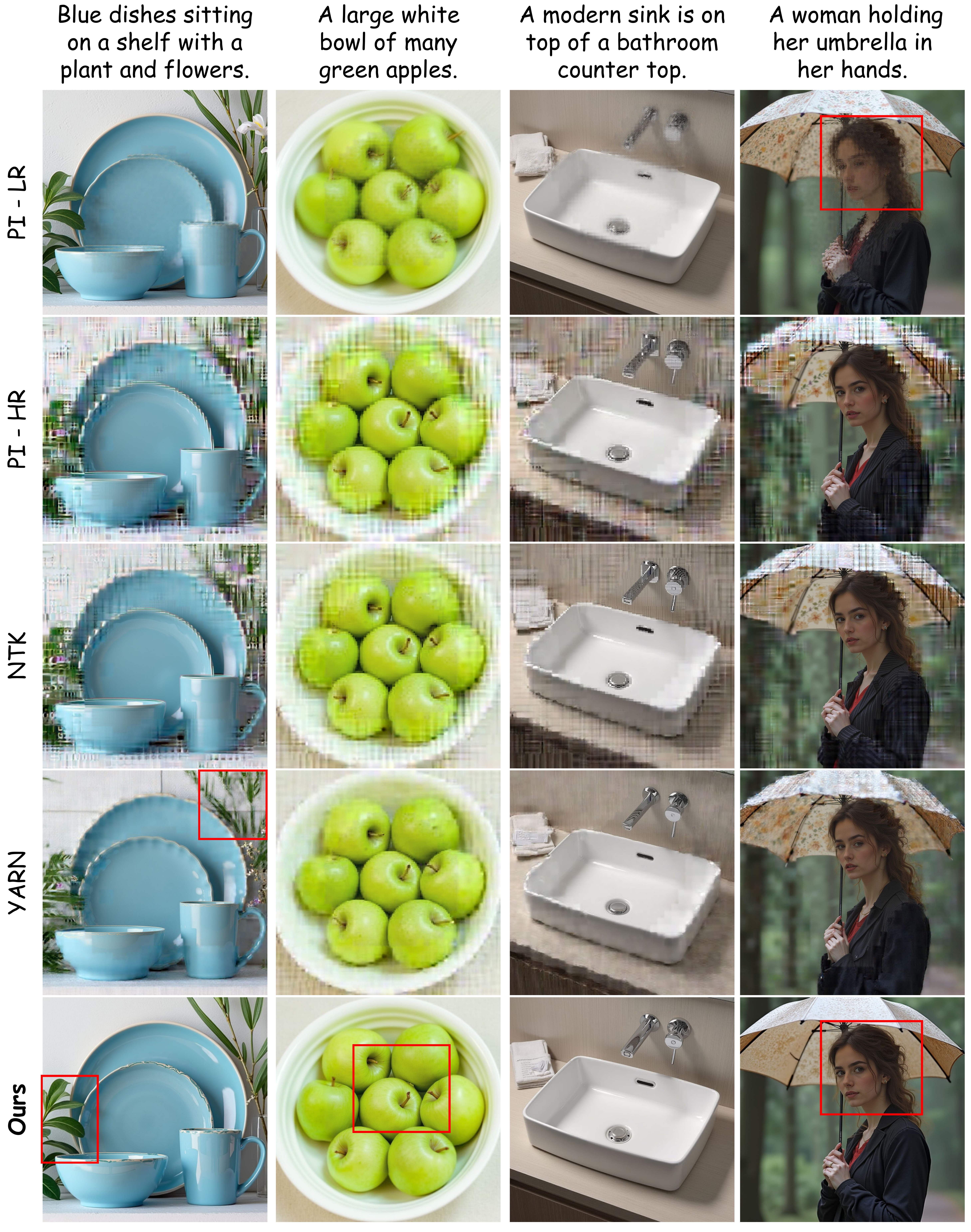}
\caption{
\textbf{Additional visual comparison with RoPE interpolation methods} applied to mixed-resolution denoising on FLUX.1-dev \cite{labs2025flux1kontextflowmatching}. As highlighted in red boxes, our method produces the most stable results.
}
\label{fig:flux_supp} 
\end{figure*}

\begin{figure*}[t]
\centering
\includegraphics[width=1\linewidth]{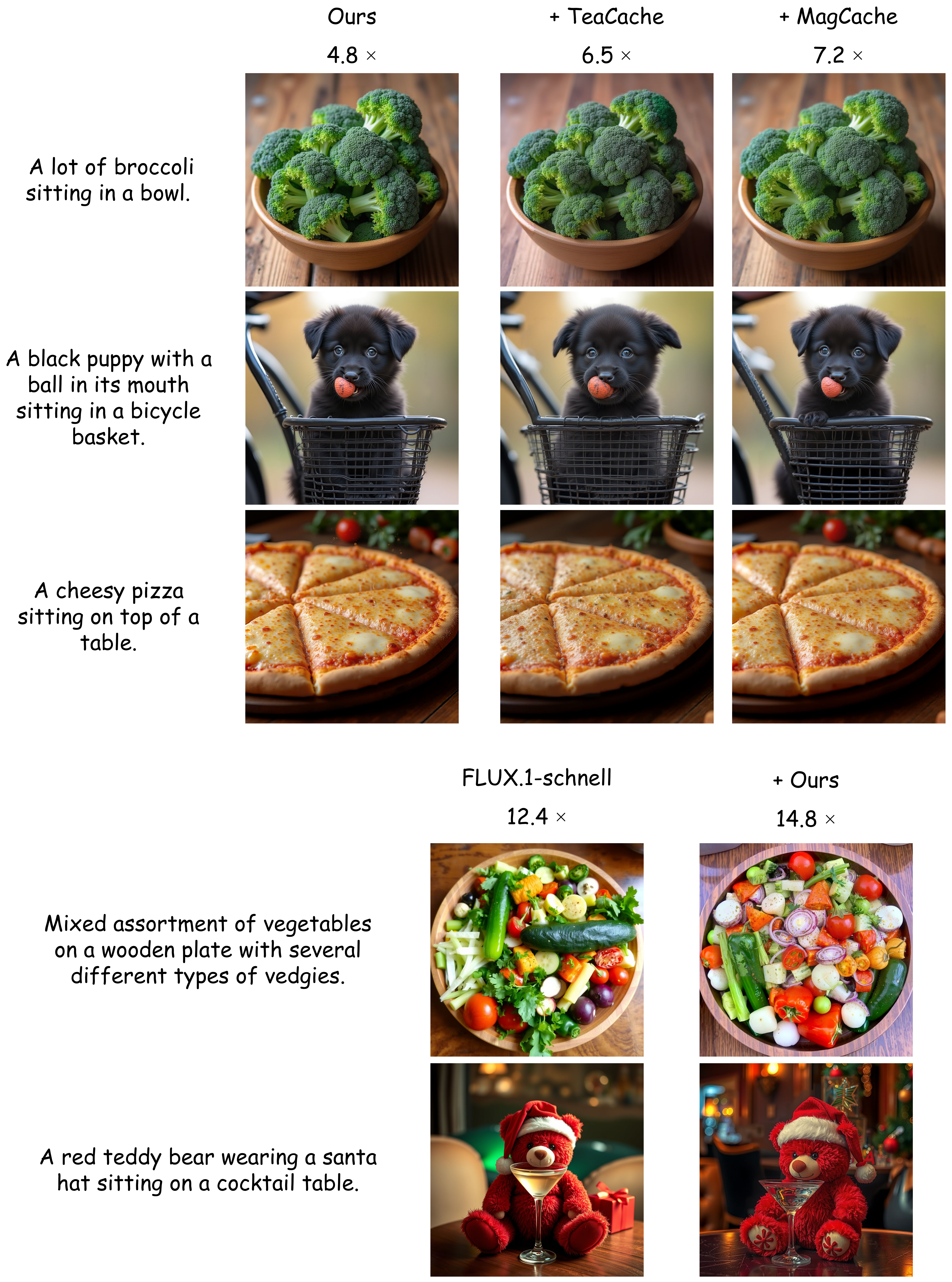} 
\caption{
\textbf{Integration with orthogonal diffusion acceleration methods} for image generation on FLUX.1-dev \cite{labs2025flux1kontextflowmatching}. We combine our method with feature caching methods and with the step-distillation model, FLUX.1-schnell \cite{blackforestlabs_flux1_schnell_2024}. Speedups are reported relative to FLUX.1-dev.
}
\label{fig:flux_integrate_acc}
\end{figure*}

Our method achieves stable and efficient mixed-resolution generation with pretrained DiTs via a phase-alignment mechanism that restores a consistent native positional scale. 
However, as shown in \cref{fig:fail_cases}, our method can still struggle for scenes with complex texture transitions between low-resolution and high-resolution regions. We can incorporate segmentation models to produce better saliency masks, which will partially alleviate this issue, but there is still substantial room for improvement.

A promising direction for future work is to move beyond inference-time correction by redesigning positional encodings and retraining DiTs accordingly, so that they can natively support mixed-resolution tokens in a more scale-invariant manner.

\section{Prompts}
\label{sec:prompts}

We provide the prompts used to generate the qualitative results shown in the paper but not included in the figures.

Figure~1 in the main text:
\begin{itemize}
    \item A minimalist studio photo of a white cotton crew-neck T-shirt with a vintage script logo reading “Credo” and small regular text “Comfortable Basics”, paired with dark blue jeans, shot against a soft light-gray background with natural lighting and a clean lifestyle aesthetic.
    \item A fluffy Pomeranian sitting inside a floral teacup on a warm wooden table, golden morning sunlight streaming in, warm bright color palette. The dog takes tiny sips of milk (no spills), blinking cutely while the camera slowly dolly-ins. Soft steam swirls, bokeh sparkles in the background, object-centric with the teacup and Pomeranian filling most of the frame.
    \item Close-up portrait of a young woman with wavy brown hair and blue eyes, neutral expression, soft studio lighting, shallow depth of field.
    \item A sugar glider perched on a ripe peach in an orchard at golden hour, warm glowing sunlight and bright colors. The glider unfurls its tiny wings and glides to a nearby branch in slow motion. A close-up of the glider’s face filling the frame.
    \item A cozy wooden cabin with smoke rising from the chimney, sitting in a green alpine meadow with dramatic mountain peaks in the background at sunset, cinematic landscape shot.
    \item A soft teddy bear barista, carefully pouring latte art into a tiny cup.
    \item Young woman in a pastel sweater sipping strawberry milk at a window cafe while rain sparkles outside, cozy happy vibe, soft natural light, cinematic close ups, slow push in.
    \item A small hedgehog holding a ripe strawberry, blinking slowly as leaves rustle behind it, subtle forward camera dolly, soft forest bokeh.
\end{itemize}

Figure~2 in the main text:
\begin{itemize}
    \item Shiny red sports car parked under bright city lights, realistic reflections, high detail, professional photography.
\end{itemize}

Figure~6 in the main text:
\begin{itemize}
    \item A panda playing on a swing set.
    \item A beautiful woman smiles.
\end{itemize}

Figure~7 in the main text:
\begin{itemize}
    \item A pile of oranges in crates topped with yellow bananas.
    \item A group of baseball players is crowded at the mound.
    \item A picture of a dog laying on the ground.
    \item A man holding a camera up over his left shoulder.
    \item A chicken sandwich in a wrapper near a cell phone.
    \item There is a small bus with several people standing next to it.
\end{itemize}

Figure~8 in the main text:
\begin{itemize}
    \item Giant sunflower slowly turning toward warm golden sunlight in a summer field.
    \item A close-up cinematic shot of a young woman standing outdoors in winter, wearing a bright red knit beanie and matching red scarf, soft natural daylight illuminating her face. She has gentle wavy light brown hair flowing slightly in the cold breeze. The background is softly blurred with cool gray urban tones, creating a shallow depth of field. She looks directly into the camera with a subtle, warm smile. Light snow begins to fall slowly around her. The camera gently pushes in, capturing fine details of her expression and the texture of the knit fabric. Soft winter ambiance, natural color grading, realistic skin tones, 4K, 60fps, shallow depth of field, cinematic, smooth handheld motion.
\end{itemize}

Figure~9 in the main text:
\begin{itemize}
    \item A building with a sign that reads One India Buildings.
    \item A little boy holding a brownie sandwich over a plate.
    \item A school bus and a silver car waiting at a railroad crossing for a train to go past.
\end{itemize}

Figure~11 in the main text:
\begin{itemize}
    \item A train going down the tracks that has just gone under a bridge.
\end{itemize}

\begin{figure*}[t]
\centering
\includegraphics[width=0.95\linewidth]{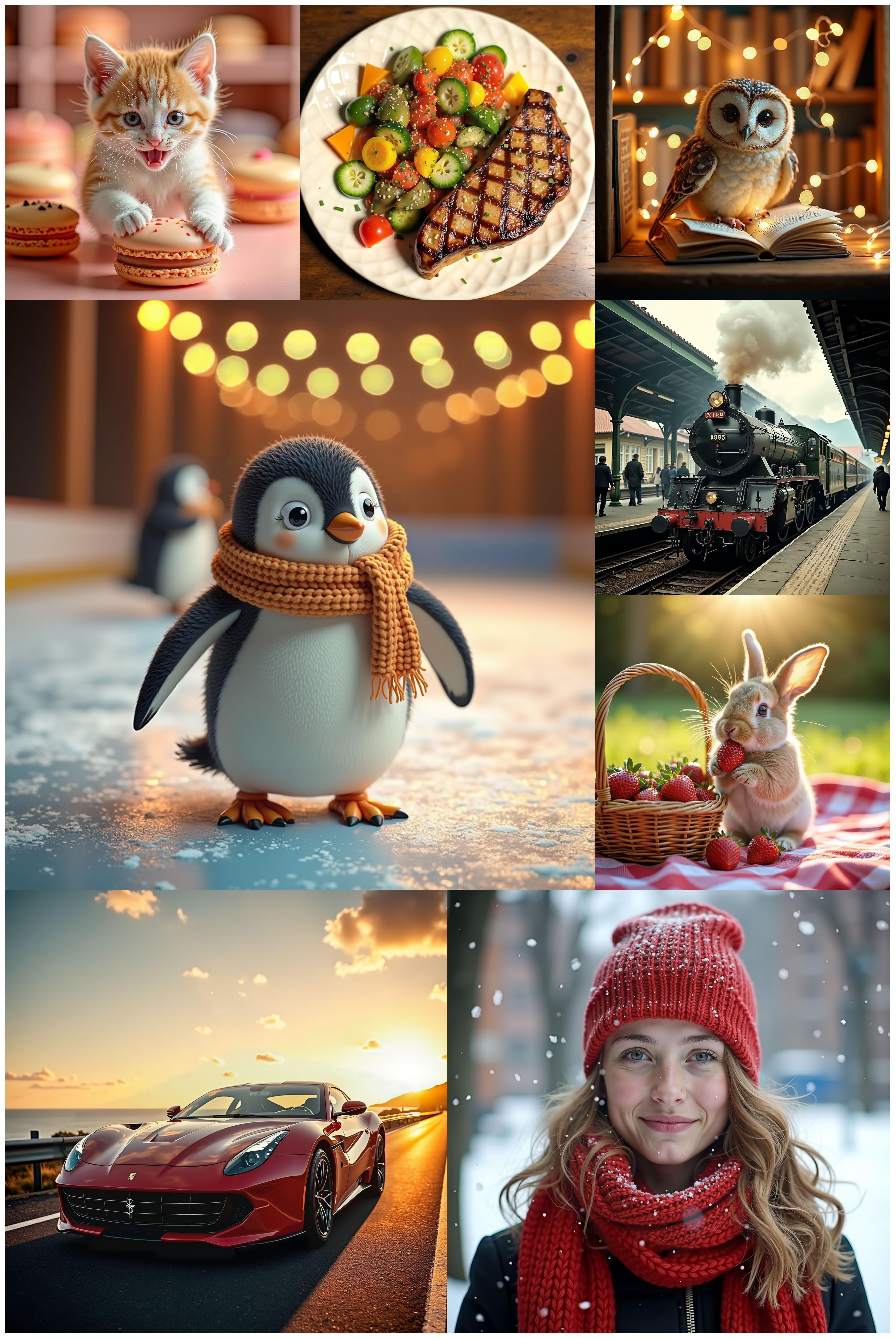} 
\caption{
2048$\times$2048 samples generated by our method, using DyPE \cite{issachar2025dype} for resolution adaptation.
}
\label{fig:flux_2k_ours} 
\end{figure*}

\begin{figure*}[t]
\centering
\includegraphics[width=1\linewidth]{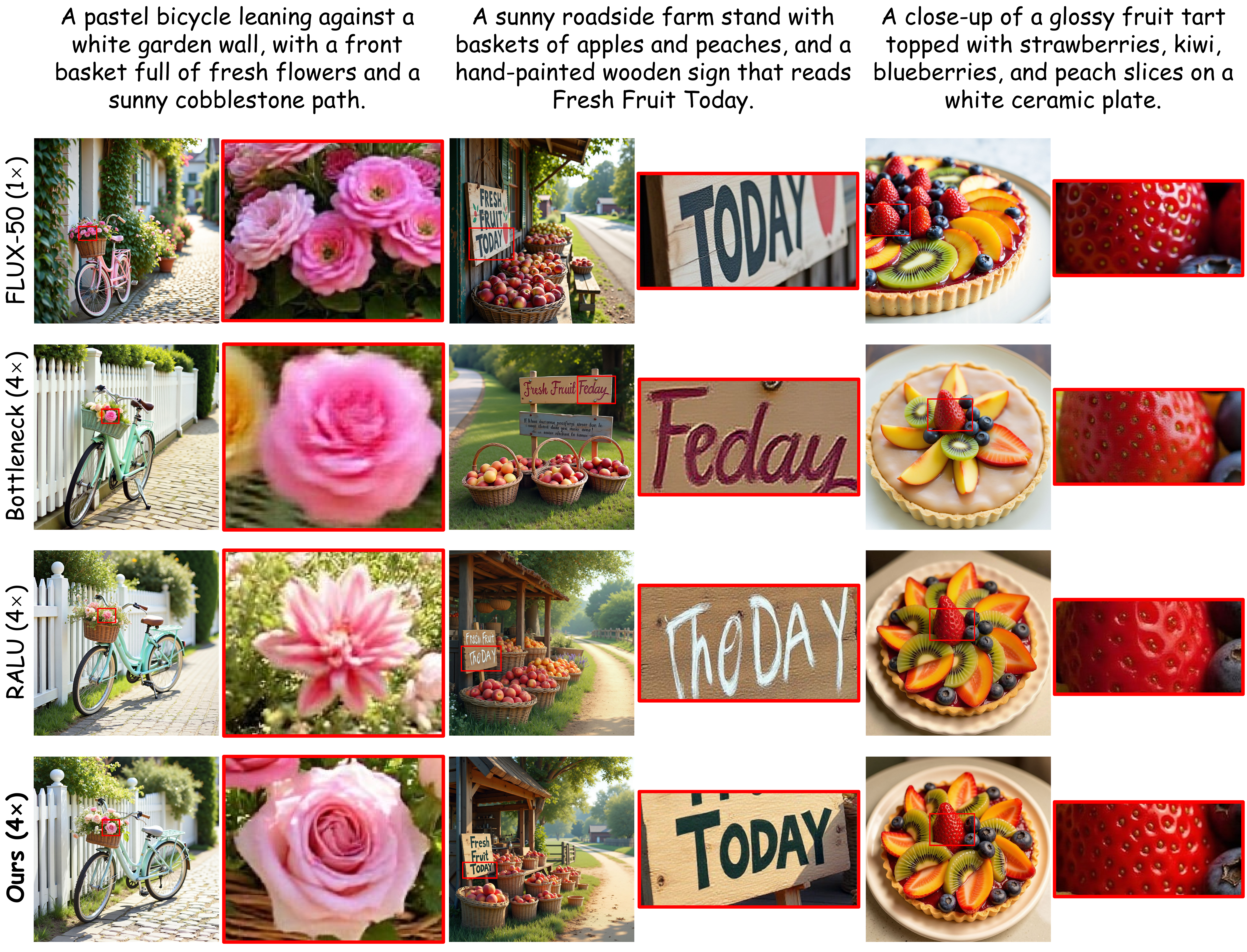}
\caption{\textbf{Comparison with acceleration methods for 2K image generation} on FLUX.1-dev \cite{labs2025flux1kontextflowmatching}. We compare at \(4\times\) speedups, with 50-step FLUX as reference.
% Red boxes highlight the improved level of detail in our generation.
}
\label{fig:rb_more_vis}
\end{figure*}

\begin{figure*}[t]
\centering
\includegraphics[width=0.45\linewidth]{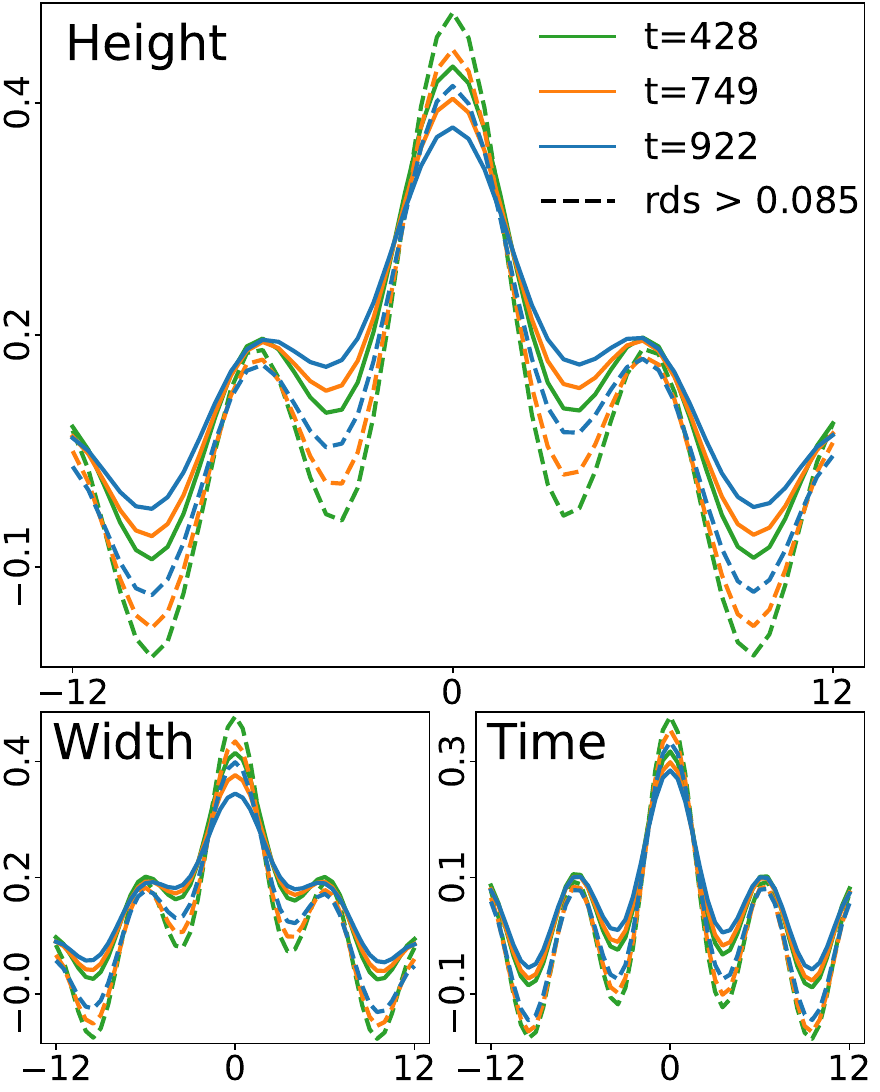}
\hspace{0.05\linewidth}
\includegraphics[width=0.45\linewidth]{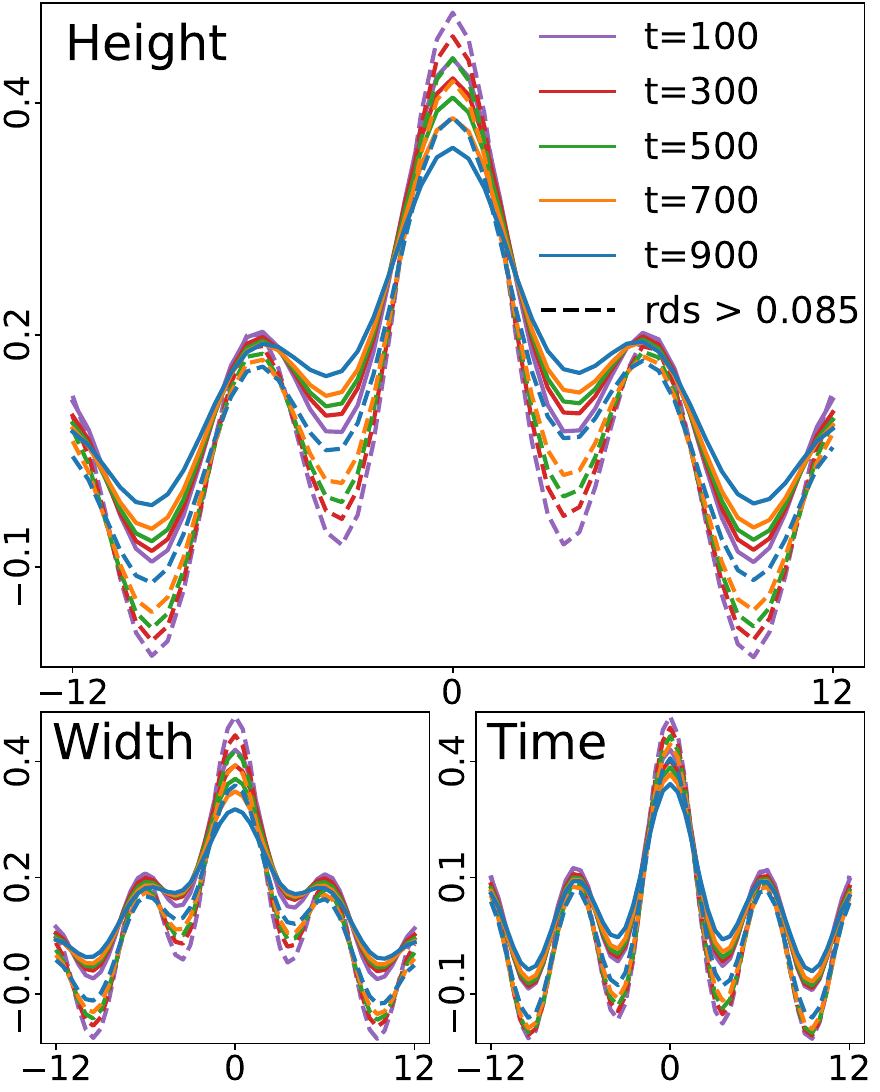}
{
\makebox[0.5\linewidth]{(a) With 1K prompts}%
\makebox[0.5\linewidth]{(b) Across broader, uniform timesteps}%
} 
\caption{Additional $\kappa(\Delta)$ curves.}
\label{fig:rb_kappa}
\end{figure*}

\Cref{fig:flux_2k_ours}:
\begin{itemize}
    \item A kitten inside a pastel macaron bakery display, warm bright lighting and creamy orange-pink tones. The kitten paws at a macaron, then boops it so it rolls toward the camera.
    \item A plate of colorful vegetables and a cut of meat.
    \item A cute owl in a tiny glowing bookshop nook under warm fairy lights, bright cozy palette. The owl flips a book page with its wing, pages fluttering softly.
    \item A chubby penguin wearing a cozy orange scarf on a warm-lit indoor ice rink set, bright amber lighting (stylized). The penguin does a cute little spin and waddles forward.
    \item Vintage train arriving at an old European station, steam billowing, passengers waiting on the platform, nostalgic mood, rich cinematic details.
    \item A bunny at a strawberry picnic on a gingham blanket in warm sunlight, bright warm palette. The bunny nibbles a strawberry and wiggles its nose; strawberries gently bounce as the bunny shifts. 
    \item Glossy red sports car parked on an empty coastal highway at sunrise, dramatic low angle shot, polished reflections, cinematic automotive commercial style.
    \item A close-up cinematic shot of a young woman standing outdoors in winter, wearing a bright red knit beanie and matching red scarf, soft natural daylight illuminating her face. She has gentle wavy light brown hair flowing slightly in the cold breeze. The background is softly blurred with cool gray urban tones, creating a shallow depth of field. She looks directly into the camera with a subtle, warm smile. Light snow begins to fall slowly around her. The camera gently pushes in, capturing fine details of her expression and the texture of the knit fabric. Soft winter ambiance, natural color grading, realistic skin tones, 4K, shallow depth of field, cinematic.
\end{itemize}

\clearpage

% put figures, except the curve, here

\begin{table*}[t]
\centering
\caption{Detailed quantitative evaluation results on VBench \cite{huang2024vbench}, corresponding to Table~1 in the main text. The \% symbol is omitted.}
\label{tab:wan_rope_detail_vbench}
% \vspace{-2mm}
\resizebox{\linewidth}{!}{
\begin{tabular}{l|cccccccccccccccc|c}
\toprule
Method &
\rotatebox{90}{Scene} &
\rotatebox{90}{Temporal Style} &
\rotatebox{90}{Overall Consistency} &
\rotatebox{90}{Human Action} &
\rotatebox{90}{Temporal Flickering} &
\rotatebox{90}{Motion Smoothness} &
\rotatebox{90}{Dynamic Degree} &
\rotatebox{90}{Spatial Relationship} &
\rotatebox{90}{Appearance Style} &
\rotatebox{90}{Subject Consistency} &
\rotatebox{90}{Background Consistency} &
\rotatebox{90}{Aesthetic Quality} &
\rotatebox{90}{Imaging Quality} &
\rotatebox{90}{Object Class} &
\rotatebox{90}{Multiple Objects} &
\rotatebox{90}{Color} &
\rotatebox{90}{Time} \\
\midrule
HR & 22.82 & 22.35 & 22.19 & 73.00 & 99.25 & 98.92 & 23.61 & 67.39 & 20.10 & 94.94 & 97.67 & 63.54 & 59.41 & 62.90 & 51.14 & 85.81 & 172.1 s \\
\midrule
PI-LR~\cite{chen2023extending} & 19.62 & 20.78 & 21.40 & 69.00 & 99.59 & 97.92 & 45.83 & 35.11 & 21.32 & 92.72 & 96.02 & 49.13 & 42.80 & 64.40 & 23.48 & 87.93 \\
PI-HR~\cite{chen2023extending} & 8.50 & 19.09 & 20.88 & 67.00 & 97.33 & 94.44 & 26.39 & 22.92 & 22.01 & 90.34 & 94.07 & 48.62 & 40.38 & 55.14 & 23.86 & 78.60 \\
NTK~\cite{peng2023ntk} & 12.57 & 19.63 & 21.59 & 69.00 & 97.98 & 95.27 & 27.78 & 31.23 & 22.17 & 90.71 & 94.02 & 50.22 & 42.37 & 59.26 & 28.43 & 82.34 & 43.2 s \\
PI+NTK & 13.23 & 21.01 & 22.20 & 71.00 & 99.55 & 97.35 & 36.11 & 40.75 & 21.69 & 91.86 & 95.14 & 52.78 & 46.16 & 61.31 & 30.64 & 90.41 \\
YARN~\cite{peng2023yarn} & 18.24 & 21.02 & 22.67 & 71.00 & 99.35 & 97.62 & 36.11 & 39.20 & 21.94 & 91.64 & 94.40 & 53.63 & 51.28 & 65.19 & 32.70 & 83.39 \\
\midrule
Ours & 21.08 & 22.31 & 23.04 & 70.00 & 99.59 & 97.65 & 48.61 & 52.76 & 21.03 & 93.59 & 95.75 & 60.70 & 61.48 & 74.13 & 47.64 & 91.23 & 43.2 s \\
\bottomrule
\end{tabular}
}
% \vspace{-2mm}
\end{table*}

\begin{table*}[t]
\centering
\caption{Detailed quantitative evaluation results on VBench~\cite{huang2024vbench}, corresponding to Table~3 in the main text. The \% symbol is omitted.}
\label{tab:wan_compare_acc_detail_vbench}
% \vspace{-2mm}
\resizebox{\linewidth}{!}{
\begin{tabular}{l|cccccccccccccccc|c}
\toprule
Method &
\rotatebox{90}{Scene} &
\rotatebox{90}{Temporal Style} &
\rotatebox{90}{Overall Consistency} &
\rotatebox{90}{Human Action} &
\rotatebox{90}{Temporal Flickering} &
\rotatebox{90}{Motion Smoothness} &
\rotatebox{90}{Dynamic Degree} &
\rotatebox{90}{Spatial Relationship} &
\rotatebox{90}{Appearance Style} &
\rotatebox{90}{Subject Consistency} &
\rotatebox{90}{Background Consistency} &
\rotatebox{90}{Aesthetic Quality} &
\rotatebox{90}{Imaging Quality} &
\rotatebox{90}{Object Class} &
\rotatebox{90}{Multiple Objects} &
\rotatebox{90}{Color} &
\rotatebox{90}{Time} \\
\midrule
HR & 22.82 & 22.35 & 22.19 & 73.00 & 99.25 & 98.92 & 23.61 & 67.39 & 20.10 & 94.94 & 97.67 & 63.54 & 59.41 & 62.90 & 51.14 & 85.81 & 172.1 s \\
\midrule
UniPC \cite{zhao2023unipc} & 11.99 & 21.12 & 21.97 & 61.00 & 99.32 & 98.55 & 44.44 & 52.82 & 21.21 & 91.82 & 96.82 & 58.10 & 49.81 & 47.23 & 31.55 & 81.19 & 45.6 s \\
DPM++ \cite{lu2025dpm} & 15.04 & 20.70 & 21.69 & 57.00 & 99.34 & 98.46 & 47.22 & 52.86 & 21.29 & 91.43 & 96.99 & 56.61 & 46.48 & 42.96 & 25.84 & 72.13 & 44.7 s \\
ToMe \cite{bolya2023token} & 2.11 & 10.99 & 10.48 & 6.00 & 98.36 & 98.39 & 36.11 & 3.93 & 22.61 & 89.84 & 95.59 & 31.86 & 24.26 & 7.44 & 1.14 & 100.00 & 48.0 s \\
TeaCache \cite{liu2025timestep} & 17.95 & 19.75 & 21.27 & 61.00 & 99.28 & 98.97 & 22.22 & 51.04 & 20.82 & 92.99 & 97.27 & 58.36 & 46.11 & 49.13 & 27.21 & 82.98 & 43.5 s \\
MagCache \cite{ma2025magcache} & 21.88 & 21.16 & 21.51 & 66.00 & 99.12 & 99.00 & 16.67 & 59.80 & 20.33 & 94.62 & 97.25 & 59.94 & 52.85 & 55.54 & 42.23 & 82.68 & 45.5 s \\
\midrule
Ours & 21.08 & 22.31 & 23.04 & 70.00 & 99.59 & 97.65 & 48.61 & 52.76 & 21.03 & 93.59 & 95.75 & 60.70 & 61.48 & 74.13 & 47.64 & 91.23 & 43.2 s \\
\bottomrule
\end{tabular}
}
% \vspace{-2mm}
\end{table*}

\end{document}